\documentclass[final,leqno,onefignum,onetabnum]{siamltex1213}

\usepackage{lipsum}

\usepackage{cite}

\usepackage[labelformat=parens]{subcaption}

\graphicspath{{./figures_compressed/}}

\usepackage{tikz}
\usetikzlibrary{arrows}

\usepackage{colortbl}
\definecolor{LightYellow}{rgb}{0.9961, 0.8510, 0.6510}
\definecolor{LightGreen}{rgb}{0.8, 0.9216, 0.7725}

\usepackage{amsmath, amssymb, mathrsfs, eufrak}
\usepackage{mathtools}

\usepackage{amsthm}

\DeclareMathOperator*{\argmax}{\arg\!\max}
\DeclareMathOperator*{\argmin}{\arg\!\min}
\newcommand{\vect}[1]{{\mathbf{#1}}}
\newcommand{\mat}[1]{{\mathbf{#1}}}
\newcommand{\set}[1]{{\mathcal{#1}}}
\newcommand{\norm}[2]{\left\| #1 \right\|_{#2}}
\newcommand{\transpose}[1]{#1^\mathrm{T}}

\newcommand{\Real}{{\mathbb R}}
\newcommand{\Nat}{{\mathbb N}}

\newcommand{\bintail}[3]{{\mathcal{B}} \left( #1,#2;#3\right)}
\newcommand{\eps}{\varepsilon}
\newcommand{\meps}{$\eps$\xspace}

\usepackage{footnote}
\usepackage{tabularx}
\usepackage{booktabs}
\usepackage{array}
\usepackage{rotating}
\usepackage{multirow}
 \usepackage{dcolumn}
\newcolumntype{d}[1]{D{.}{.}{#1}}
\newcolumntype{.}{D{.}{.}{-1}}
\makeatletter
\newcolumntype{B}[1]{>{\boldmath\DC@{.}{.}{#1}}c<{\DC@end}}
\makeatother

\usepackage[algo2e,ruled,vlined,linesnumbered]{algorithm2e}
\SetKwComment{Comment}{//\ }{}

\title{A BI-CLUSTERING FRAMEWORK FOR CONSENSUS PROBLEMS\thanks{This work was partially supported by NSF, ONR, NGA, ARO, and NSSEFF.}}

\author{Mariano Tepper\footnotemark[2] and Guillermo Sapiro\footnotemark[2]}

\begin{document}
\maketitle
\slugger{mms}{xxxx}{xx}{x}{x--x}%slugger should be set to mms, siap, sicomp, sicon, sidma, sima, simax, sinum, siopt, sisc, or sirev

\renewcommand{\thefootnote}{\fnsymbol{footnote}}
\footnotetext[2]{Department of Electrical and Computer Engineering, Duke University, Durham, NC 27708 (\email{mariano.tepper@duke.edu}, \email{guillermo.sapiro@duke.edu})}
\renewcommand{\thefootnote}{\arabic{footnote}}

\begin{abstract}
We consider grouping as a general characterization for problems such as clustering, community detection in networks, and multiple parametric model estimation.
We are interested in merging solutions from different grouping algorithms, distilling all their good qualities into a \emph{consensus} solution.
In this paper, we propose a bi-clustering framework and perspective for reaching consensus in such grouping problems.
In particular, this is the first time that the task of finding/fitting multiple parametric models to a dataset is formally posed as a consensus problem. We highlight the equivalence of these tasks and establish the connection with the computational Gestalt program, that seeks to provide a psychologically-inspired detection theory for visual events.
We also present a simple but powerful bi-clustering algorithm, specially tuned to the nature of the problem we address, though general enough to handle many different instances inscribed within our characterization.
The presentation is accompanied with diverse and extensive experimental results in clustering, community detection, and multiple parametric model estimation in image processing applications.
\end{abstract}


%\begin{AMS}62H35, 62H30, 62H15, 65f30, 68U10, 90C26, 90C90\end{AMS}

\pagestyle{myheadings}
\thispagestyle{plain}
\markboth{M.~TEPPER AND G.~SAPIRO}{A BI-CLUSTERING FRAMEWORK FOR CONSENSUS PROBLEMS}

\section{Introduction}

This paper addresses the problem of grouping data corrupted by noise and outliers.
In our context, the data is a set $\set{X}$ of $m$ elements, described by
\begin{equation}
\set{X} = \left( \bigcup_i \set{X}_i \right) \cup \set{O} \quad \text{with} \quad (\forall i) \set{X}_i \cap \set{O} = \emptyset,
\end{equation}
where each $\set{X}_i$ is a group (the number of groups is unknown), and $\set{O}$ contains the outliers.
Depending on the application context, $\set{O}$ might be empty.
Generically, we consider that a grouping algorithm provides a set $\set{C}$ of candidate groups ($\set{C} \subset \mathbb{P} (\set{X})$, where $\mathbb{P} (\set{X})$ is the power set of $\set{X}$). The groups in $\set{C}$ do not need to form a partition of $\set{X}$ nor to cover $\set{X}$.
This general characterization subsumes many different grouping problems. For example, in this paper we address the following problems:
\begin{itemize}
	\item If $\set{C}$ is a partition of $\set{X}$, then we are considering \textbf{clustering} algorithms. Traditionally, these algorithms assume $\set{O} = \emptyset$. For a broad perspective on clustering techniques, we refer the reader to the overview reported in~\cite{jain09}.
	\item If our dataset is a network, i.e., $\set{X}$ are its vertices, then we have a \textbf{community detection} problem (notice that communities are not required to form a partition of $\set{X}$). See~\cite[and references therein]{plantie13} for a comprehensive survey on community detection.
	\item If $\set{C}$ contains a single group that was obtained by fitting a parametric model, we are dealing with a \textbf{parametric model estimation} problem.
\end{itemize}
Of course, many other problems fit into this characterization, such as image and video segmentation, or object detection in images and videos, to name just a few.
Let us consider that we are provided with a pool (universe) $\{ \set{C}_k \}_{k=1}^{c}$ of $c$ such candidate sets. These candidates might come from:
\begin{itemize}
    \item running different grouping algorithms;
    \item running one algorithm with different parameters;
    \item running a grouping algorithm on different modalities of the same data;
    \item all of the above simultaneously.
\end{itemize}
Given the pool $\{ \set{C}_k \}_{k=1}^{c}$, we ask: which is the set of candidates $\set{C}_k$ most representative of the actual grouping structure of $\set{X}$? For this, one would need a criterion to select a specific set and discard all others. This has proven a rather difficult task, where even standard measures, such as modularity or normalized cuts, have known critical shortcomings~\cite{kleinberg02,lancichinetti11}. But why do we have to settle with selecting one solution from the pool? We argue that a better option is to combine the information of the different results in the pool into a new result consistent with them.

Recently, there have been interesting attempts to formulate this consensus problem in a sound way.
Let us assume that we have a quality measure $\omega$ for each group in $\bigcup_{k=1}^{c} \set{C}_k$. We establish a relation that says if two groups $C \in \set{C}_k, C' \in \set{C}_{k'}$ are mutually consistent (e.g., they do not overlap, $C \cap C' = \emptyset$). We then browse through the pool $\bigcup_{k=1}^{c} \set{C}_k$ and build a new solution by combining the subset of mutually consistent groups with higher quality.
This can be posed as a maximum-weight clique problem. This type of formulation was simultaneously introduced in~\cite{brendel10} and~\cite{ion11} for image segmentation and extended for clustering~\cite{li2012nips} and community detection~\cite{tepper13comDet}.

These two approaches, picking one candidate set $\set{C}_k$ from the pool or combining all candidate sets to form a new solution, share a common issue: we need a sound and \emph{general} quality measure (in the sense that it should allow to compare algorithms based on different principles, applied to the same data). 
In the clustering field, a plethora of methods to assess or classify clustering algorithms have been developed, some of them with very interesting results, e.g.,~\cite{kleinberg02,kannan04,ben-david08,memoli10}. Unfortunately, the lack of a general definition (i.e., algorithm independent) makes difficult to find a unifying clustering theory and/or measure.
In community detection for example, the best way to establish the community structure of a network is also an elusive task and still disputed, e.g.,~\cite{lancichinetti11}.

An alternative approach is to find and exploit the consistencies between the different grouping candidates in $\{ \set{C}_k \}_{k=1}^{c}$.
Consensus/ensemble clustering is a well known family of techniques used in data analysis to solve this type of problems. Typically, the goal is to search for the so-called ``mean'' (or consensus) partition, i.e., the partition that is most similar, on average, to all the input partitions.

The most common form of consensus clustering involves creating an $m \times m$ co-occurrence matrix (recall $m$ is the number of data elements), defined as
\begin{equation}
	\mat{B} = \tfrac{1}{c} \sum_{k=1}^{c} \mat{B}_k ,
	\quad \text{where} \quad
    (\mat{B}_k)_{ij} =
    \begin{cases}
        1 & \text{if } (\exists C \in \set{C}_k)\ i, j \in C; \\
        0 & \text{otherwise.}
    \end{cases}
	\label{eq:co-occurrence}
\end{equation}
There are many algorithms for analyzing $\mat{B}$, from simple techniques such as applying a clustering algorithm to it (e.g., $k$-means, hierarchical or spectral clustering), to more complex techniques. See~\cite{vegapons2011} for a thorough survey of the area. In one way or another, all these techniques try to find a binary low-rank decomposition of $\mat{B}$ such that
\begin{equation}
    \mat{B}^* = \argmin_{\tilde{\mat{B}} \in \{ 0, 1 \}^{m \times m} } \sum_{k=1}^{c} \norm{\mat{B}_k - \tilde{\mat{B}}}{F}^2 = \argmin_{\tilde{\mat{B}}  \in \{ 0, 1 \}^{m \times m} } \norm{\mat{B} - \tilde{\mat{B}}}{F}^2 .
    \label{eq:li2007}
\end{equation}
In~\cite{li2007}, for example, the task is relaxed into a symmetric non-negative matrix factorization problem, replacing the binary constraint by non-negativity.

In the context of community detection, two works have explicitly addressed consensus within the standard framework just described.
In~\cite{campigotto2013}, the matrix $\mat{B}$ is simply thresholded, and its connected components give the final result. In~\cite{lancichinetti2012consensus}, $\mat{B}$ is considered as the adjacency matrix of a new weighted network. Then the following steps are iteratively applied: (1) a unique non-deterministic algorithm is applied $c$ times, (2) form $\mat{B}$ and threshold it to make it sparse, (3) stop if $\mat{B}$ is block diagonal, and (4) build a new network from $\mat{B}$ and go to (1).

The mentioned aggregation process used to build $\mat{B}$ involves loosing information contained in the individual matrices $\mat{B}_k$. In particular, only pairwise relations are conserved, while relations involving larger groups of nodes might be lost. In addition, using the average of several partitions might not be robust if some of them are of poor quality.
All these methods involve working with an $m \times m$ matrix, which is highly prohibitive when the number of nodes $m$ in the network becomes large.

\textbf{Contributions.}
We propose a novel framework and perspective for reaching consensus in grouping problems by posing them as a bi-clustering problem.\footnote{A preliminary version of this work, restricted to community detection in networks was published in the conference proceedings~\cite{tepper14comDet}.} The proposed approach has two main advantages: (1) all relations are conserved (instead of only keeping pairwise relations) and contribute to the consensus search, and (2) we use a much smaller matrix, rendering the problem tractable for large datasets. We also propose a new \textit{parameterless} bi-clustering algorithm, fit for the type of matrices we analyze. We stress that our goal is not finding a better optimum for the objective function of a given grouping method, but obtaining an overall good solution via consensus search.

In addition, this is the first time that the task of finding/fitting multiple parametric models to a dataset is formally posed as a consensus/bi-clustering problem. The equivalence of these tasks is highlighted by the proposed framework and we devote special attention to explain the rationale behind this new characterization.

Finally, we make a formal connection with the computational Gestalt program~\cite{desolneux08}, that seeks to provide a psychologically-inspired detection theory for visual events. The proposed framework allows to consider this theory from a new perspective both from the statistical and algorithmic viewpoints. We provide some insights that show the suitability of our approach as a new research direction in this field.

\textbf{Organization.}
The remainder of this paper is organized as follows. In Section~\ref{sec:biclustering} we present the proposed approach for our general grouping framework. In sections~\ref{sec:clustering},~\ref{sec:communities}, and~\ref{sec:mpme} we show how this framework applies to the problems of clustering, community detection, and multiple model parametric estimation, respectively. Each of these sections is accompanied with diverse and extensive experimental results. In Section~\ref{sec:computational_gestalt} we discuss the links with the computational Gestalt theory. Finally, we provide some closing remarks in Section~\ref{sec:conclusions}.

\section{Reaching consensus by solving a bi-clustering problem}
\label{sec:biclustering}

\begin{figure}[t]
    \centering

	\begin{subfigure}[b]{.49\textwidth}
		\centering
		\includegraphics[height=.55\columnwidth]{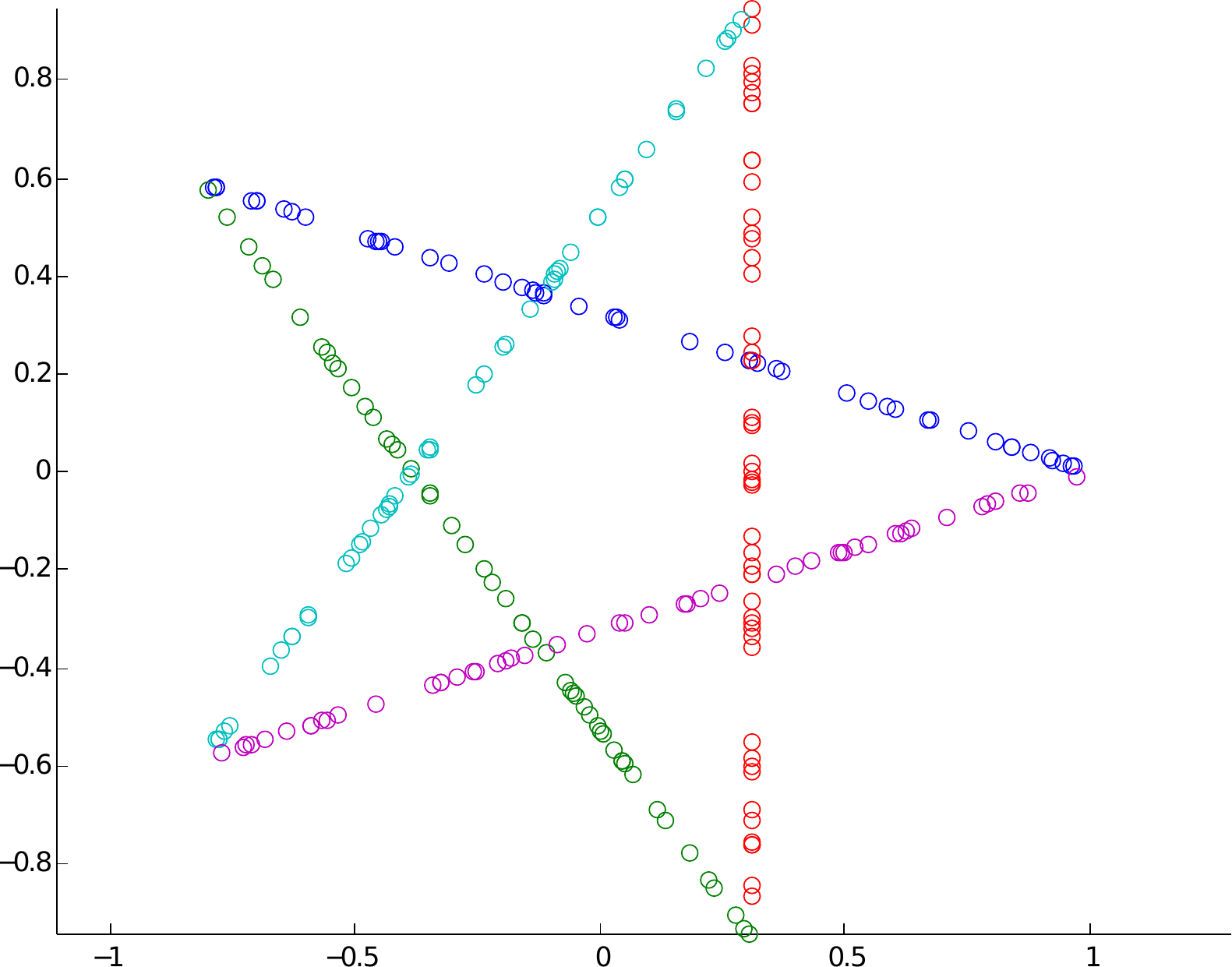}\\
		\includegraphics[width=.8\columnwidth]{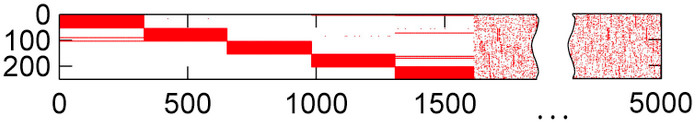}
		
		\caption{Parametric model estimation}
		\label{fig:exampleMPME}
	\end{subfigure}
	\hfill
	\begin{subfigure}[b]{.49\textwidth}
		\centering

		\begin{minipage}[c]{.6\columnwidth}
			\includegraphics[width=\columnwidth]{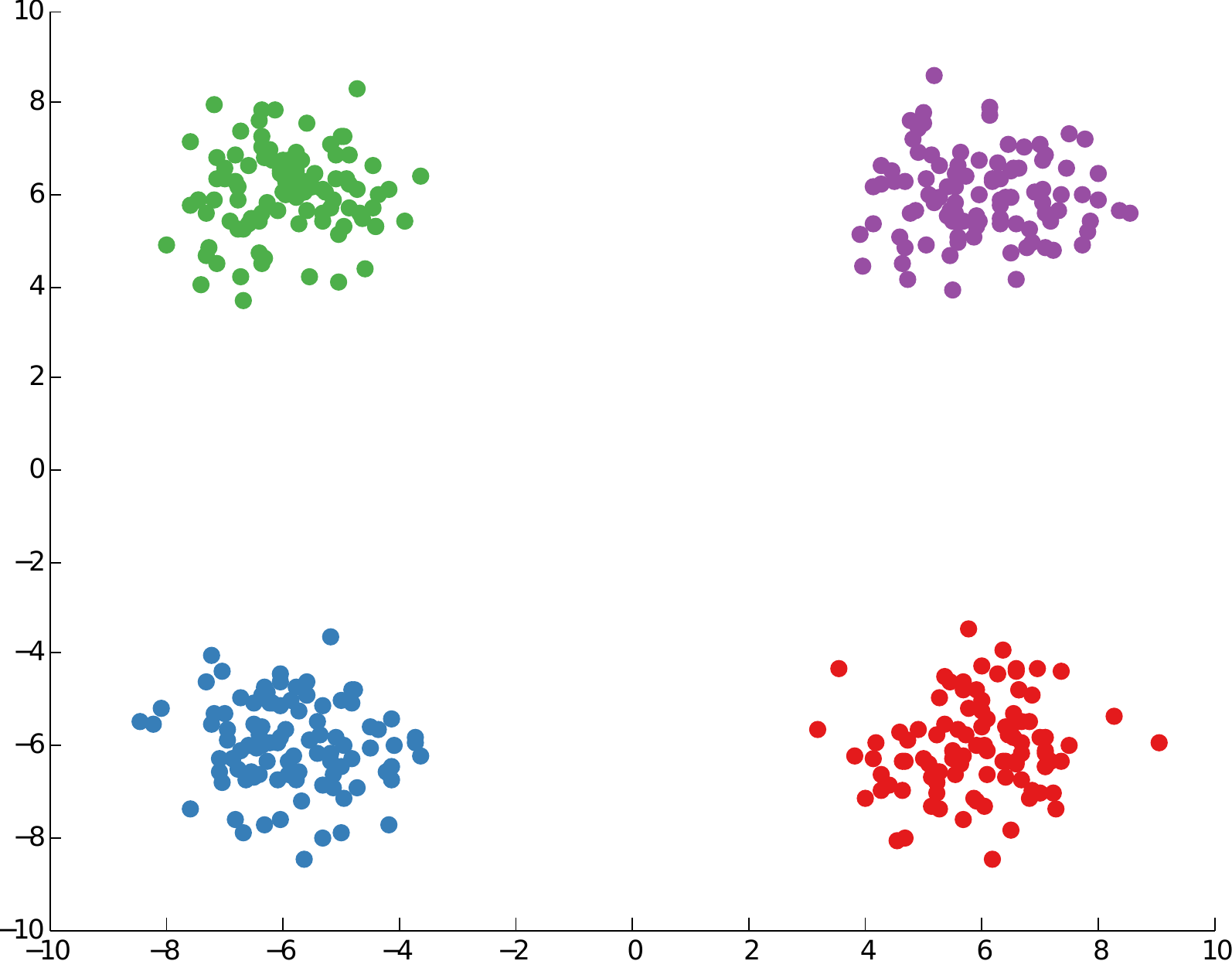}%
		\end{minipage}%
		\hspace{4pt}
		\begin{minipage}[c]{.13\columnwidth}
		\includegraphics[width=\columnwidth]{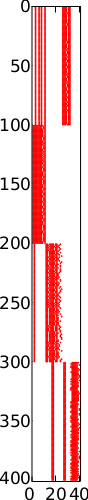}%
		\end{minipage}%
		
		\caption{Clustering}
		\label{fig:exampleClustering}
	\end{subfigure}
	
    \caption{Two examples of preference matrix. \subref{fig:exampleMPME} The data (objects) consist of 250 points on five segments (models) forming a star. The groups are potential parametric models. \subref{fig:exampleClustering} The data (objects) consist of 400 points on four clusters. In both cases  the preference matrix $\mat{A}$ was reordered (permuted) by group for improved visualization.}
    \label{fig:preferenceMatrix}
\end{figure}

The input of the consensus algorithm is a pool $\{ \set{C}_k \}_{k=1}^{c}$ of candidates, that defines the universe of candidates $\set{U} = \bigcup_{k=1}^{c} \mathcal{C}_k$. We also assign a weight $w_j \in \Real^+$ to each group candidate $C_j \in \set{U}$. From the data $\set{X}$ and $\set{U}$, we define an $m \times n$ matrix $\mat{A}$, whose rows and columns represent the $m=|\set{X}|$ data elements and the $n=|\set{U}|$ candidates, respectively; the element $(\mat{A})_{ij} = w_j$ if the $i$-th element belongs to the $j$-th group, and $0$ otherwise. We call $\mat{A}$ a preference matrix. Fig.~\ref{fig:preferenceMatrix} presents two simplified examples.

The group weights indicate the importance assigned to each candidate and can take any form. The simplest form uses uniform weights $(\forall j)\ w_j=1$, in which case $\mat{A}$ becomes a binary matrix. In this case, no prior information is used about the quality of the input group candidates. If we have such information, it can be simply incorporated in these weights.

Although a clustering of elements/objects, using as features the set of sampled models they belong to, might lead to good results in certain applications, it does not fully address the problem at hand. The relationship of the objects with the sampled models is the actual focus of interest (notice the block structure of $\mat{A}$ in Fig.~\ref{fig:preferenceMatrix}).
A pattern-discovery algorithm is needed in this relationship space.
We are thus interested in finding clusters in the product set $\set{X} \times \set{U}$. Such a problem is known in the literature as bi-clustering~\cite[and references therein]{papalexakis2013}, and we are therefore formally connecting it here for the first time with consensus algorithms for grouping problems.

The main contribution of this work is therefore to address the problem of grouping by bi-clustering the preference matrix $\mat{A}$.
This provides a very intuitive rationale since, for each bi-cluster, we are jointly selecting a subset of elements and a subset of groups such that the former belongs to the latter.
By directly analyzing $\mat{A}$ we keep all the information contained in $\set{U}$ ($\mat{A}$ is a complete representation of $\set{U}$).
More classical consensus algorithms analyze an $m \times m$ matrix, see definition~(\ref{eq:co-occurrence}), while we, in contrast, work with a much smaller matrix since for common grouping problems $m \gg n$. In multiple parametric model estimation, commonly $m \ll n$ and we will show how to prune $\set{U}$ to go back to the general scenario.
Another important feature for analyzing very large datasets is that each base algorithm does not need to see the complete dataset. We can split the network in several (preferably overlapping) chunks, run one or more algorithms in each chunk, and let our bi-clustering formulation perform the stitching via consensus.

Notice of course that if all the base algorithms consistently make the same mistakes, these will be translated to the consensus solution, a characteristic common to all consensus algorithms, since there is no natural way to detect such consistent mistake.

\subsection{Solving the bi-clustering problem}

Among many tools for bi-clustering, see~\cite{papalexakis2013} and references therein, the Penalized Matrix Decomposition (PMD)~\cite{witten2009ssvd} and the Sparse Singular Value Decomposition (SSVD)~\cite{lee2010ssvd} have shown great promise, mainly due to their conceptual and algorithmic simplicity.
We are looking for interpretable row-column associations within $\mat{A}$.
Both algorithms iterate two steps until some stopping criterion is met: (1) find one bi-cluster $\{\vect{u}, \vect{v}, s\}$, where $\vect{u} \in \Real^m$, $\vect{v} \in \Real^n$, $s \in \Real^+$; (2) set $\mat{A} = \mat{A} - s \vect{u} \transpose{\vect{v}}$. For step (1), these algorithms solve
\begin{align}
    &\text{(PMD)} &
    \min_{\vect{u}, \vect{v}, s} \norm{\mat{A} - s \vect{u} \transpose{\vect{v}}}{F}^2
    \quad \text{s.t.} \quad
    \begin{gathered}
    \norm{\vect{u}}{2} = 1, \norm{\vect{v}}{2} = 1, \\
    \norm{\vect{u}}{1} \leq c_1, \norm{\vect{v}}{1} \leq c_2,
    \end{gathered}
    \\
    &\text{(SSVD)} &
    \min_{\vect{u}, \vect{v}, s} \norm{\mat{A} - s \vect{u} \transpose{\vect{v}}}{F}^2 +
    \lambda_1 \norm{\vect{u}}{1} +
    \lambda_2 \norm{\vect{v}}{1} .
\end{align}
The sparsity-inducing constraints on $\vect{u}$, $\vect{v}$ lead to approximating $\mat{A}$ with only a few rows and columns of $\vect{u} \transpose{\vect{v}}$.
Let $\set{R}, \set{Q}$ be the active sets (the sets of nonzero elements) of $\vect{u}$, $\vect{v}$, respectively.
$\set{R}, \set{Q}$ act as indicators of the presence of a rank-one submatrix in $\mat{A}$: $\set{R}$ selects rows (elements), while $\set{Q}$ selects columns (groups). This behavior makes both PMD and  SSVD very suitable for bi-clustering.

Correctly setting the parameters $c_1, c_2, \lambda_1, \lambda_2$ is crucial, since they determine de size of the bi-clusters (via the sparsity of $\vect{u},\vect{v}$). PMD sets $c_1, c_2$ via cross-validation, while SSVD uses the Bayesian information criterion. In~\cite{ramirez2013biclustering} a minimum description length criterion is used to set $\lambda_1, \lambda_2$ and the number of iterations for SSVD. In our experiments, finding the correct values for these parameters has proven extremely challenging, since each experiment needs specifically tuned set of values. This motivates in part the development of the algorithm described next.

We propose to follow a different path for solving the bi-clustering problem at hand.
Let us first notice that non-negative matrix factorization (NMF)~\cite{paatero94} pursues a similar objective.
For $1 \leq q \leq \min \{ m, n \}$, NMF solves the problem
\begin{equation}
    \min_{\substack{ \mat{X} \in \Real^{m \times q}, \mat{Y} \in \Real^{q \times n}}} \norm{\mat{A} - \mat{X} \mat{Y}}{F}^2
    \quad \text{s.t.} \quad
    \mat{X}, \mat{Y} \geq 0.
    \label{eq:classical_nmf}
\end{equation}
$\mat{A}$ is, in our application, a sparse non-negative matrix. Notice that the positivity constraints on $\mat{X}, \mat{Y}$ have a sparsifying effect on them. The intuition behind this is that when approximating a sparse non-negative matrix, the non-negative factors will only create a sparse approximation if they are themselves sparse.
We thus obtain sparse factors $\mat{X}, \mat{Y}$ as in PMD and SSVD without introducing any (difficult to set) parameters.
Another consequence of the sparsity of $\mat{A}$ is that the Frobenius norm is not entirely well suited for analyzing it. It is more appropriate to use instead an L1 fitting term,
\begin{equation}
    \min_{\substack{ \mat{X} \in \Real^{m \times q}, \mat{Y} \in \Real^{q \times n}}} \norm{\mat{A} - \mat{X} \mat{Y}}{1}
    \quad \text{s.t.} \quad
    \quad \mat{X}, \mat{Y} \geq 0 .
    \label{eq:nmf}
\end{equation}
With this change, we are now aiming at obtaining a ``median'' type of result instead of the mean, providing robustness to poor group candidates present in the pool (i.e., columns of $\mat{A}$ in our representation).

Any standard NMF algorithm can be adapted to use the L1 norm and solve~(\ref{eq:nmf}); in this work, we modify the method in~\cite{xu2012nmf}, that has shown good performance in practice. The algorithmic details are provided in Appendix~\ref{sec:nmf_algorithm}.

A challenge with NMF is that $q$ is not an easy parameter to set. To avoid a cumbersome decision process, we propose to set $q=1$ and inscribe the L1-NMF approach in an iterative loop, as with PMD and SSVD. The rank-one factorization $\mat{X} \mat{Y}$ will thus approximate a subset of $\mat{A}$ (because of the sparsity-inducing L1-norm), correctly detecting a single bi-cluster. Note that this is related to a common problem in clustering known as masking, where a conceptually similar iterative procedure can address the issue~\cite{tepper11mstClustering}.

\subsection{Algorithmic decisions and parameters}

Algorithm~\ref{algo:biclustering} summarizes the proposed non-negative bi-clustering approach. Notice that instead of subtracting the product $\mat{X} \mat{Y}$ from $\mat{A}$ as in PMD and SSVD, we set the corresponding rows and columns to zero, enforcing disjoint active sets between the successive $\mat{X}_t$ and $\mat{Y}_t$, and hence orthogonality. This also ensures that non-negativity is maintained throughout the iterations. If the bi-clusters are allowed to share elements, we do not change the rows of $\mat{A}$. The proposed algorithm is very efficient, simple to code, and demonstrated to work well in the experimental results that we will present later.

The iterations should stop (1) when $\mat{A}$ is empty (line~\ref{algo:biclustering_stop1}), or (2) when $\mat{A}$ only contains noise (no structured patterns). The second case is controlled by the parameters $\tau_{\text{R}}$ and $\tau_{\text{C}}$ (line~\ref{algo:biclustering_stop2}). These parameters determine the minimum bi-cluster size, $\tau_{\text{R}}$ rows and $\tau_{\text{C}}$ columns, necessary for being considered a structured pattern.
Note that in contrast with the parameters discussed before, these are intuitive, related to the physics of the problem, and easier to set: $\tau_{\text{R}}$ encodes the minimum number of elements that a bi-cluster should contain, while $\tau_{\text{C}}$ encodes the minimum number of candidate groups that need to be in agreement to form a bi-cluster.

In theory, $\tau_{\text{R}}$ and $\tau_{\text{C}}$ depend on $m$, $n$, and the probability of having a non-zero entry in the preference matrix (the probability of success in a Bernoulli trial).
It would be interesting to further explore these dependencies from a theoretical point of view.
In all experiments in this paper, we set $\tau_{\text{C}} = 3$. For clustering and community detection, we use $\tau_{\text{R}} = 6$ for all experiments.
The number and the diversity of the experiments in which these values work show that, in practice, they do not need to be carefully tuned for each case. This is due in part to the fact that in the clustering and community detection experiments, the bi-clusters are not allowed to intersect, enabling the proposed algorithm to quickly eliminate spurious entries from the preference matrix.

In the case of multiple parametric model estimation, where the bi-clusters do intersect, we set $\tau_{\text{R}} = 1$ and enforce a posteriori a more strict (higher) value for it using the computational Gestalt theory. This adjustment can be done during the bi-clustering process but we decided to do it a posteriori to improve clarity and homogeneity between the different applications.

Notice that, as an alternative, we could consider the proposed bi-clustering algorithm as a lossy compression encoder; set $\tau_{\text{R}} = \tau_{\text{C}} = 1$, and select the first $T$ bi-clusters that yield maximum compression (low $T$) with minimal loss (low fitting error).

\begin{algorithm2e}[t]
%    \DontPrintSemicolon
    \SetKwInOut{Input}{input}\SetKwInOut{Output}{output}

    \begin{small}
    \Input{Preference matrix $\mat{A} \in \Real^{m \times n}$, stopping parameters $\tau_{\text{R}}, \tau_{\text{C}} \in \Nat$}
    \Output{Bi-clusters $\{ (\set{R}_t, \set{Q}_t) \}_{t=1}^{T}$}
            
    $T \gets 0$\;
    \While{$\mat{A} \neq \mat{0}$}{ \nllabel{algo:biclustering_stop1}
        $T \gets T+1$\;

        Solve Problem~(\ref{eq:nmf}) for $\mat{X}, \mat{Y}$ with $q=1$\;
        $\set{R}_T \gets \{ i,\ 1 \leq i \leq m,\ (\mat{X})_{i,1} \neq 0 \}$\;
        $\set{Q}_T \gets \{ j,\ 1\leq j \leq n,\ (\mat{Y})_{1,j} \neq 0 \}$\;

	\If{$|\set{R}_T| \leq \tau_{\text{R}} \lor |\set{Q}_T| \leq \tau_{\text{C}}$}{ \nllabel{algo:biclustering_stop2}
		$T \gets T-1$\;
		break\;
	}

%        \TinyLineComment{application-dependent clause:}
        \If{different bi-clusters are not allowed to share rows}{
            $(\forall i,j) \ i \in \set{R}_T, 1\leq j \leq n, \; (\mat{A})_{ij} \gets 0$\;
        }
        $(\forall i,j) \ 1 \leq i \leq m, j \in \set{Q}_T, \; (\mat{A})_{ij} \gets 0$\;
    }
    \Return $\{ \set{R}_t \}_{t=1}^{T}$, $\{ \set{Q}_t \}_{t=1}^{T}$
    \end{small}
    
%    \caption{Bi-clustering algorithm with inputs $\mat{A} \in \{0,1\}^{N \times K}$, $\lambda_u$, $\lambda_v$.}
    \caption{Bi-clustering algorithm.}
    \label{algo:biclustering}
\end{algorithm2e}

\textbf{Note.}
A seemingly related approach~\cite{li2007} involves using symmetric non-negative matrix factorization (SNMF) to cluster the co-occurrence matrix, defined in Equation~(\ref{eq:co-occurrence}), into $q$ clusters. This is achieved by approximating problem~(\ref{eq:li2007}) with
\begin{equation}
	\min_{ \substack{ \mat{H} \in \Real^{m \times q} \\ \mat{D} \in \Real^{q \times q} } } \norm{\mat{B} -  \mat{H} \mat{D} \transpose{\mat{H}} }{F}^2
	\quad \text{s.t.} \quad
	\begin{gathered}
	\mat{H}, \mat{D} \geq 0 ,\\
	\transpose{\mat{H}} \mat{H} = \mat{I} ,
	\end{gathered}
	\label{eq:li2007snmf}
\end{equation}
where $\mat{H} \mat{D} \transpose{\mat{H}}$ provides a non-negative low-rank approximation of $\mat{B}$, and $\mat{D}$ is a matrix that encodes the sizes of the $q$ obtained clusters. As in all matrix decomposition methods, the number of clusters $q$ is a critical parameter and it is in fact one of the parameters we are interested in discovering!

Digging deeper, we point out that
\begin{equation}
	\sum_{i=1}^{n} (\mat{A})_i \transpose{(\mat{A})_i} = \mat{B} ,
	\label{eq:AAt_B}
\end{equation}
where $(\mat{A})_i$ is a column of $\mat{A}$ and, when bi-clusters are not allowed to share rows, by construction we have $(\forall t, t'), t \neq t', \set{R}_t \cap \set{R}_{t'} \ = \emptyset$. We thus have the same orthogonality constraints as in Equation~(\ref{eq:li2007snmf}). The proposed formulation presents all the benefits of Equation~(\ref{eq:li2007snmf}), but with increased robustness. A double averaging is present in the original problem~(\ref{eq:li2007}) and thus in problem~(\ref{eq:li2007snmf}). First, Equation~(\ref{eq:AAt_B}) acts like a pooling operator in $\mat{A}$, loosing critical information. Second, the Frobenius norm is known to be non-resilient to outliers. On the other hand, our formulation computes a robust median approximation to the preference matrix, which carries all needed information.

\subsection{On the input pool of group candidates}

There is a key ingredient in every consensus problem: the quality of the input group candidates. Performing the consensus of many extremely poor groups will not yield a good consensus solution. Among the candidates fed to any consensus algorithm, there needs to be a certain number of reasonably good groups and, at the same time, not too many bad groups. Otherwise, if the number of bad groups overwhelms the number of good groups, a masking phenomenon will occur and we will be facing an extremely hard pattern-discovery problem.

We will thus assume that, in the general case, we have a set of reasonably good candidate groups, contaminated with a few bad groups (by few we mean a number not overwhelmingly large). In the particular case of parametric model estimation, we have at our disposal more information about the nature of the groups (they are parametric) and thus employ a simple but powerful testing procedure to eliminate the vast majority of bad candidate groups. We are currently investigating how to perform similar tests in non-parametric scenarios.

The nature of the group candidates plays a non-negligible role in the consensus grouping: we would like the mistakes of the grouping algorithms to be as uncorrelated as possible (this is a general observation that applies to all consensus approaches). The algorithms should ideally fit the data well, with their ``mistakes'' being caused by different factors in the data and/or different algorithmic decisions/artifacts and hence not systematically appearing over and over. This a key observation that is not trivially enforced in practice; often it is enough to use a pool of candidates that exhibit some consistency in the results while keeping some variety.

Following the presentation of our general consensus grouping framework and procedure, we now proceed to show different application contexts. We focus on clustering, community detection in networks, and multiple parametric model estimation.

% !TEX root = main.tex
\section{Consensus clustering}
\label{sec:clustering}

\begin{figure}[t]
	\centering
	\begin{small} 
	
	Base algorithms\\[6pt]
	\begin{tabular}{ m{.02\textwidth} *{5}{ @{\hspace{4pt}} m{.14\textwidth}} }
	
		\begin{sideways}GMM\end{sideways} &
		\includegraphics[width=.14\textwidth]{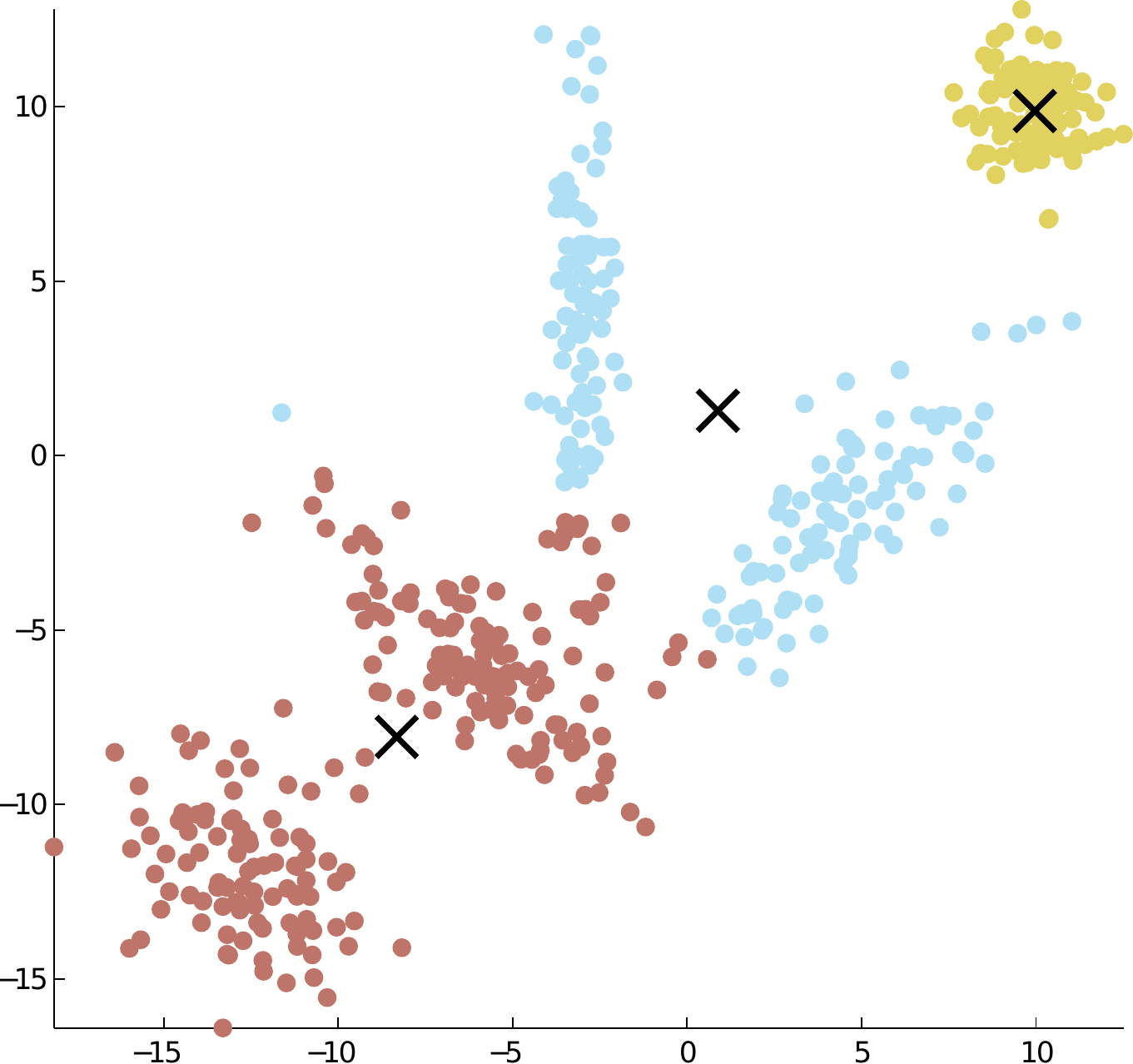} &
		\includegraphics[width=.14\textwidth]{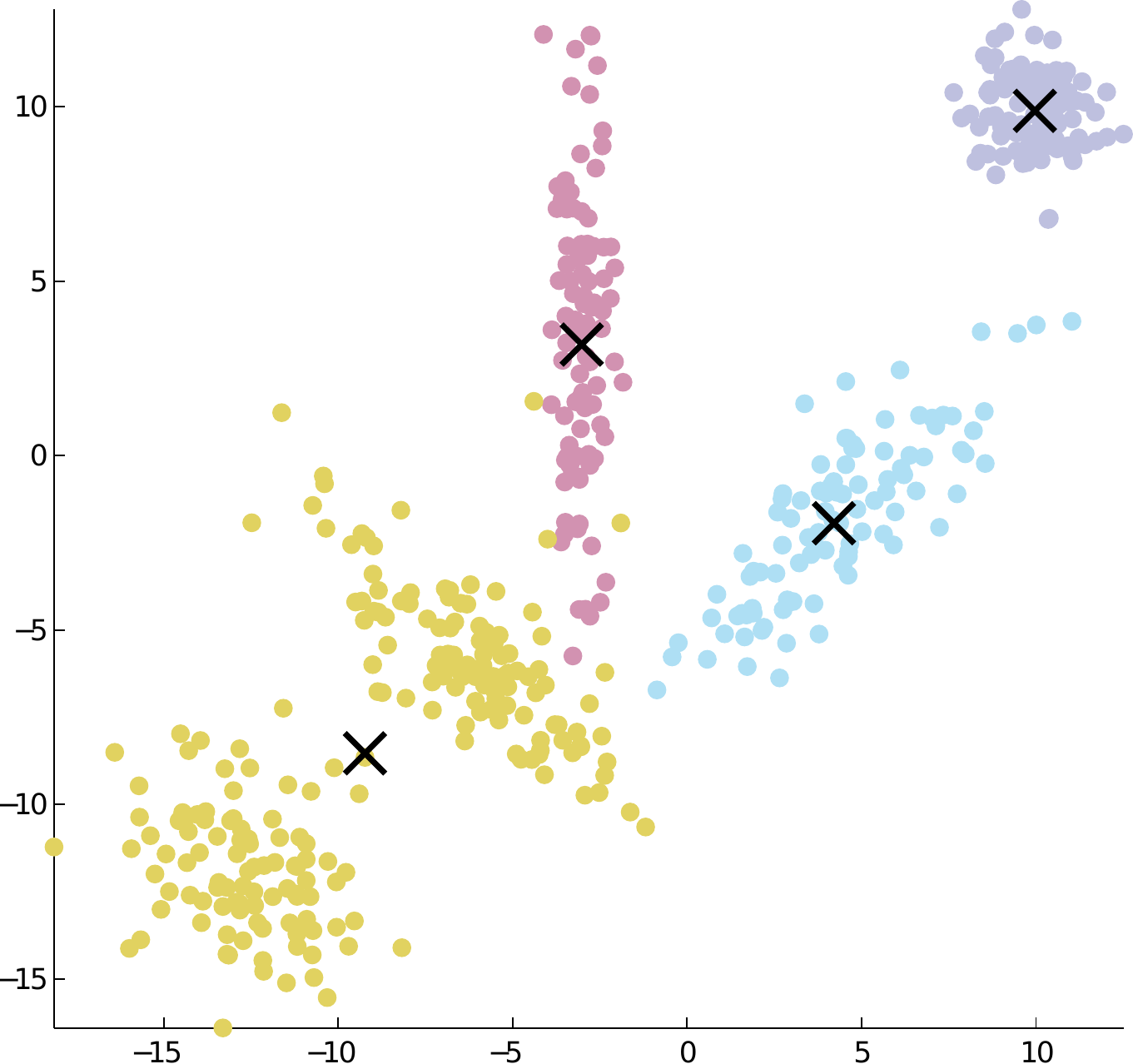} &
		\includegraphics[width=.14\textwidth]{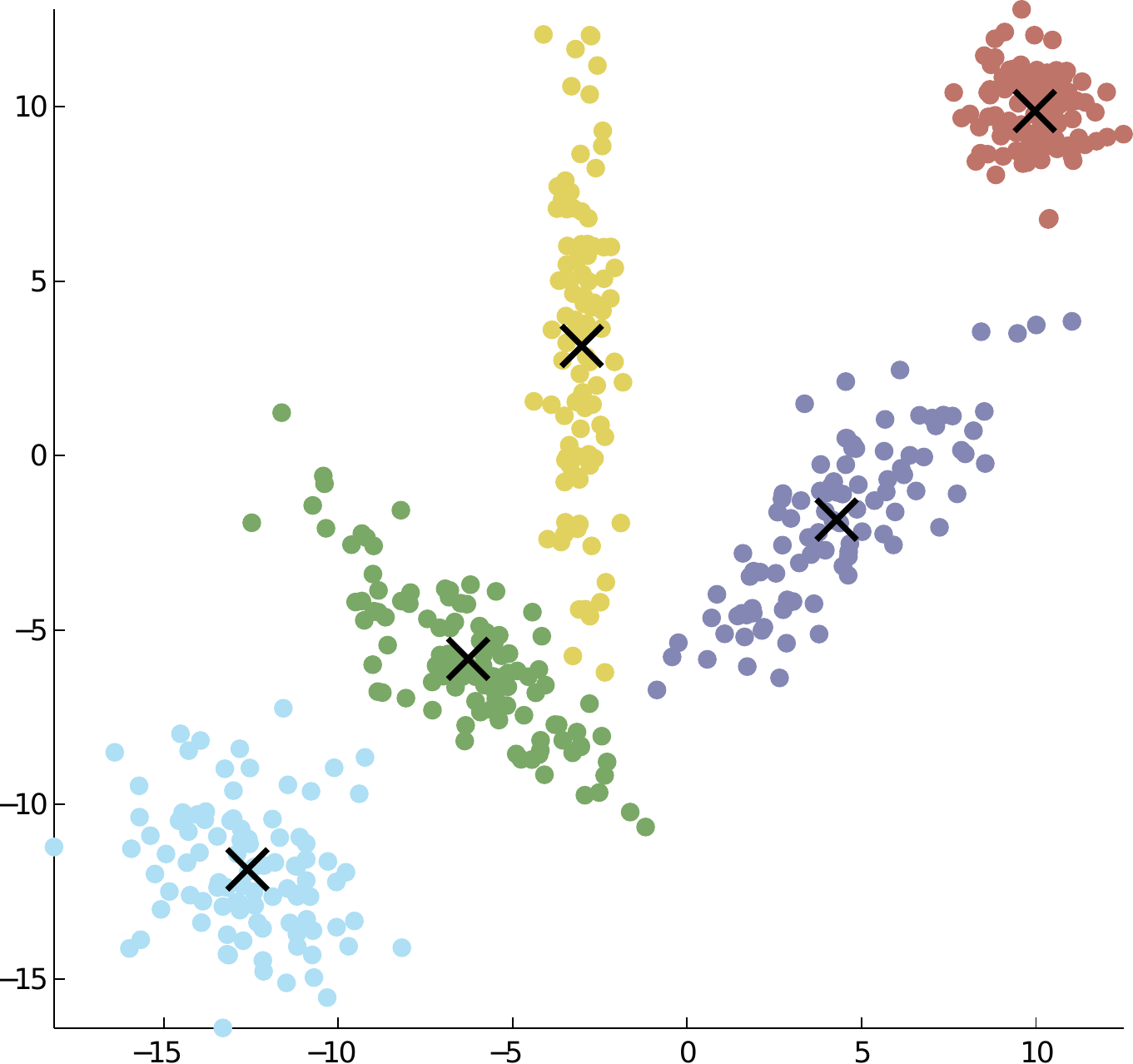} &
		\includegraphics[width=.14\textwidth]{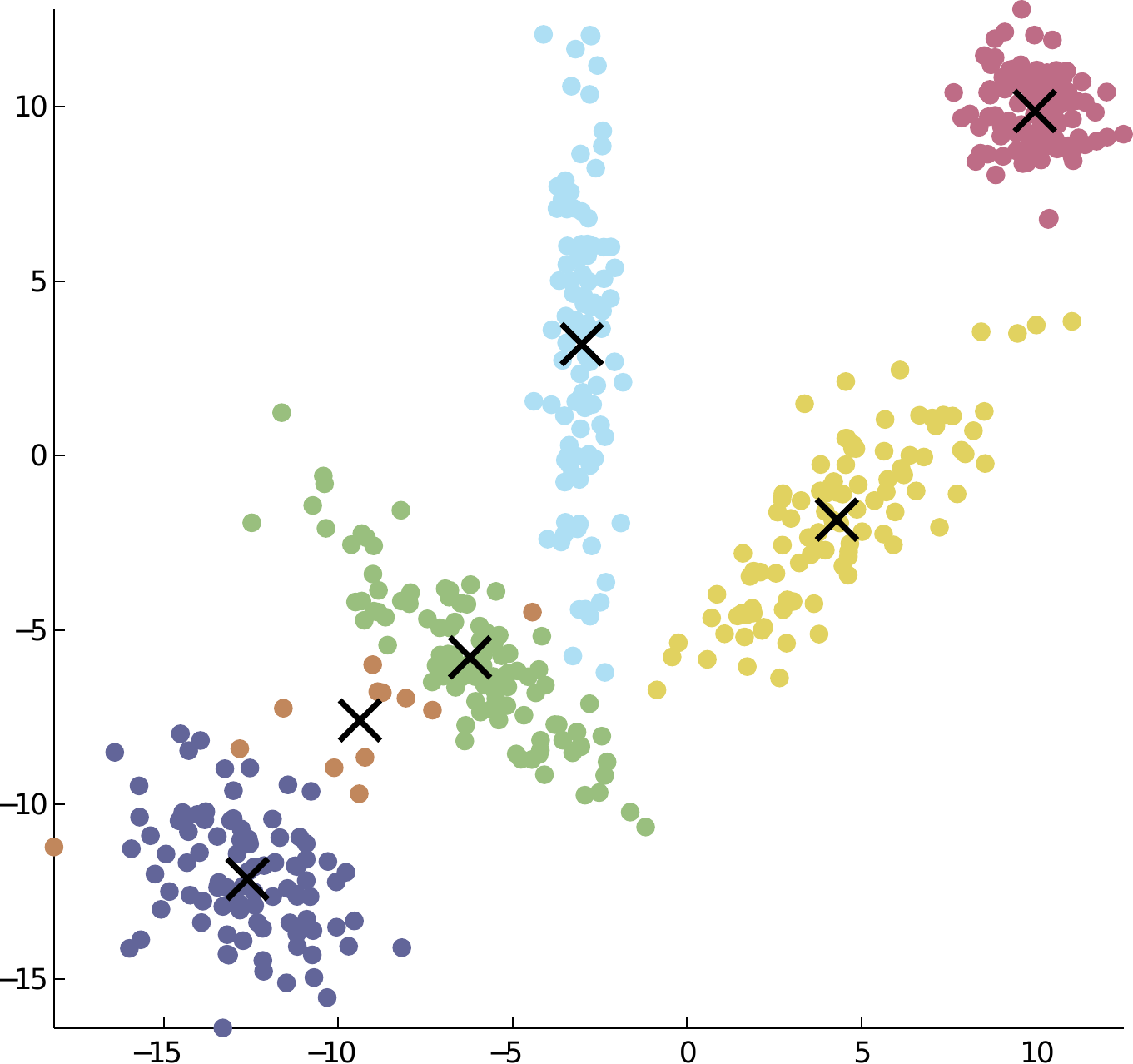} &
		\includegraphics[width=.14\textwidth]{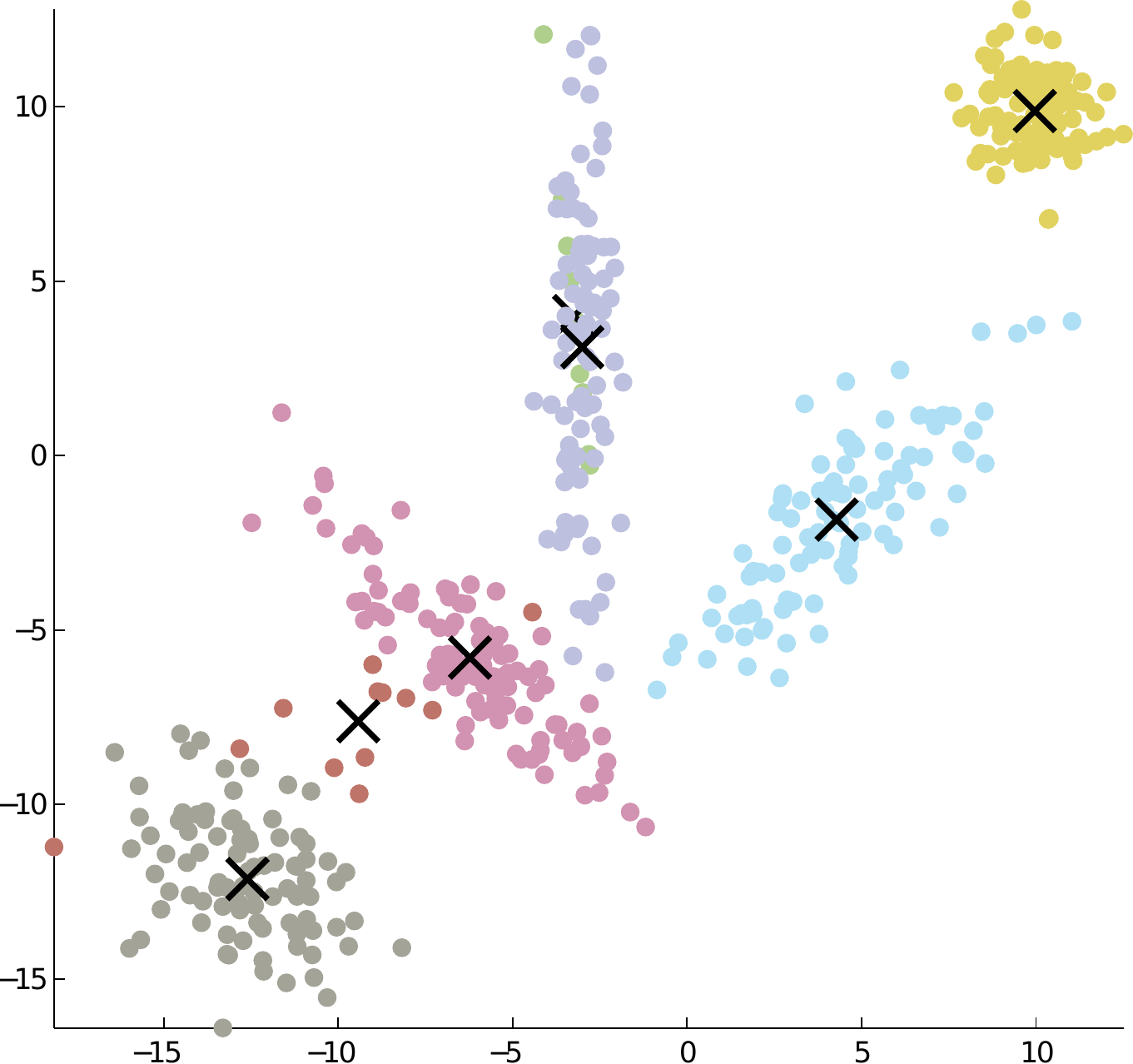} \\	

		&
		\multicolumn{1}{c}{\textsc{nmi}=0.738} &
		\multicolumn{1}{c}{\textsc{nmi}=0.800} &
		\multicolumn{1}{c}{\textsc{nmi}=0.903} &
		\multicolumn{1}{c}{\textsc{nmi}=0.880} &
		\multicolumn{1}{c}{\textsc{nmi}=0.865} \\

		&
		\multicolumn{1}{c}{\textsc{f-m}=72.52} &
		\multicolumn{1}{c}{\textsc{f-m}=82.42} &
		\multicolumn{1}{c}{\textsc{f-m}=95.79} &
		\multicolumn{1}{c}{\textsc{f-m}=94.67} &
		\multicolumn{1}{c}{\textsc{f-m}=93.60} \\[6pt]		
		
		\begin{sideways}$K$-means\end{sideways} &
		\includegraphics[width=.14\textwidth]{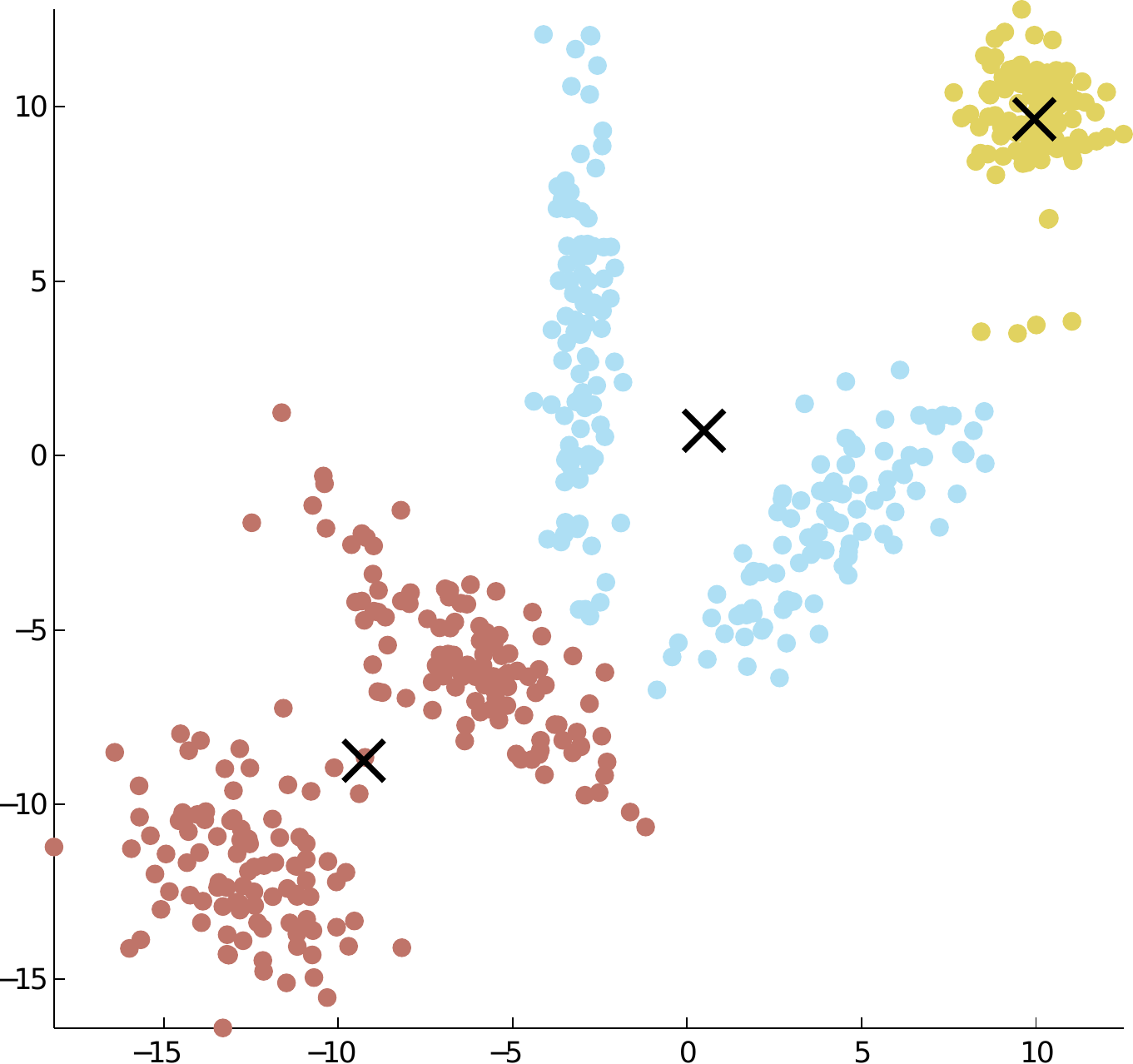} &
		\includegraphics[width=.14\textwidth]{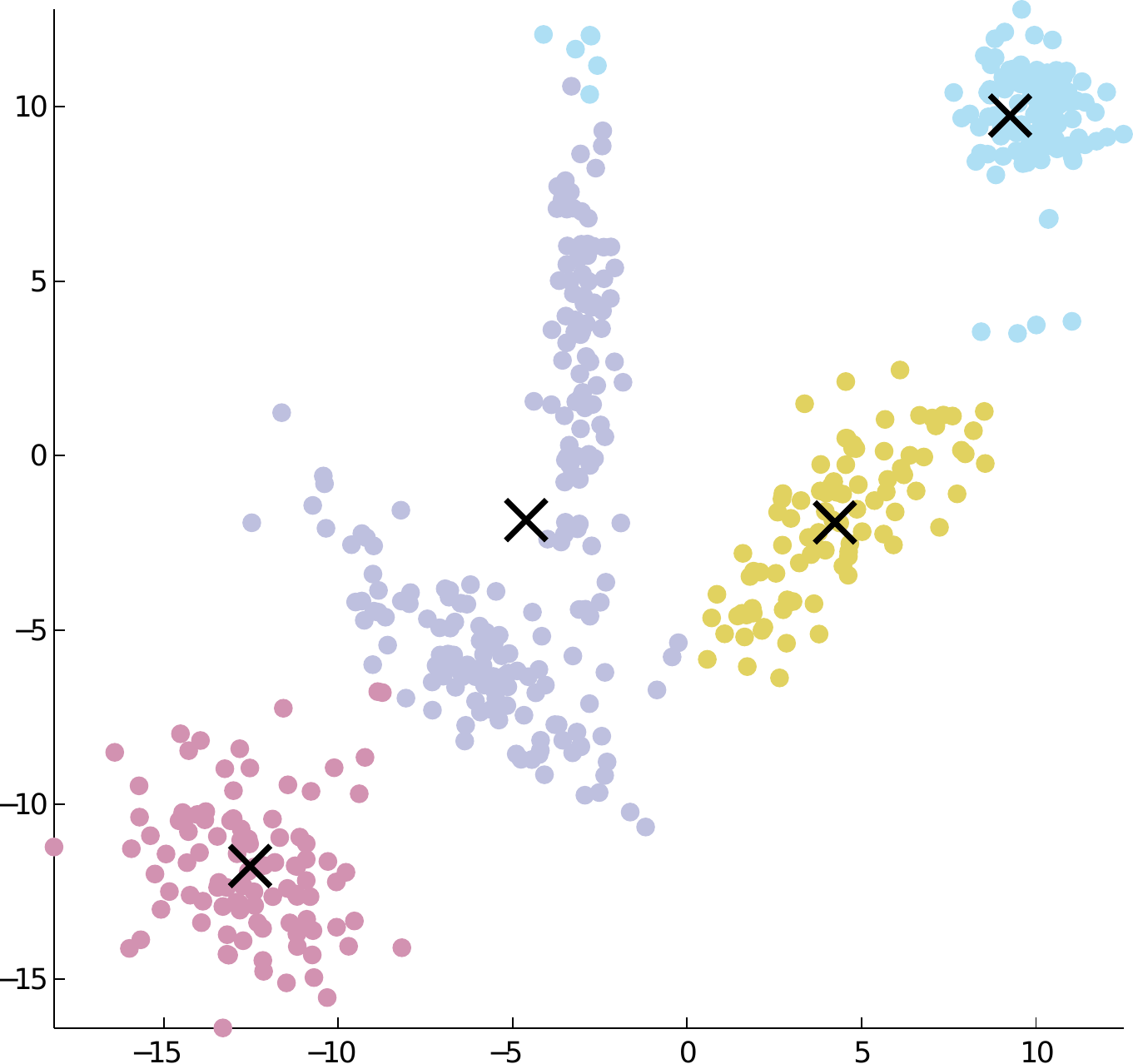} &
		\includegraphics[width=.14\textwidth]{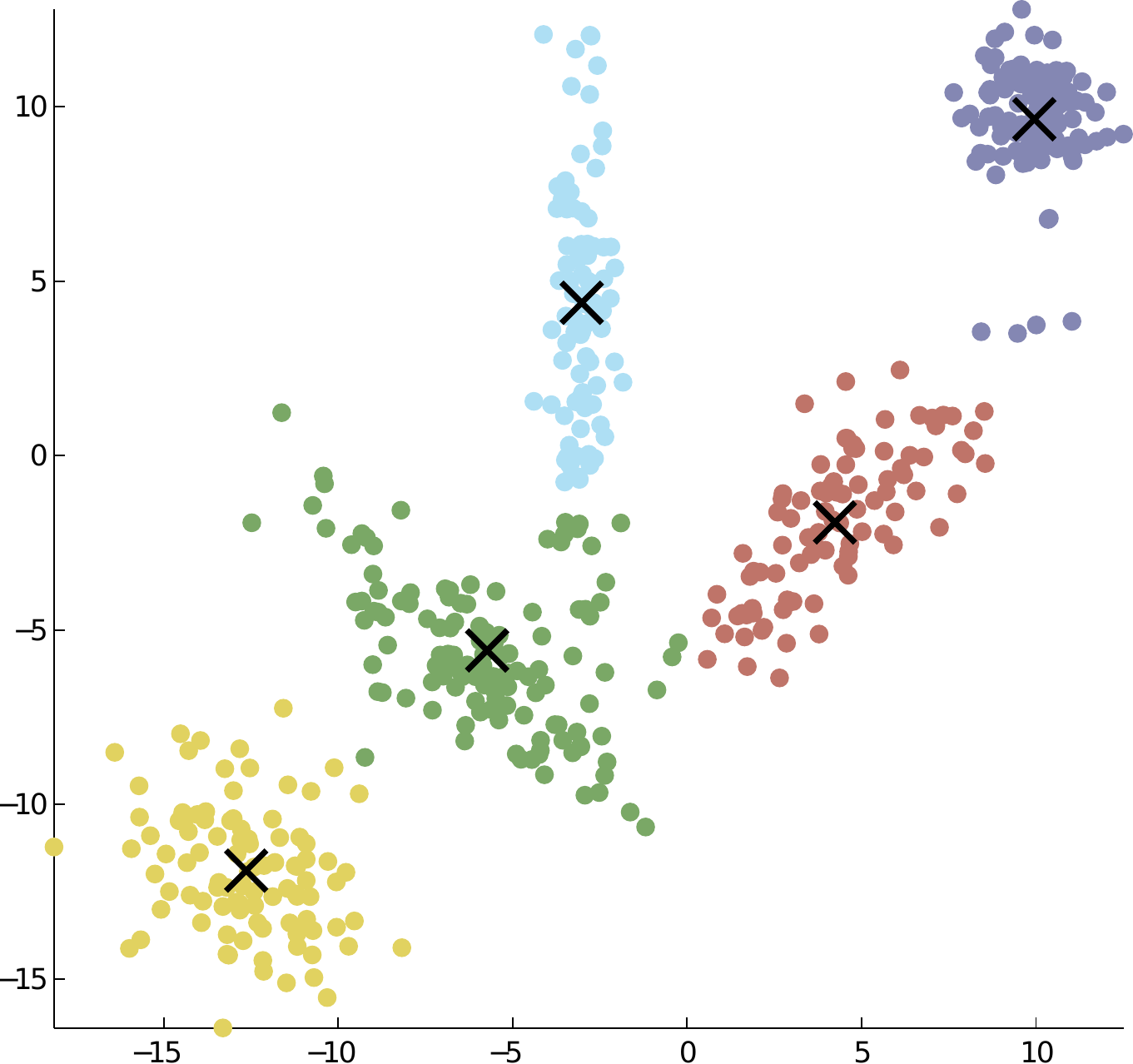} &
		\includegraphics[width=.14\textwidth]{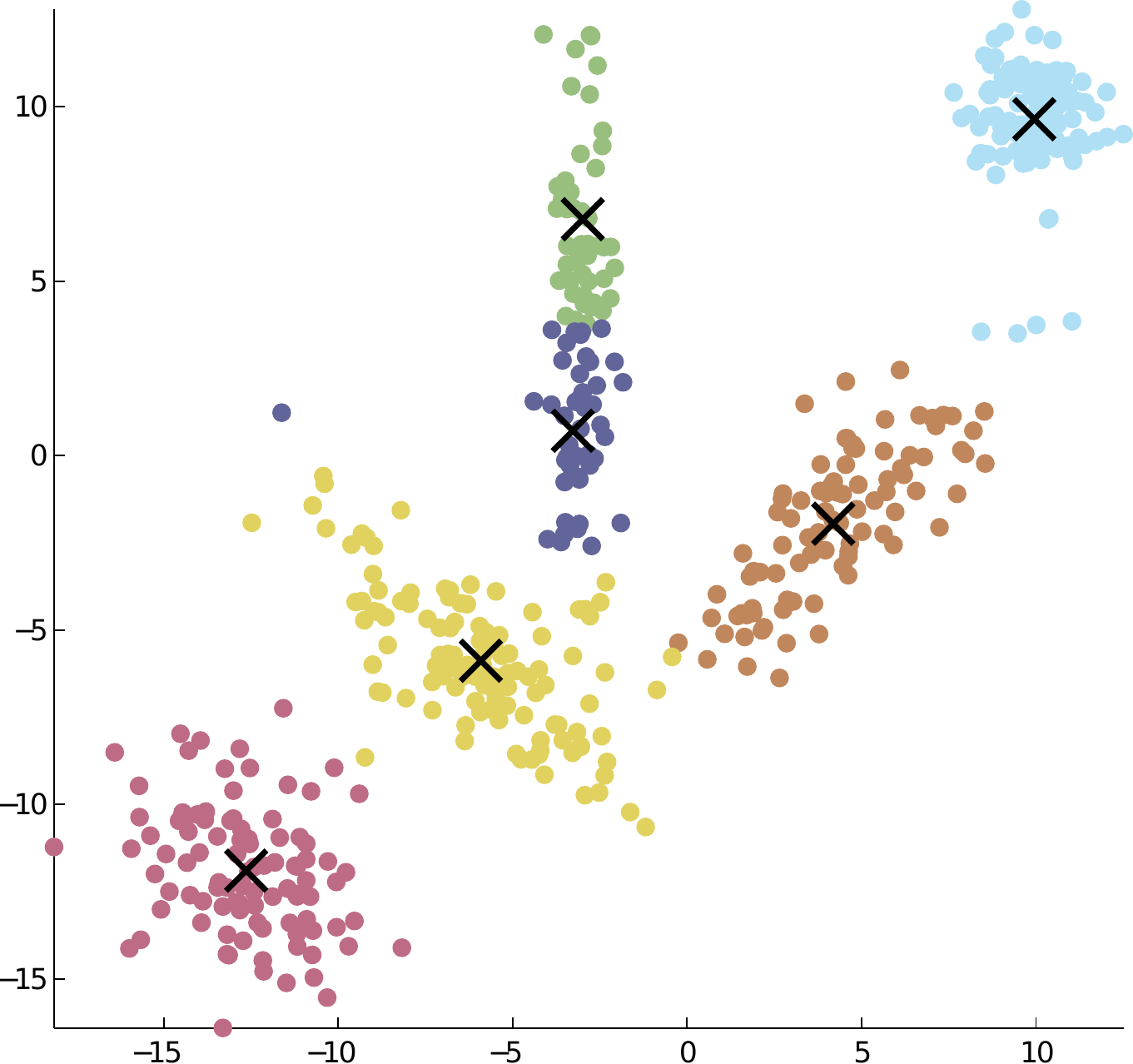} &
		\includegraphics[width=.14\textwidth]{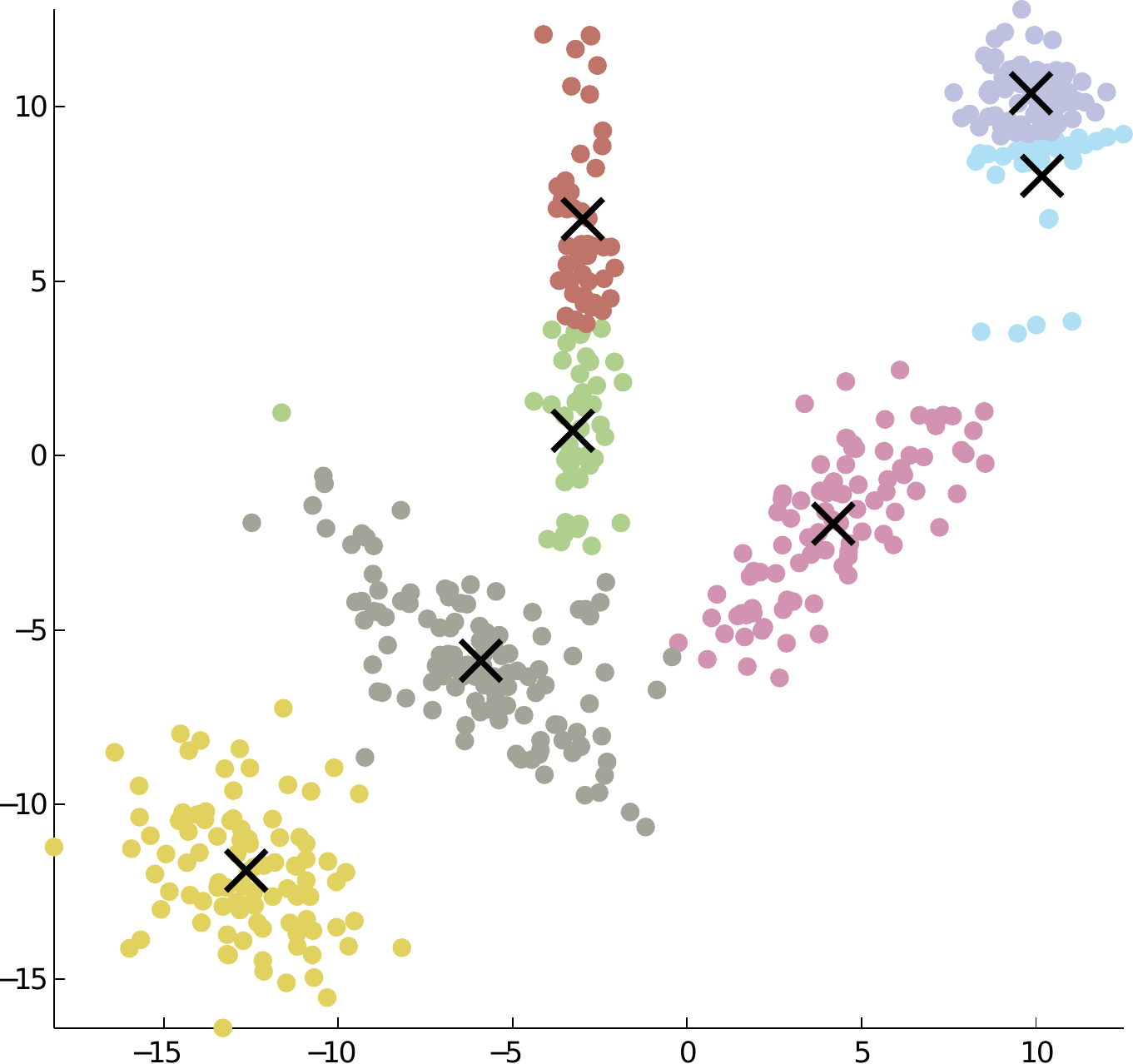} \\

		&
		\multicolumn{1}{c}{\textsc{nmi}=0.727} &
		\multicolumn{1}{c}{\textsc{nmi}=0.852} &
		\multicolumn{1}{c}{\textsc{nmi}=0.976} &
		\multicolumn{1}{c}{\textsc{nmi}=0.939} &
		\multicolumn{1}{c}{\textsc{nmi}=0.904} \\

		&
		\multicolumn{1}{c}{\textsc{f-m}=71.85} &
		\multicolumn{1}{c}{\textsc{f-m}=85.13} &
		\multicolumn{1}{c}{\textsc{f-m}=99.20} &
		\multicolumn{1}{c}{\textsc{f-m}=94.19} &
		\multicolumn{1}{c}{\textsc{f-m}=90.28} \\

	\end{tabular}

	\vskip.8em
	
	\begin{tabular}{cccc}
		Transposed preference matrix $\transpose{\mat{A}}$ &
		Bi-clustered preference matrix \\
		
		\rotatebox{90}{\reflectbox{\includegraphics[width=.2\textwidth]{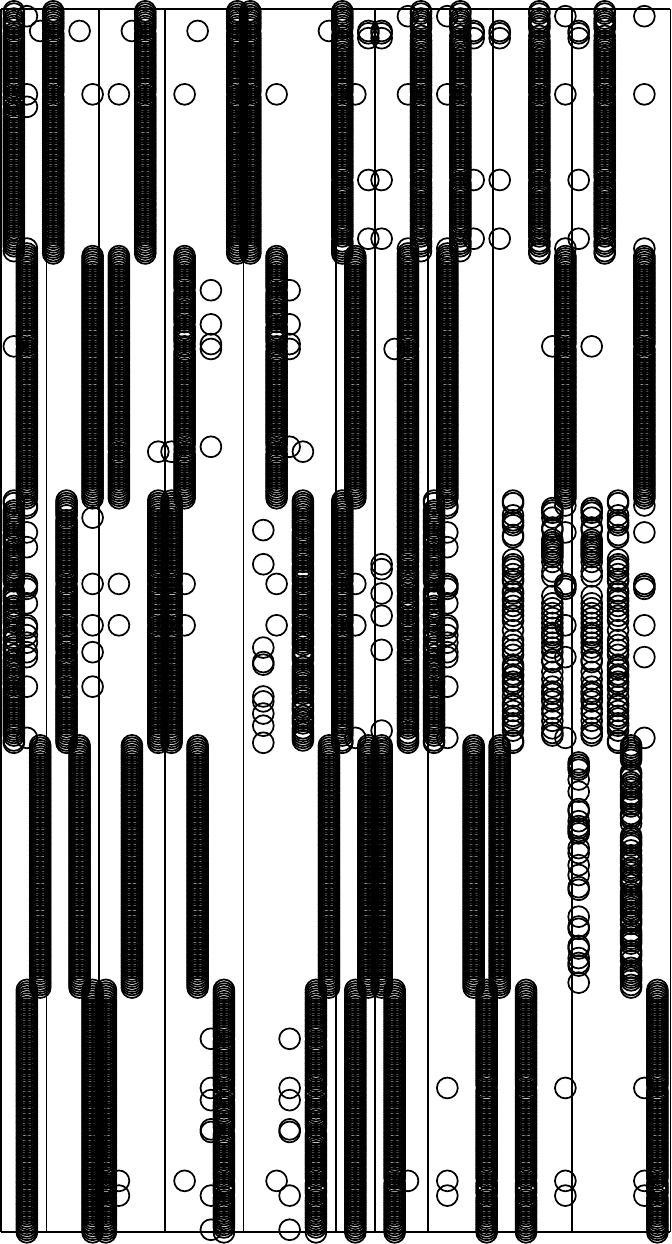}}} &
		\rotatebox{90}{\reflectbox{\includegraphics[width=.2\textwidth]{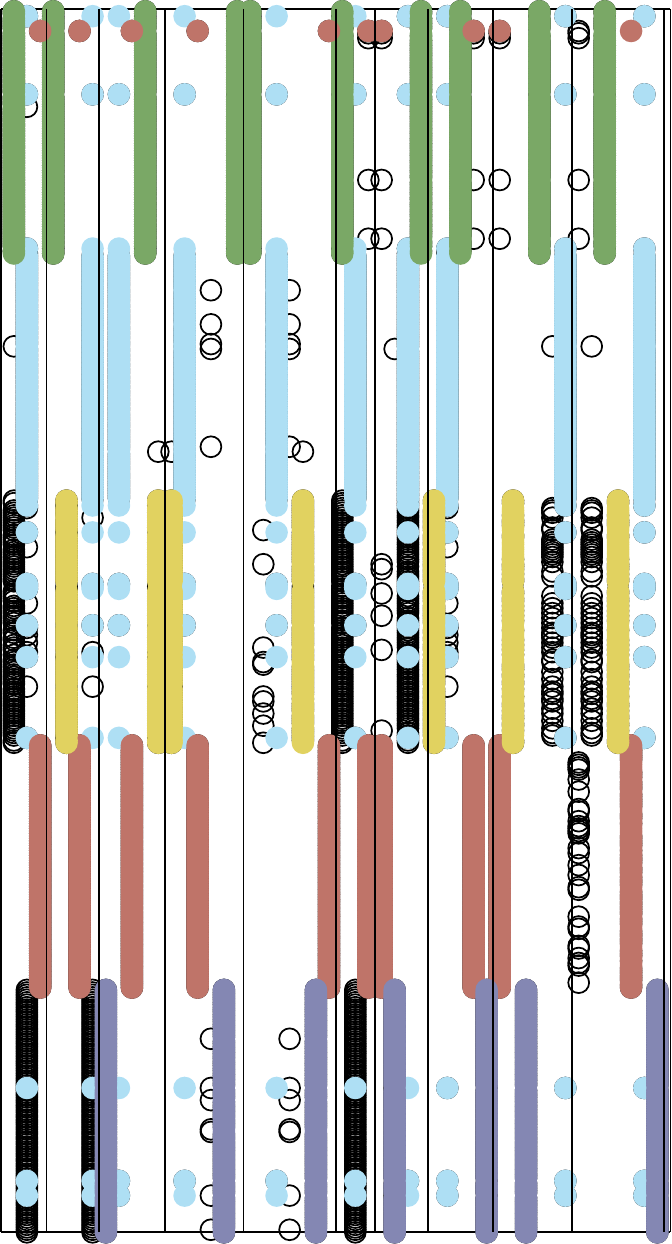}}} \\
	\end{tabular}
	
	\vskip.8em
	
	\begin{tabular}{ccc}
		\multirow{2}{*}{Ground truth} &
		\multicolumn{2}{c}{Consensus (\textsc{nmi}=0.941, \textsc{f-m}=97.80)} \\
		\cmidrule(lr){2-3}
		&
		Point assignments &
		Cluster assignments \\
	
		\includegraphics[width=.24\textwidth]{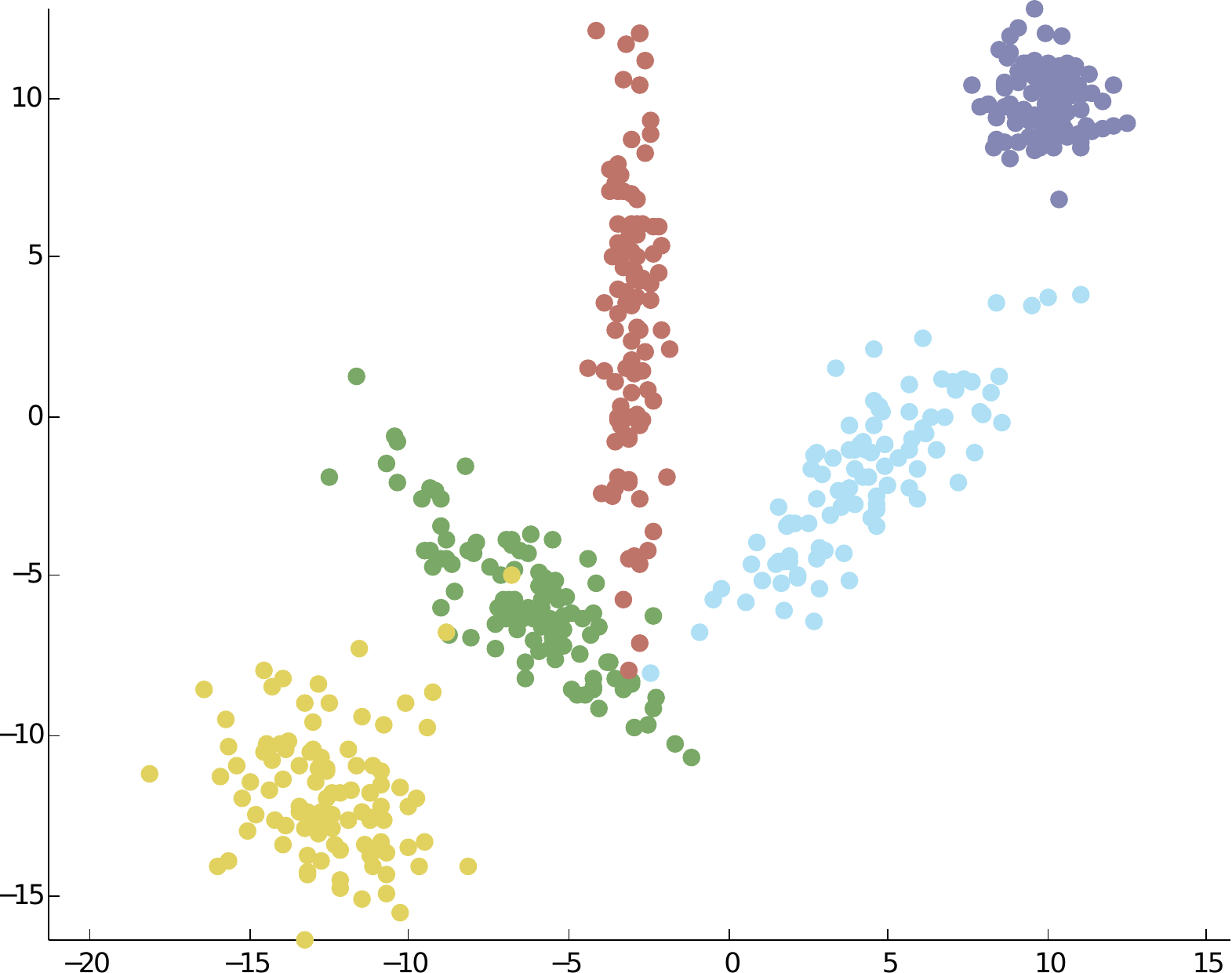} &
		\includegraphics[width=.24\textwidth]{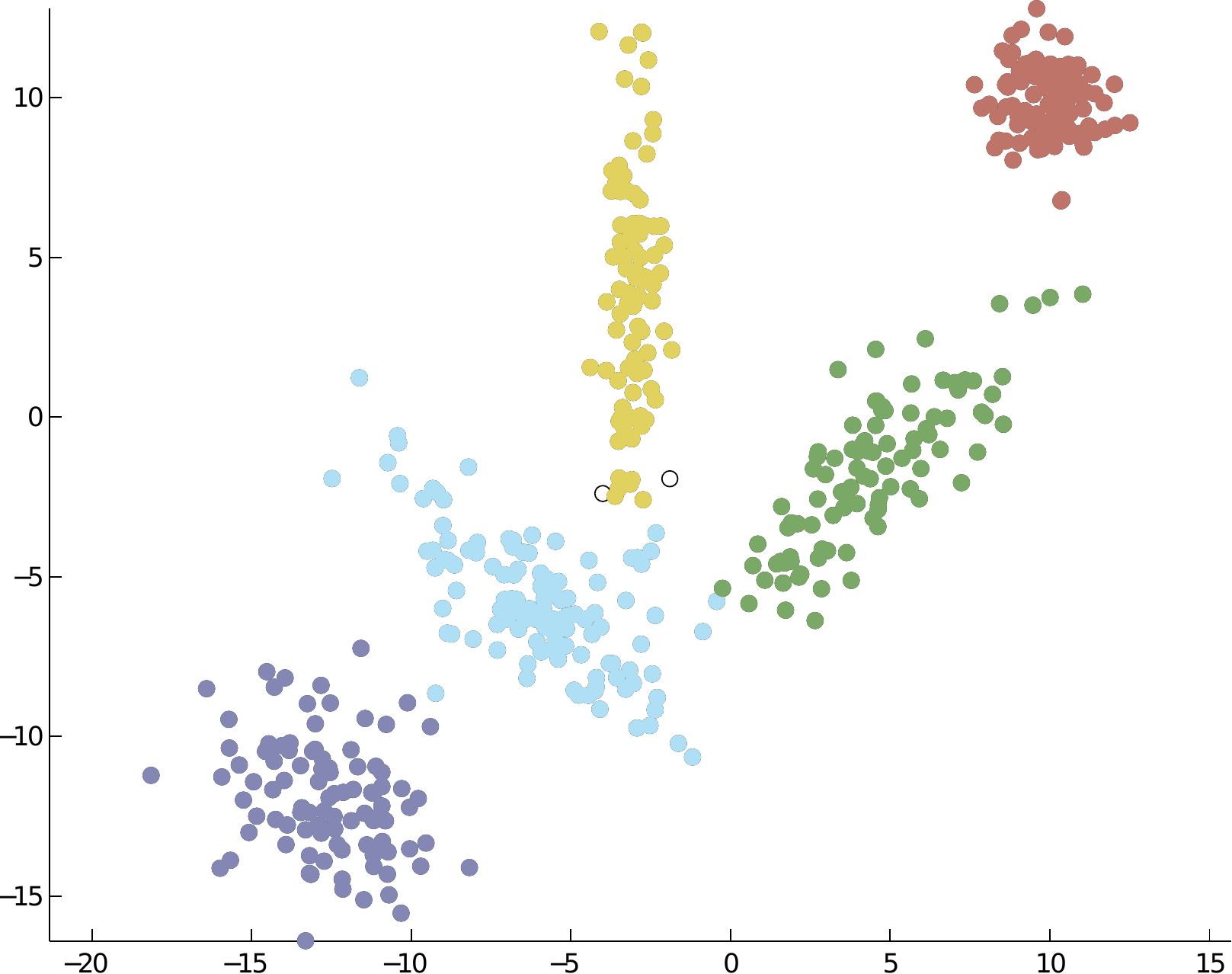} &
		\includegraphics[width=.24\textwidth]{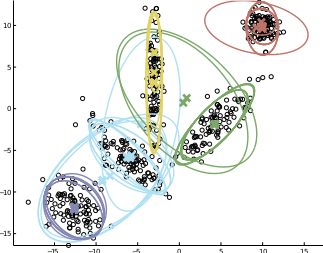} \\
	\end{tabular}	
	\end{small}
	
	\caption{Synthetic example of the proposed bi-clustering consensus computed from different instances of GMM and $K$-means, varying the number of classes. Each bi-cluster corresponds to jointly selected point and cluster assignments (see the bi-clustered preference matrix), we thus show with the same color its points and the covariance matrices of its constituting clusters, represented by ellipses.  NMI and F-M stand for normalized mutual information and F-measure, respectively.}
	\label{fig:clustering_synthetic_gmm}
\end{figure}

\begin{figure}[t]

	\centering
	\begin{small}
	
	Base algorithms\\[6pt]
	\begin{tabular}{ @{\hspace{0pt}} c *{6}{ @{\hspace{2pt}} c} @{\hspace{0pt}} }
	
		\includegraphics[width=.138\textwidth]{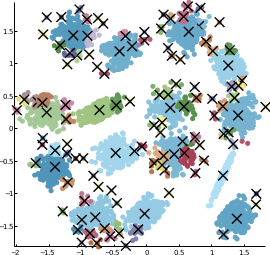} &
		\includegraphics[width=.138\textwidth]{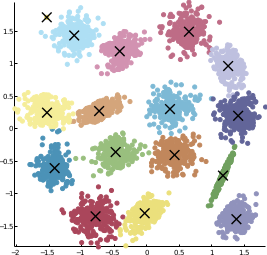} &
		\includegraphics[width=.138\textwidth]{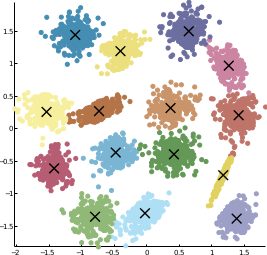} &
		\includegraphics[width=.138\textwidth]{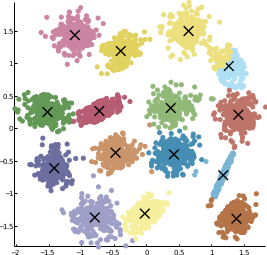} &
		\includegraphics[width=.138\textwidth]{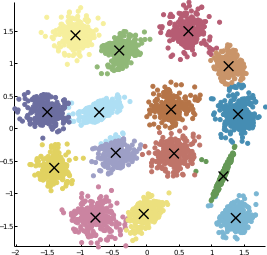} &
		\includegraphics[width=.138\textwidth]{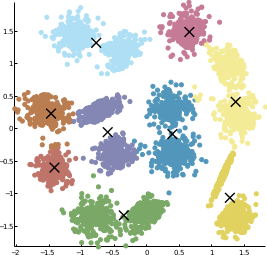} &
		\includegraphics[width=.138\textwidth]{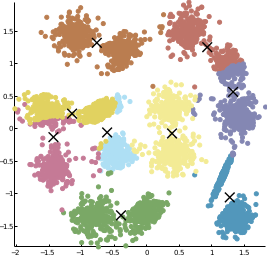} \\

		\textsc{nmi}=0.905 &
		\textsc{nmi}=0.985 &
		\textsc{nmi}=0.983 &
		\textsc{nmi}=0.971 &
		\textsc{nmi}=0.976 &
		\textsc{nmi}=0.872 &
		\textsc{nmi}=0.833 \\

		\textsc{f-m}=93.07 &
		\textsc{f-m}=99.27 &
		\textsc{f-m}=99.08 &
		\textsc{f-m}=97.92 &
		\textsc{f-m}=98.70 &
		\textsc{f-m}=73.01 &
		\textsc{f-m}=70.10 \\
				
	\end{tabular}
		
	\vskip.5em

	\begin{tabular}{ccc}
		\multirow{2}{*}{Ground truth} &
		\multicolumn{2}{c}{Consensus (\textsc{nmi}=0.986, \textsc{f-m}=99.32)} \\
		\cmidrule(lr){2-3}
		&
		Point assignments &
		Cluster assignments \\
	
		\includegraphics[width=.24\textwidth]{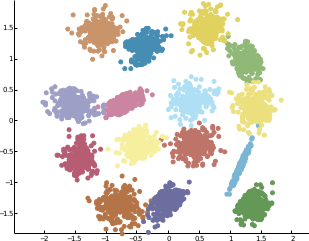} &
		\includegraphics[width=.24\textwidth]{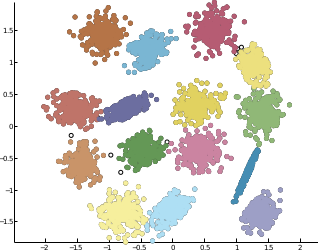} &
		\includegraphics[width=.24\textwidth]{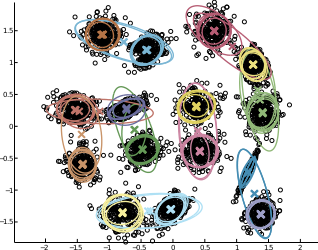} \\
	\end{tabular}
	\end{small}
		
	\caption{Synthetic example of the proposed bi-clustering consensus computed from different mean-shift instances, varying the kernel size. The bi-clusters correspond to jointly selected point and cluster assignments (see the bi-clustered preference matrix), we thus show with the same color its points and the corresponding covariance matrix, represented by an ellipse.  NMI and F-M stand for normalized mutual information and F-measure, respectively.}
	\label{fig:clustering_synthetic_mean-shift}
\end{figure}

Clustering seeks to group observations into subsets (called clusters) so that, in some sense, intra-cluster observations are more similar than inter-cluster ones. It is one of the key components of exploratory data mining and a common technique for statistical data analysis, used in such diverse fields as machine learning, pattern recognition, image analysis, information retrieval, and bioinformatics. See~\cite{jain09} for a broad overview of the field.

Consensus clustering is the most straightforward application of our grouping framework. Each set $\set{C}_k$ of candidate groups in the pool $\{ \set{C}_k \}_{k=1}^c$ is the result of an instance of a clustering algorithm, and the preference matrix $\mat{A}$ is therefore constructed in a straightforward fashion.

\subsection{Experimental results}
In these experiments, we use uniform weights  in $\mat{A}$. For assessing the quality of a solution when ground truth is available, we use the standard normalized mutual information (NMI) and F-measure to evaluate the results.
Let $\set{G}, \set{C}$ be the ground truth and the evaluated solution, respectively.
The normalized mutual information is defined as
\begin{equation}
	\operatorname{NMI}(\set{G}, \set{C}) = 2 \frac{ \operatorname{H}(\set{G}) + \operatorname{H}(\set{C}) - \operatorname{H}(\set{G}, \set{C}) }{ \operatorname{H}(\set{G}) + \operatorname{H}(\set{C}) } ,
\end{equation}
where $\operatorname{H}(\set{G}), \operatorname{H}(\set{C})$ are marginal entropies, and $\operatorname{H}(\set{G}, \set{C})$ is the joint entropy of $\set{G}$ and $\set{C}$.
The F-measure is defined as
\begin{equation}
	\operatorname{F_\beta}(\set{G}, \set{C}) = \frac{ (\beta^2 + 1) \cdot \operatorname{P}_{\set{G}} (\set{C}) \cdot \operatorname{R}_{\set{G}} (\set{C}) }{ \beta^2 \cdot \operatorname{P}_{\set{G}} (\set{C}) + \operatorname{R}_{\set{G}}(\set{C}) } ,
\end{equation}
where $\operatorname{P}_{\set{G}} (\set{C}), \operatorname{R}_{\set{G}} (\set{C})$ are the precision and recall rates of $\set{C}$ with respect to $\set{G}$. In these experiments we use $\beta=1$, and express the F-measure as a percentage.

In figures~\ref{fig:clustering_synthetic_gmm} and~\ref{fig:clustering_synthetic_mean-shift} we observe two synthetic examples, where the datasets are different mixtures of Gaussians. In Figure~\ref{fig:clustering_synthetic_gmm} we run several instances of a Gaussian mixture model (GMM) clustering algorithm and $K$-means, where each instance has a different number of pre-specified clusters. In Figure~\ref{fig:clustering_synthetic_mean-shift} we run several instances of the mean-shift algorithm, each one with a different kernel size. We show the bi-clusters obtained with the proposed consensus algorithm. In both cases and without tuning any parameters, the quality of the solution approximates quite well the ground truth both qualitatively (by visual inspection) and quantitatively. In Figure~\ref{fig:clustering_synthetic_gmm}, the consensus solution ranks second best when comparing with the base algorithms in terms of NMI and F-measure; in Figure~\ref{fig:clustering_synthetic_mean-shift}, it ranks first.

\begin{figure}
	\begin{subfigure}[b]{.5\textwidth}
		\centering
		$\mat{X}$ factor\\[2pt]
		\includegraphics[width=\textwidth]{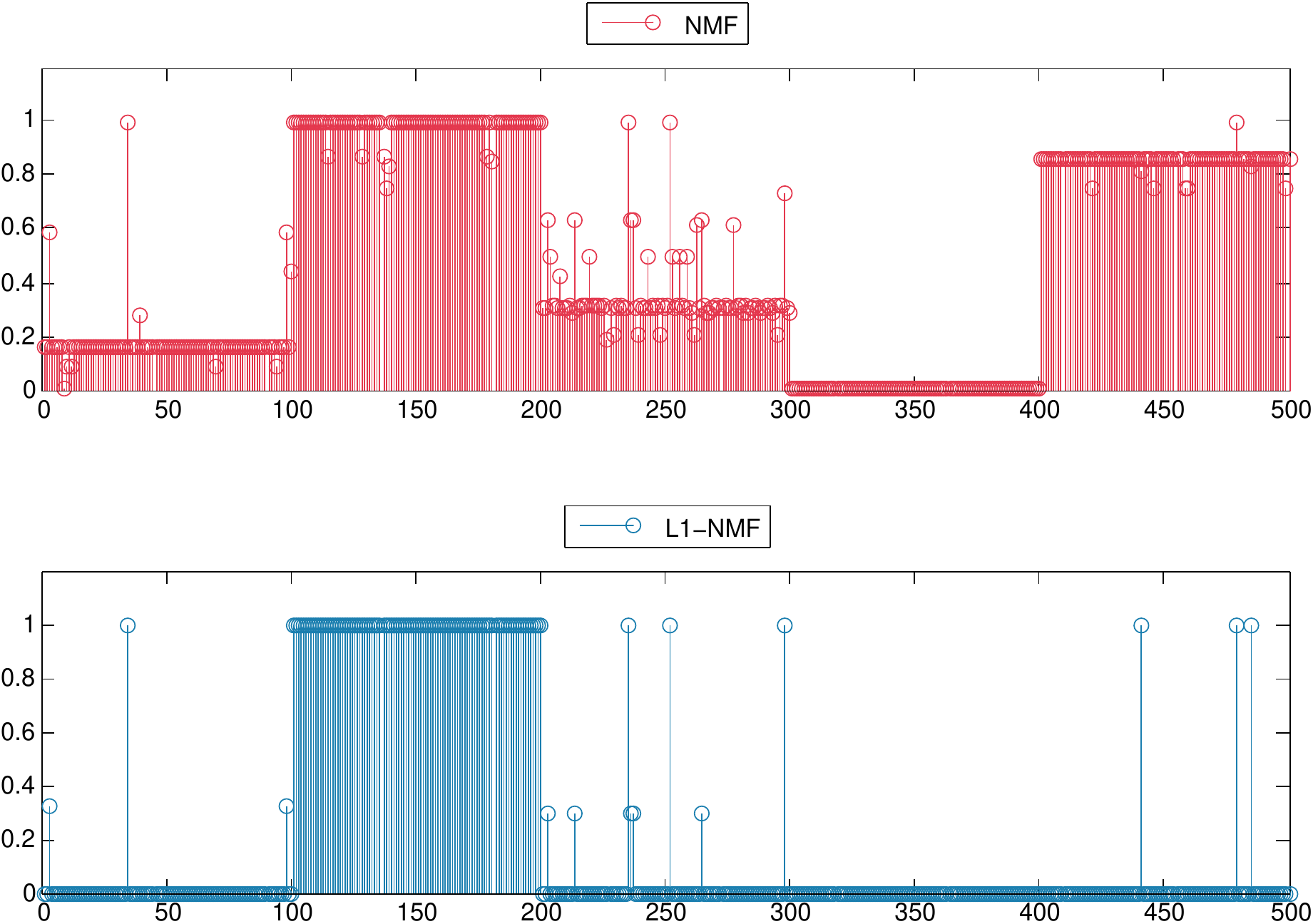}
	\end{subfigure}
	\begin{subfigure}[b]{.5\textwidth}
		\centering
		$\mat{Y}$ factor\\[2pt]
		\includegraphics[width=\textwidth]{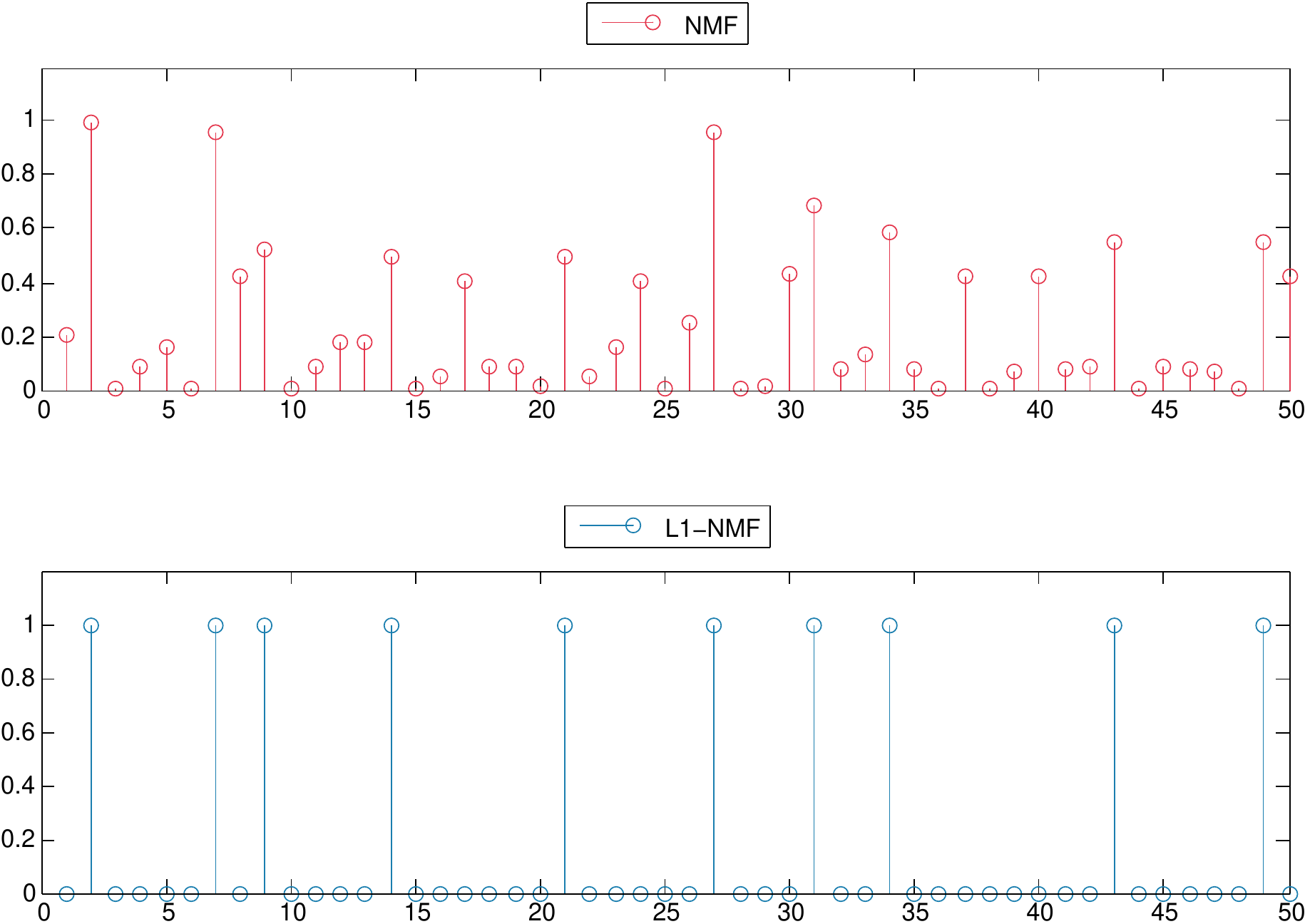}
	\end{subfigure}
	
	\caption{Bi-cluster extraction using NMF (problem~(\ref{eq:classical_nmf})) versus L1-NMF (problem~(\ref{eq:nmf})) in Algorithm~\ref{algo:biclustering}. We show the active set of the first extracted bi-cluster in the example of Figure~\ref{fig:clustering_synthetic_gmm} (depicted there in light blue). Sparsely fitting the preference matrix, by using the L1-norm, helps to detect a bi-cluster that corresponds to a single ground-truth group, without tuning active-set thresholds.}
	\label{fig:clustering_synthetic_gmm_frob_vs_l1}
\end{figure}

It is interesting to visualize the differences between the more classical NMF, see problem~(\ref{eq:classical_nmf}), and L1-NMF,  see problem~(\ref{eq:nmf}). In Figure~\ref{fig:clustering_synthetic_gmm_frob_vs_l1} we compare the active-sets of the rank-one factors $\mat{X}, \mat{Y}$ obtained with both formulations at the first iteration of Algorithm~\ref{algo:biclustering} for the example of Figure~\ref{fig:clustering_synthetic_gmm}. NMF approximates in average the preference matrix as a whole; thus, the active-set of the analyzed bi-cluster merges information from almost all the candidate groups (i.e., its $\mat{Y}$ factor is non-sparse), trying to approximate more than a single group (i.e., its $\mat{X}$ factor is non-sparse). Contrarily, the proposed L1-NMF formulation robustly fits a subset of the preference matrix entries; it thus selects a sparse number of candidate groups and closely fits a single ground-truth group.

\begin{figure}
	
	\begin{subfigure}[b]{\textwidth}
		\centering
		
		\begin{minipage}{.8\textwidth}
			\centering
			\begin{small}
			Base algorithms\\
			\centerline{
				\shortstack{
					\includegraphics[width=.24\textwidth]{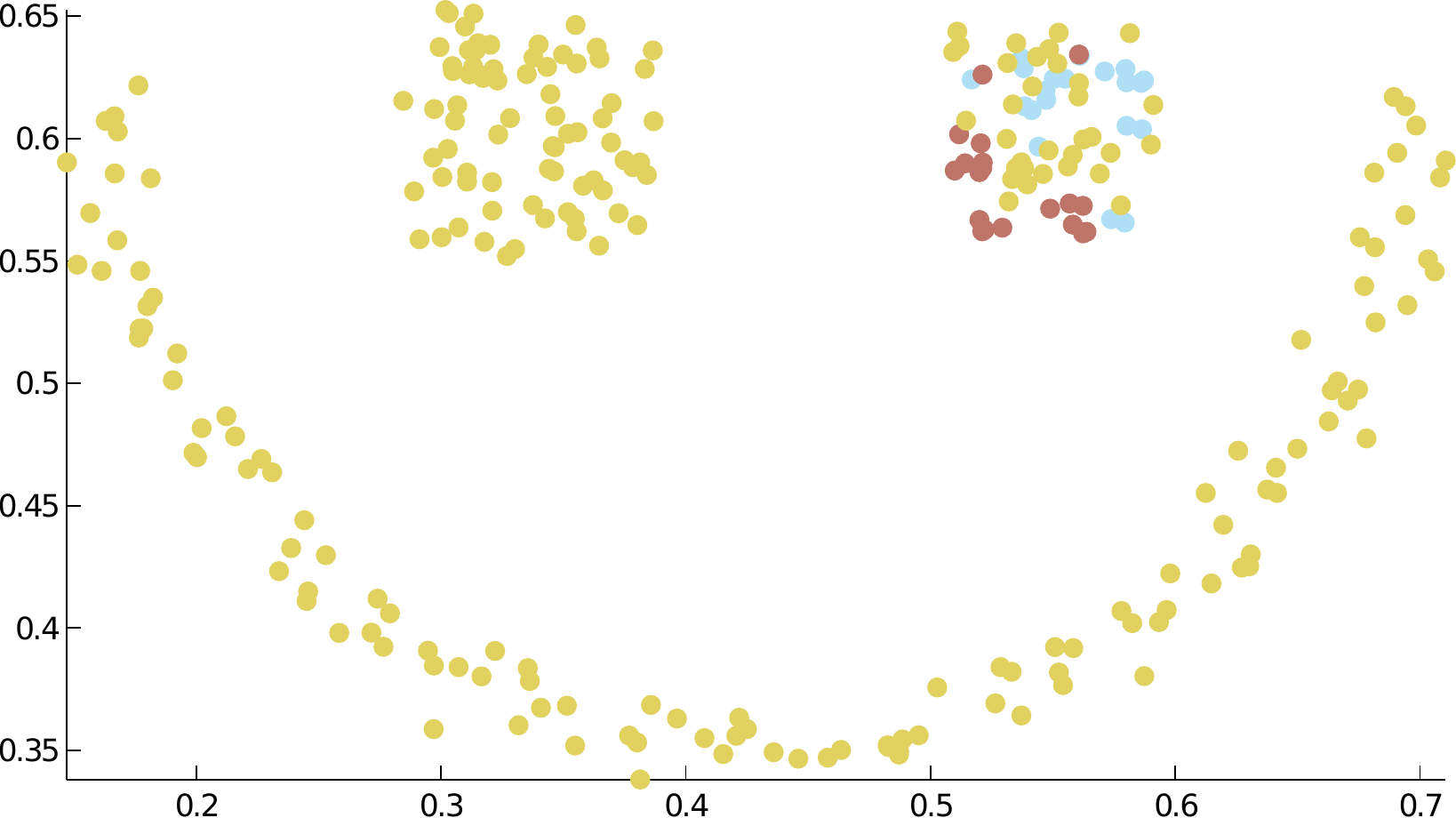}\\
					\textsc{nmi}=0.293\\
					\textsc{f-m}=56.75
				}
				\hfill
				\shortstack{
					\includegraphics[width=.24\textwidth]{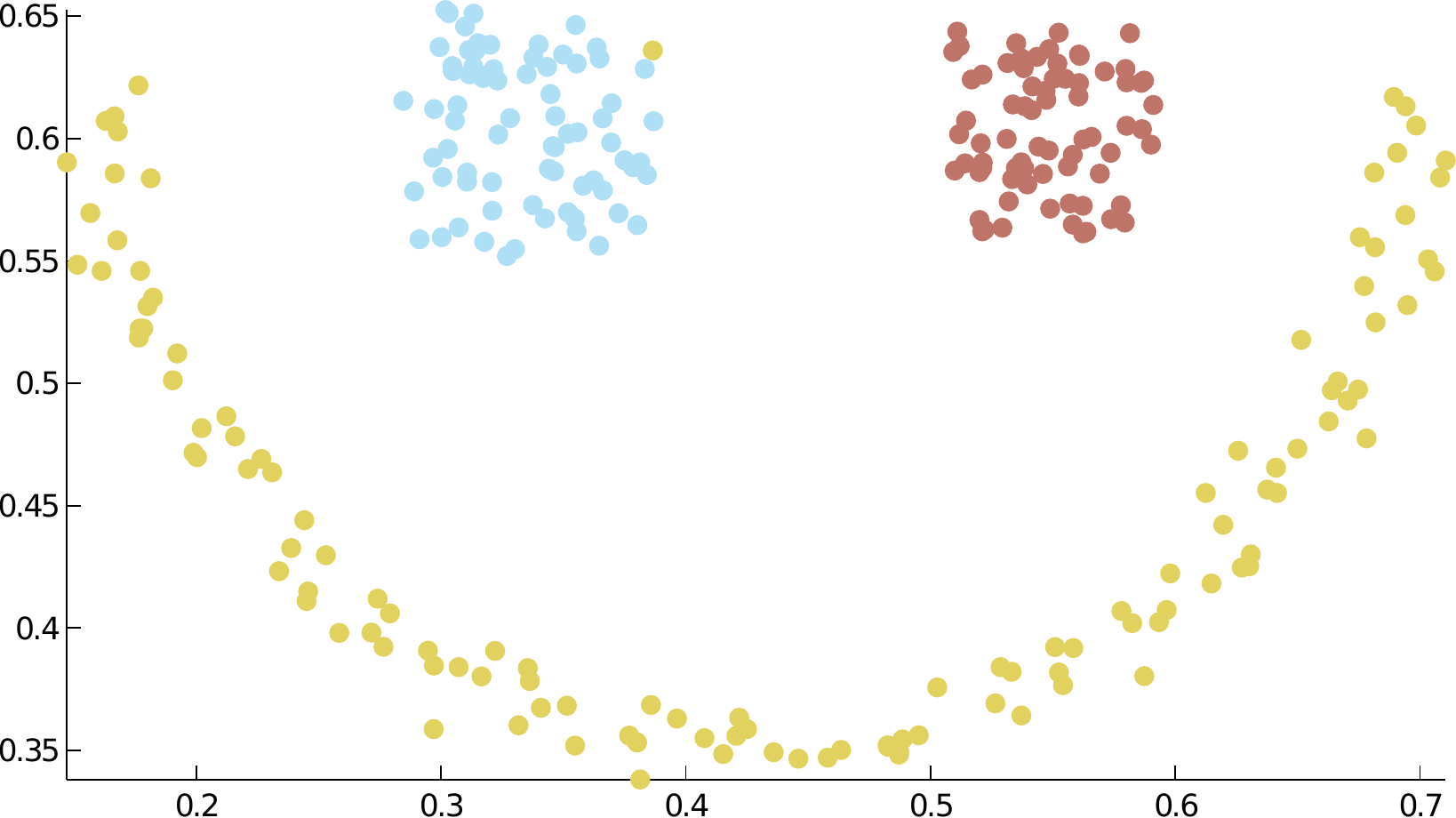}\\
					\textsc{nmi}=0.981\\
					\textsc{f-m}=99.62
				}
				\hfill
				\shortstack{
					\includegraphics[width=.24\textwidth]{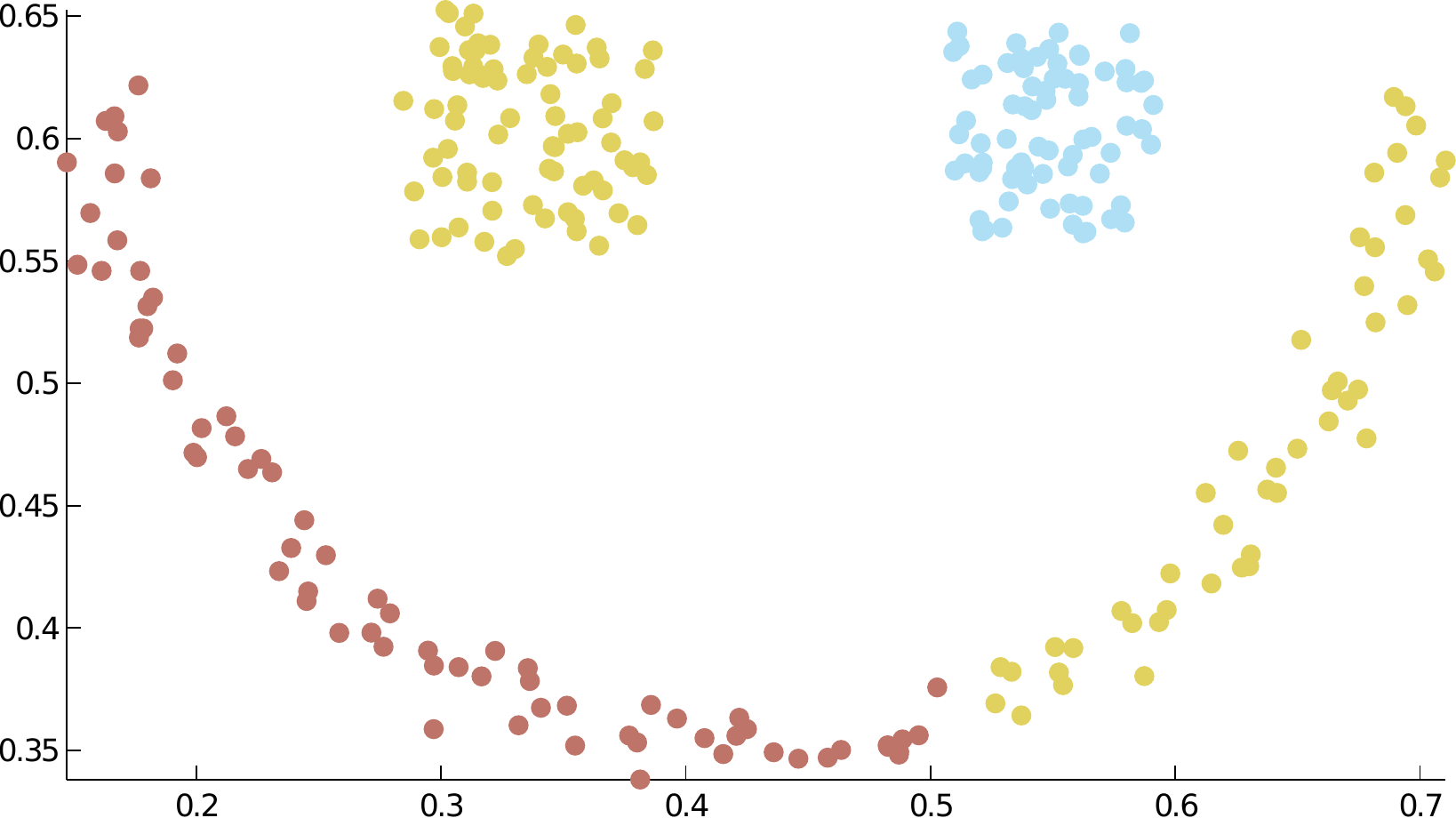}\\
					\textsc{nmi}=0.717\\
					\textsc{f-m}=82.24
				}
				\hfill
				\shortstack{
					\includegraphics[width=.24\textwidth]{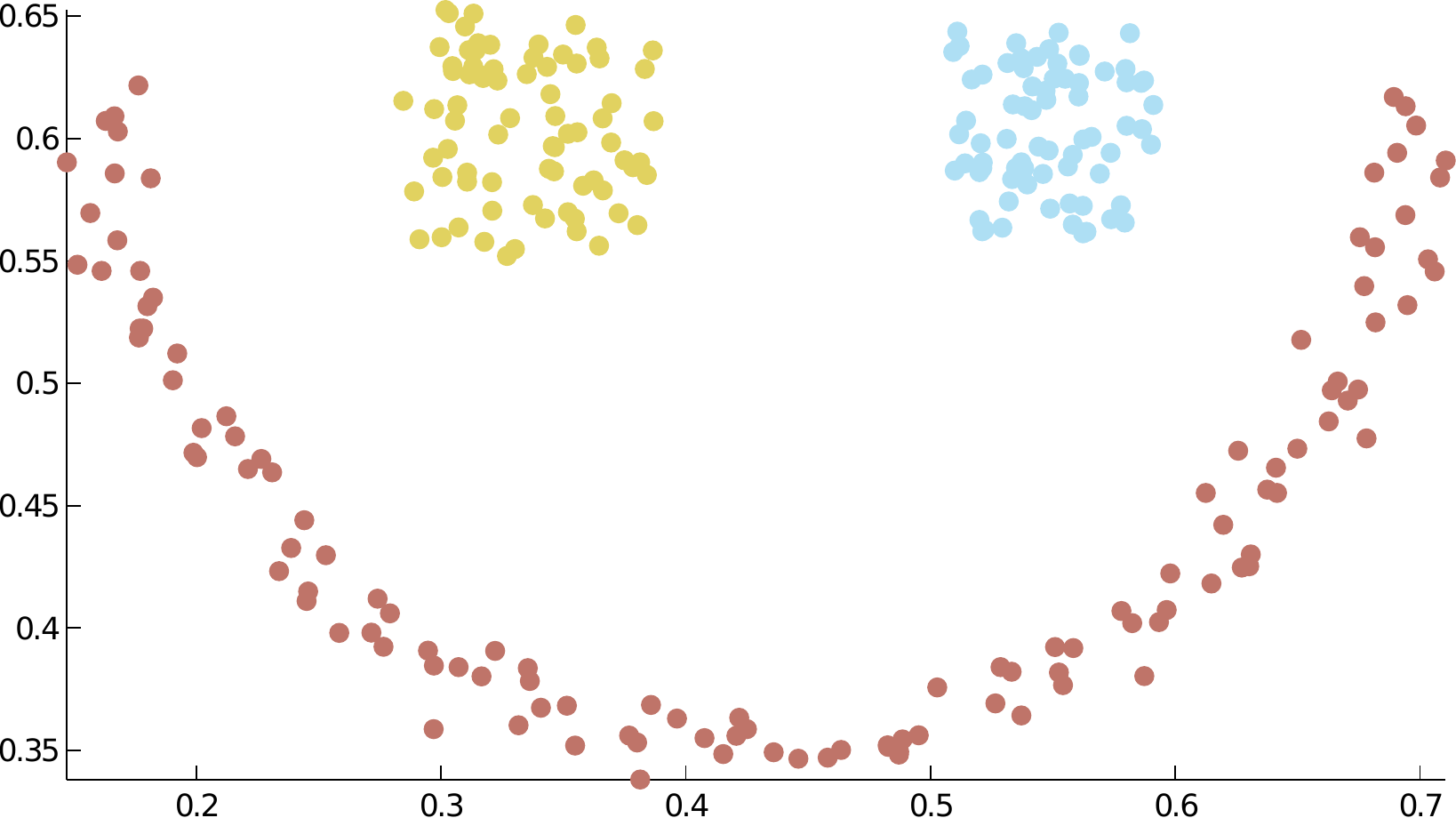}\\
					\textsc{nmi}=1.000 \\
					\textsc{f-m}=100.00
				}
			}
			
			\vskip.5em
			
			\centerline{	
				\shortstack{				
					\includegraphics[width=.24\textwidth]{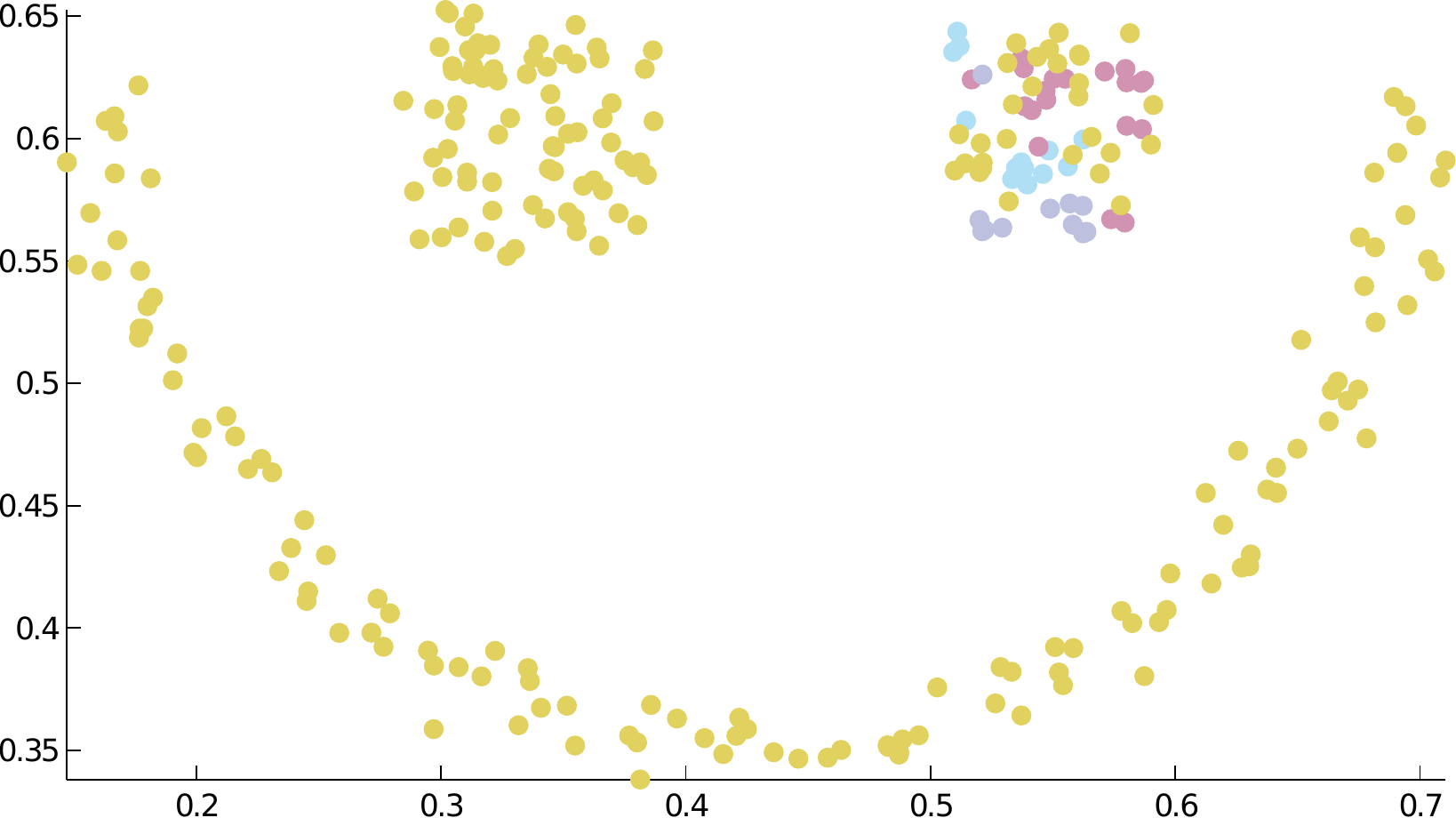}\\
				\textsc{nmi}=0.311\\
				\textsc{f-m}=56.84
				}
				\shortstack{
					\includegraphics[width=.24\textwidth]{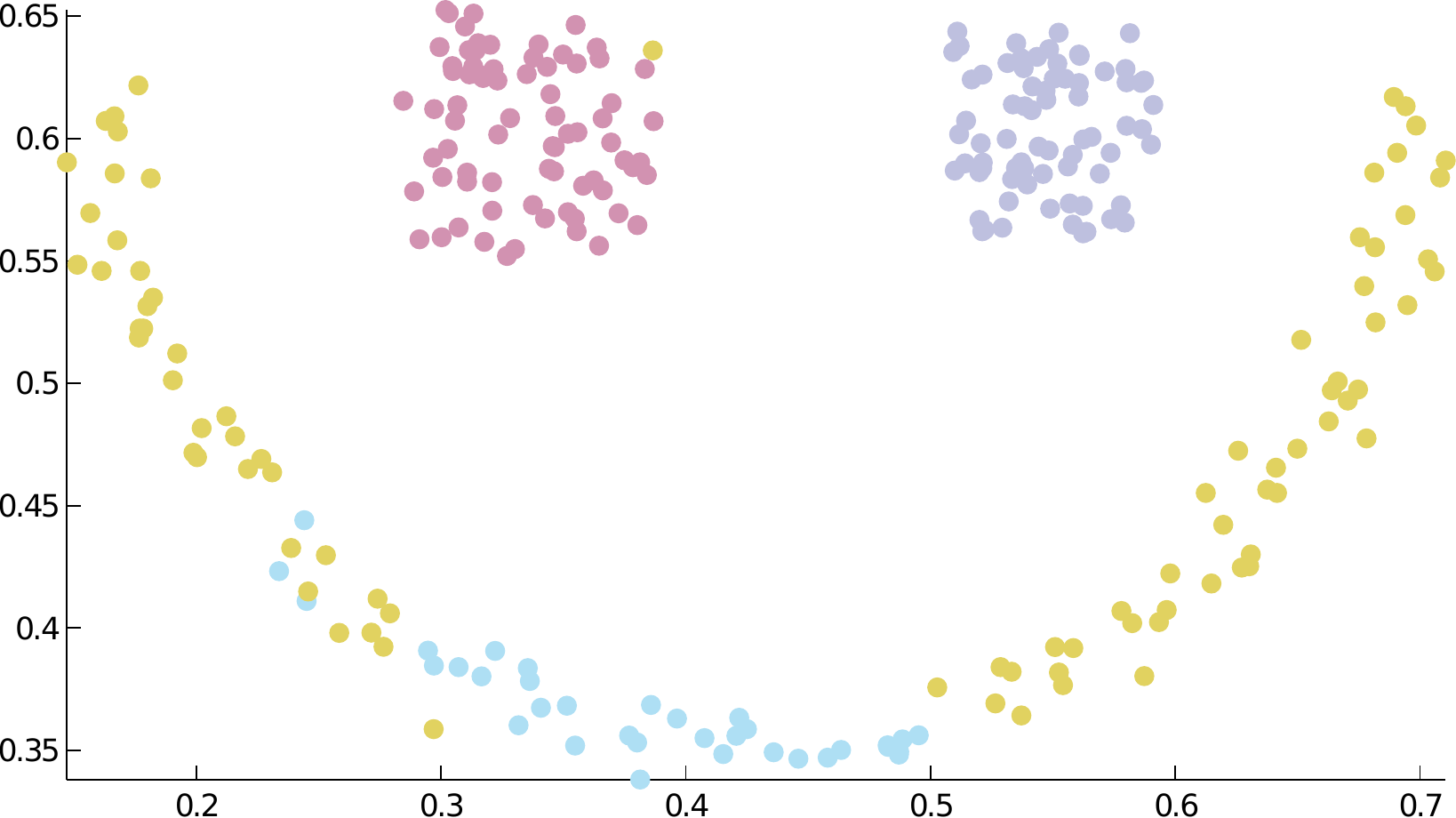}\\
					\textsc{nmi}=0.873\\
					\textsc{f-m}=92.16
				}
				\shortstack{
					\includegraphics[width=.24\textwidth]{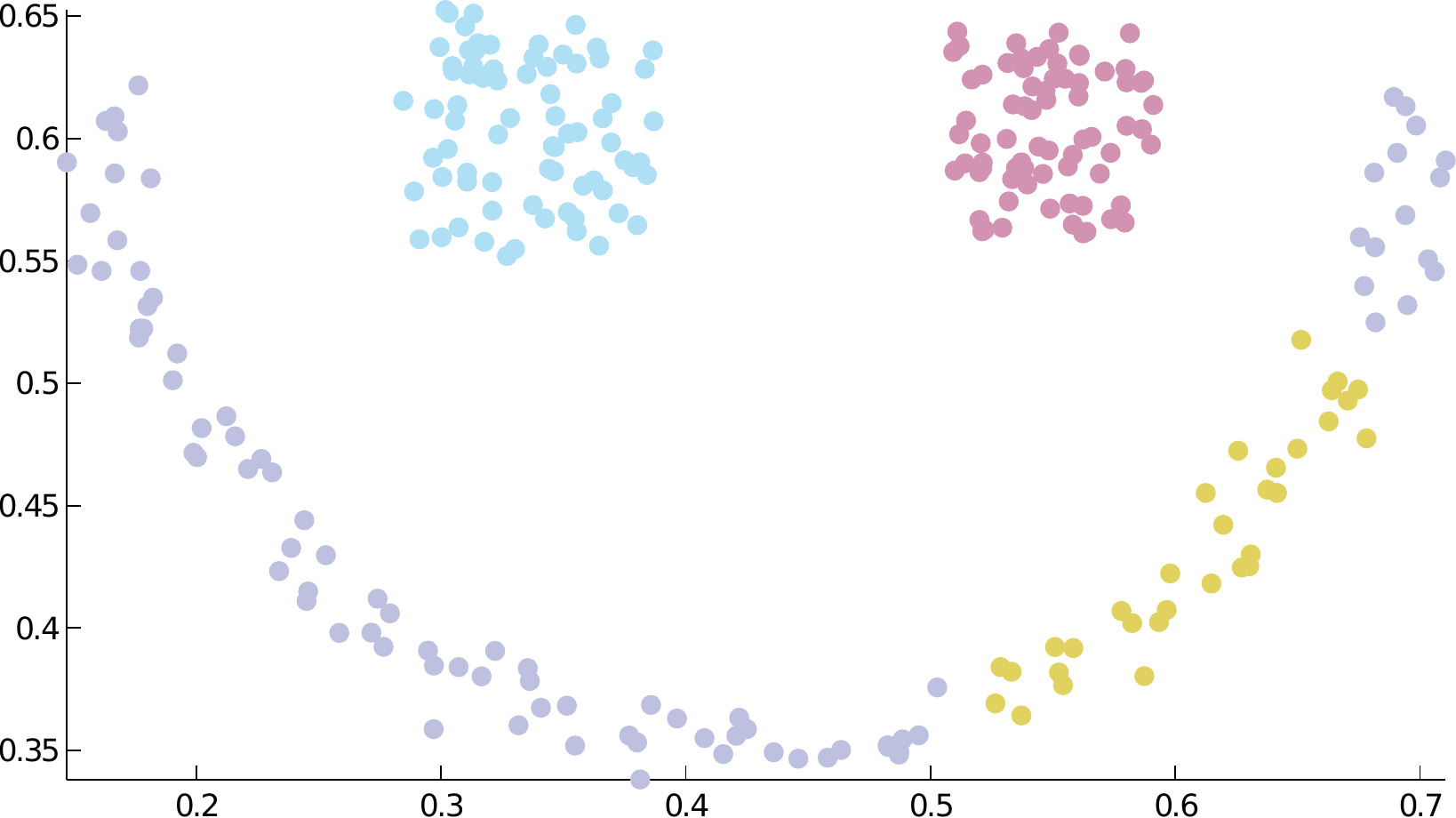}\\
					\textsc{nmi}=0.892\\
					\textsc{f-m}=93.04
				}
				\shortstack{
					\includegraphics[width=.24\textwidth]{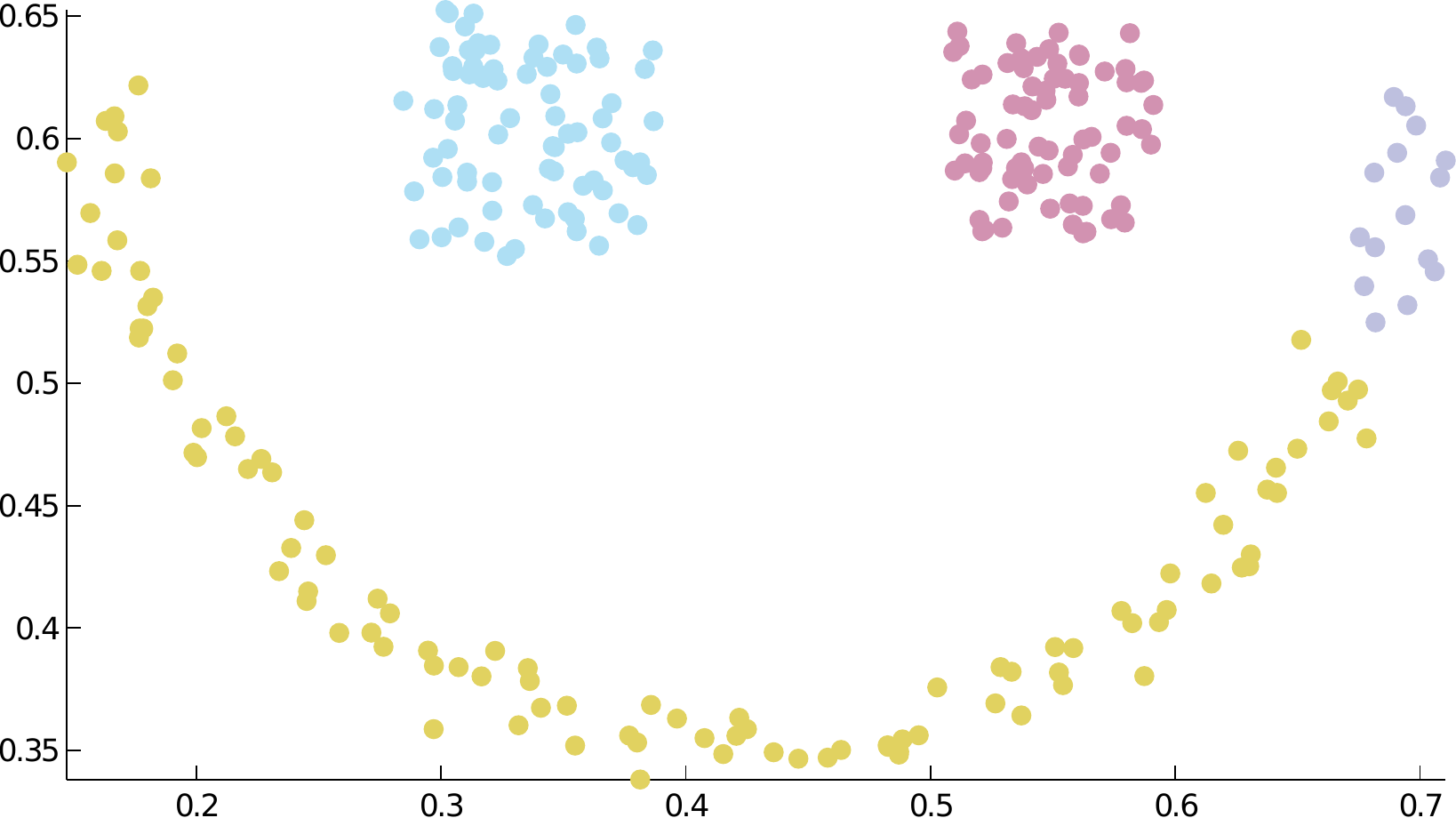} \\
					\textsc{nmi}=0.927\\
					\textsc{f-m}=96.99
				}
			}
			\end{small}
		\end{minipage}
		\hfill
		\begin{minipage}{.19\textwidth}
			\centering
			\begin{small}
			Co-occurrence\\matrix\\[4pt]
			\fbox{\includegraphics[width=\textwidth]{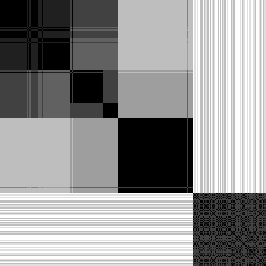}}
			\end{small}
		\end{minipage}
		
		\caption{The base algorithms are different spectral clustering instances, obtained by varying the kernel weighting $\sigma \in \{ 0.005, 0.01, 0.015, 0.02 \}$ for producing the similarity matrix (from left to right), and the desired number of classes $K \in \{ 3, 4 \}$ (from top to bottom).}
		\label{fig:smiley_base}
	\end{subfigure}

	\begin{subfigure}[b]{\textwidth}
		\centerline{
			\hfill
			\shortstack{
				\includegraphics[width=.2\textwidth]{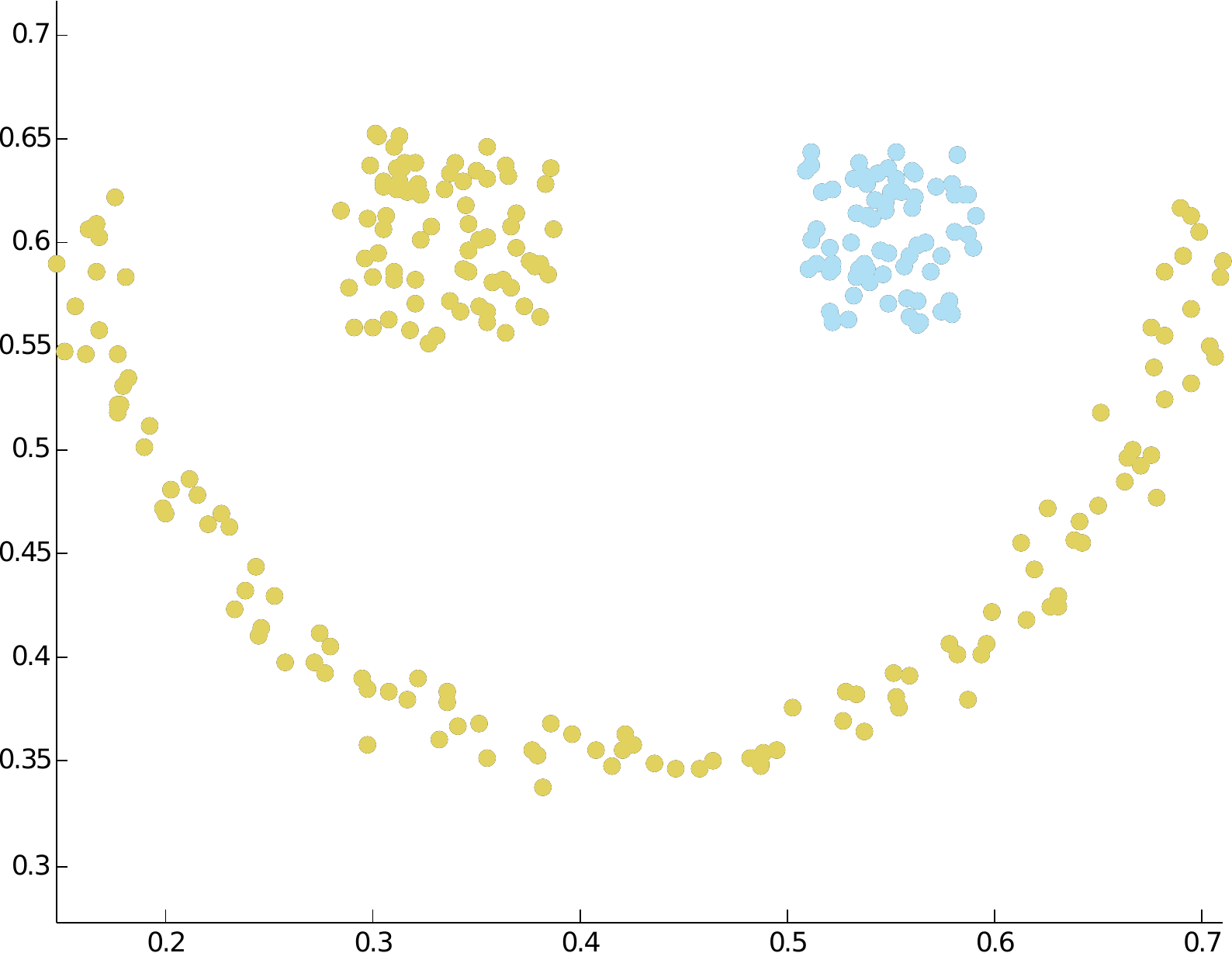}
				\fbox{\includegraphics[width=.12\textwidth]{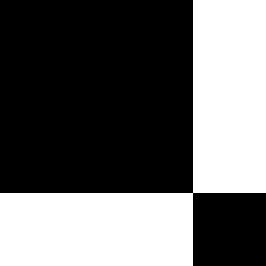}}\\
				\textsc{nmi}=0.708\\
				\textsc{f-m}=76.89
			}
			\hfill
			\shortstack{
				\includegraphics[width=.2\textwidth]{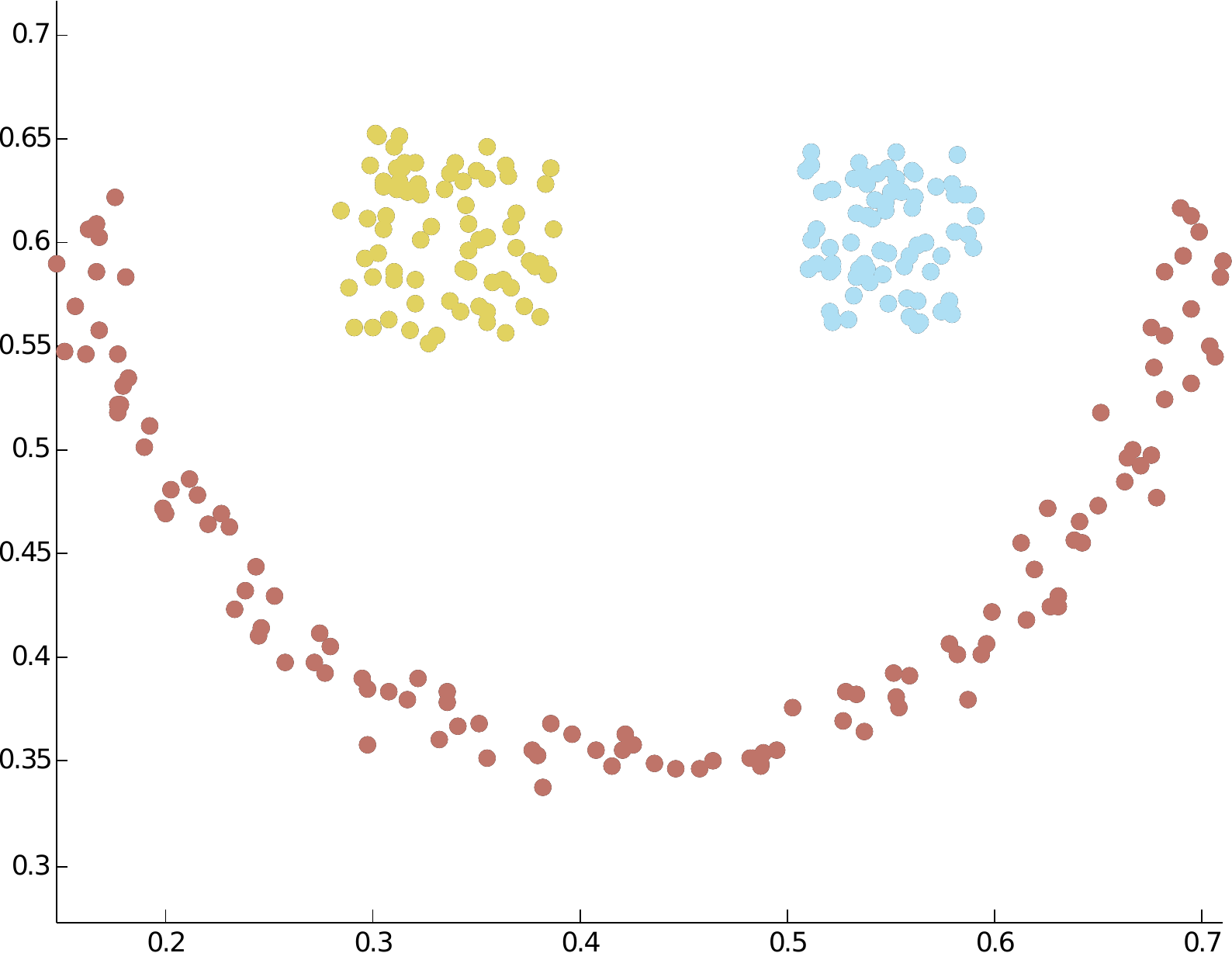}
				\fbox{\includegraphics[width=.12\textwidth]{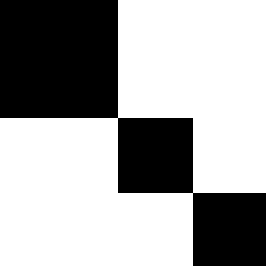}}\\
				\textsc{nmi}=1.000\\
				\textsc{f-m}=100.00				
			}
			\hfill
			\shortstack{
				\includegraphics[width=.2\textwidth]{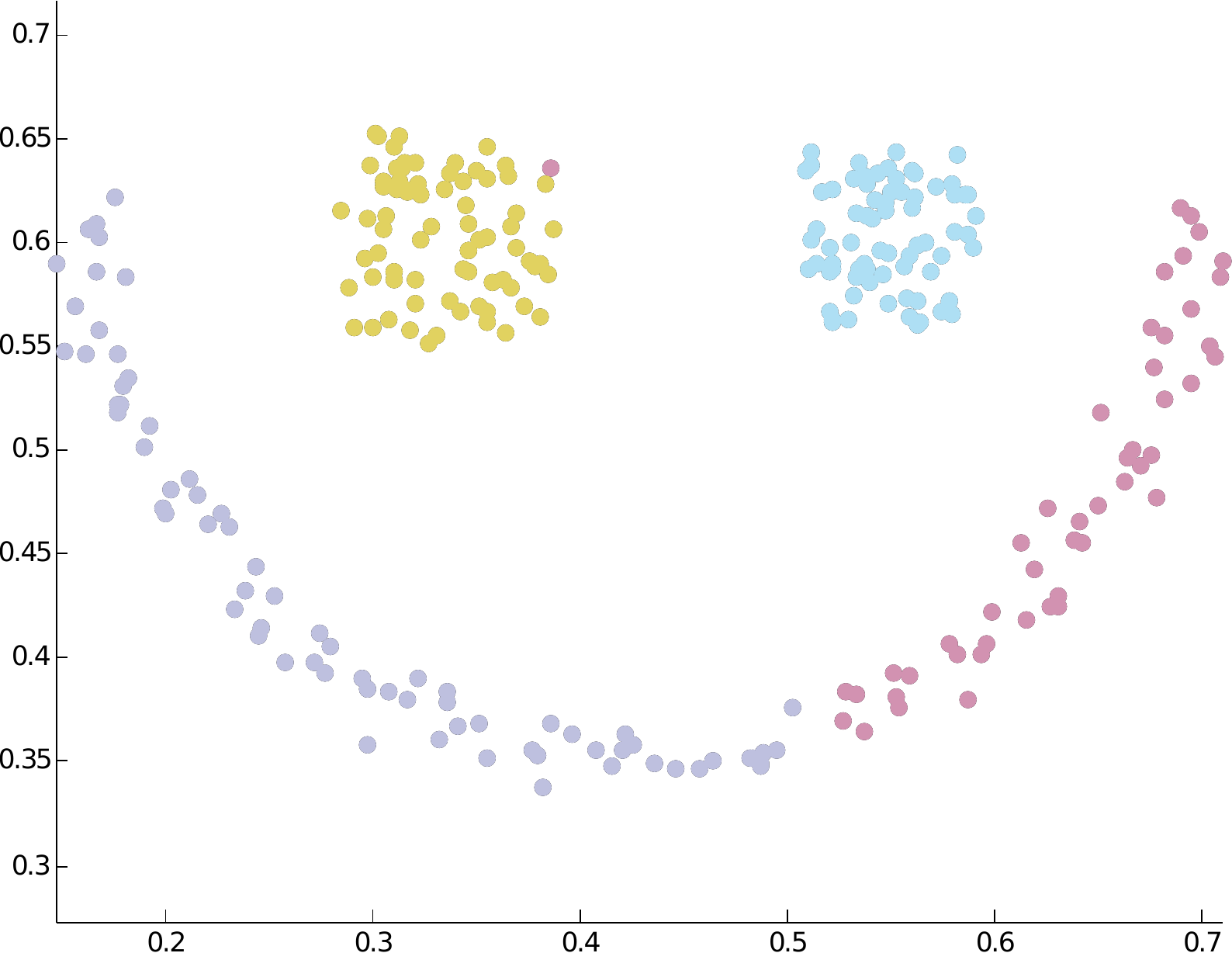}
				\fbox{\includegraphics[width=.12\textwidth]{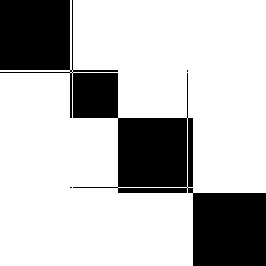}}\\
				\textsc{nmi}=0.862\\
				\textsc{f-m}=88.78
			}
			\hfill
		}
				
		\caption{Spectral clustering~\cite{ng01} (from left to right, $K = 2, 3, 4$)}
		\label{fig:smiley_spectral}
	\end{subfigure}
	
	\begin{subfigure}[b]{.325\textwidth}
		\centering
		\shortstack{
			\includegraphics[width=.6\textwidth]{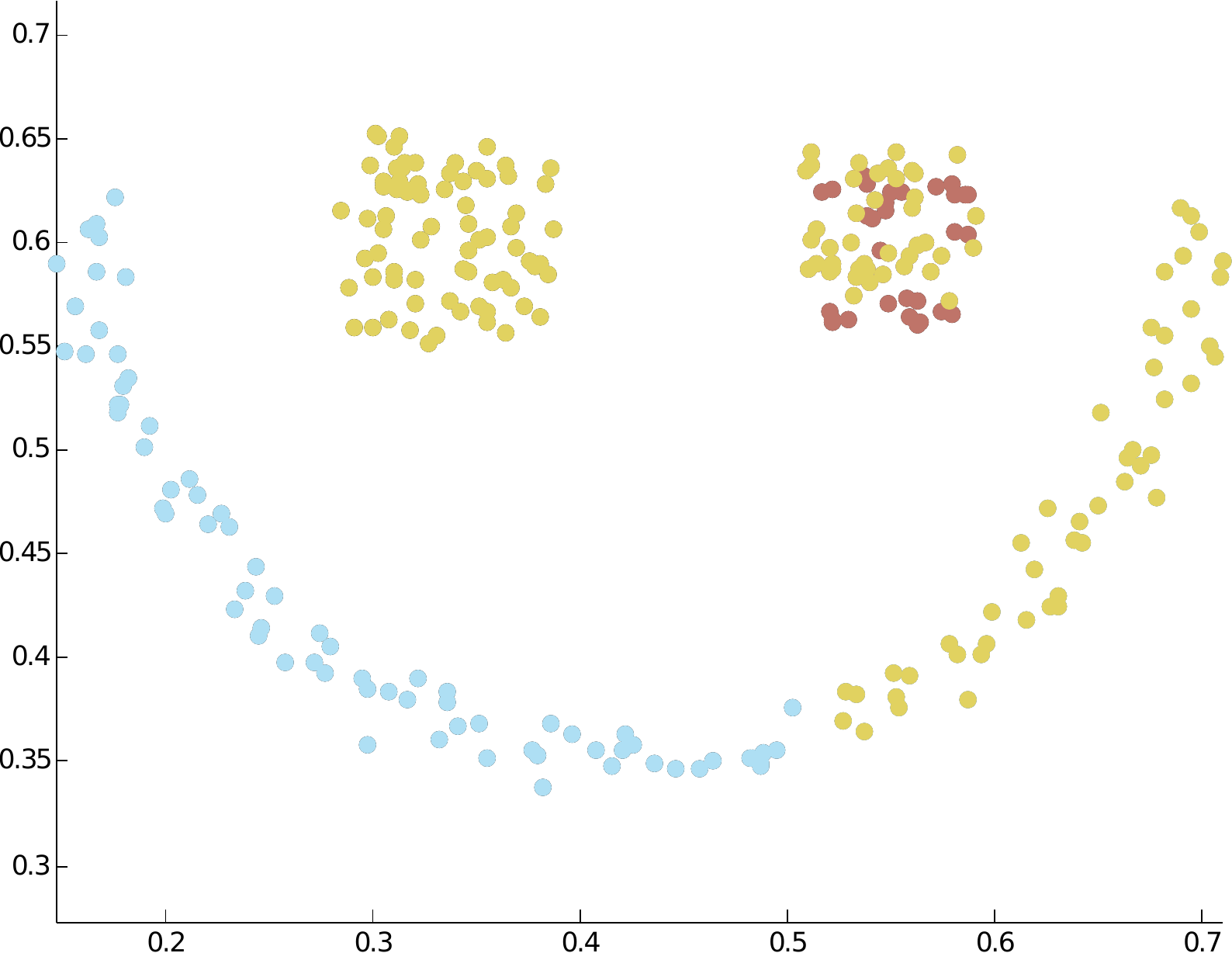}
			\fbox{\includegraphics[width=.36\textwidth]{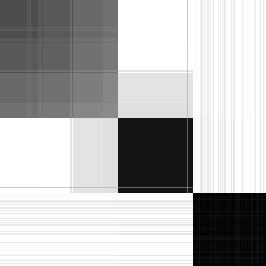}}\\
			\textsc{nmi}=0.708\\
			\textsc{f-m}=76.89
		}
		
		\caption{SNMF~\cite{li2007} ($q=3$)}
		\label{fig:smiley_snmf}
	\end{subfigure}
	\hfill
	\begin{subfigure}[b]{.325\textwidth}
		\centering
		\shortstack{
			\includegraphics[width=.6\textwidth]{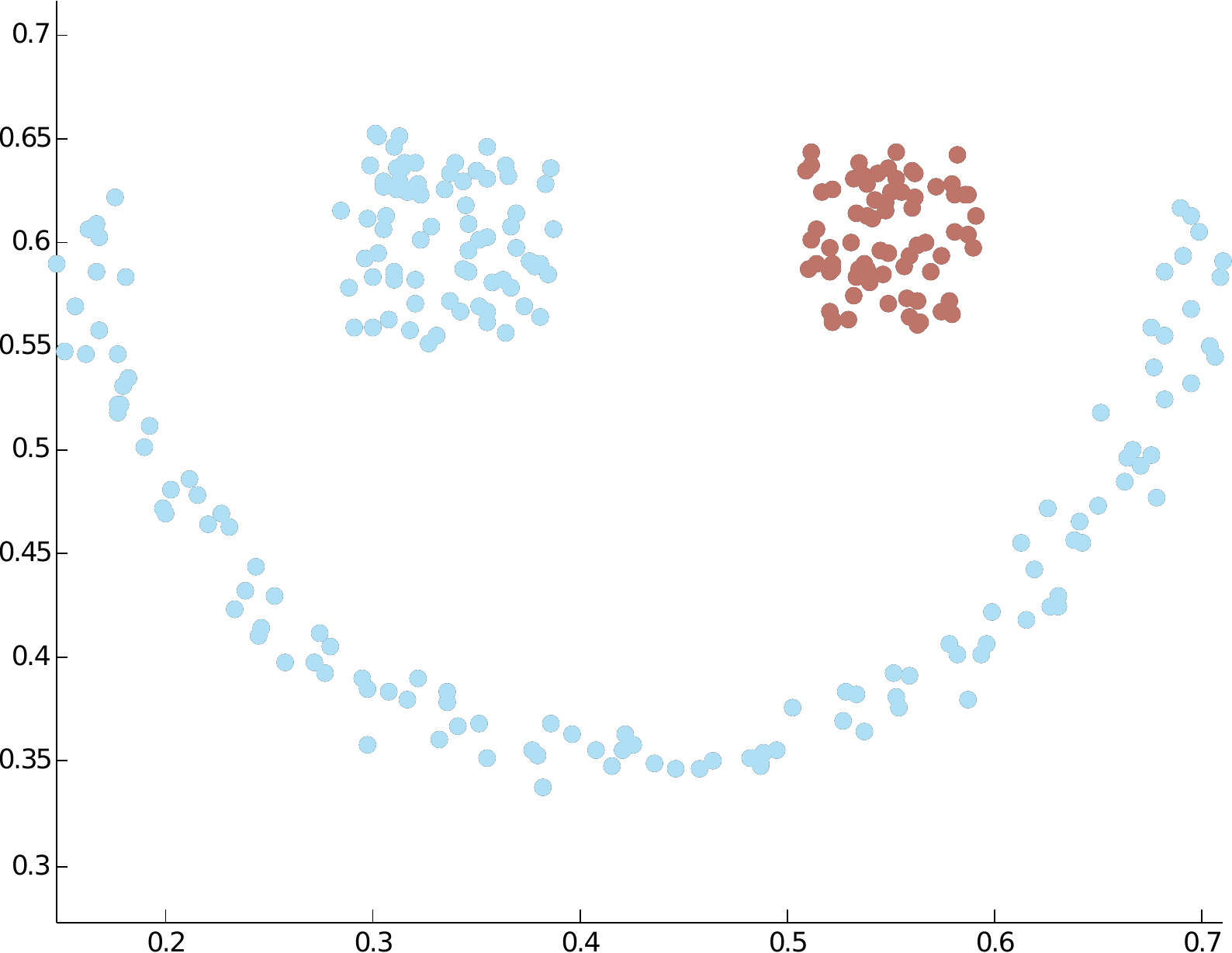}
			\fbox{\includegraphics[width=.36\textwidth]{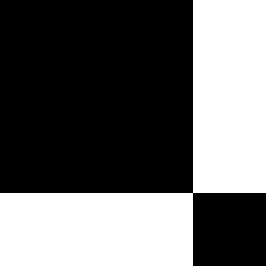}}\\
			\textsc{nmi}=0.421\\
			\textsc{f-m}=67.39
		}
		
		\caption{J-linkage~\cite{toldo08}}
		\label{fig:smiley_jl}
	\end{subfigure}
	\hfill
	\begin{subfigure}[b]{.325\textwidth}
		\centering
		\shortstack{
			\includegraphics[width=.6\textwidth]{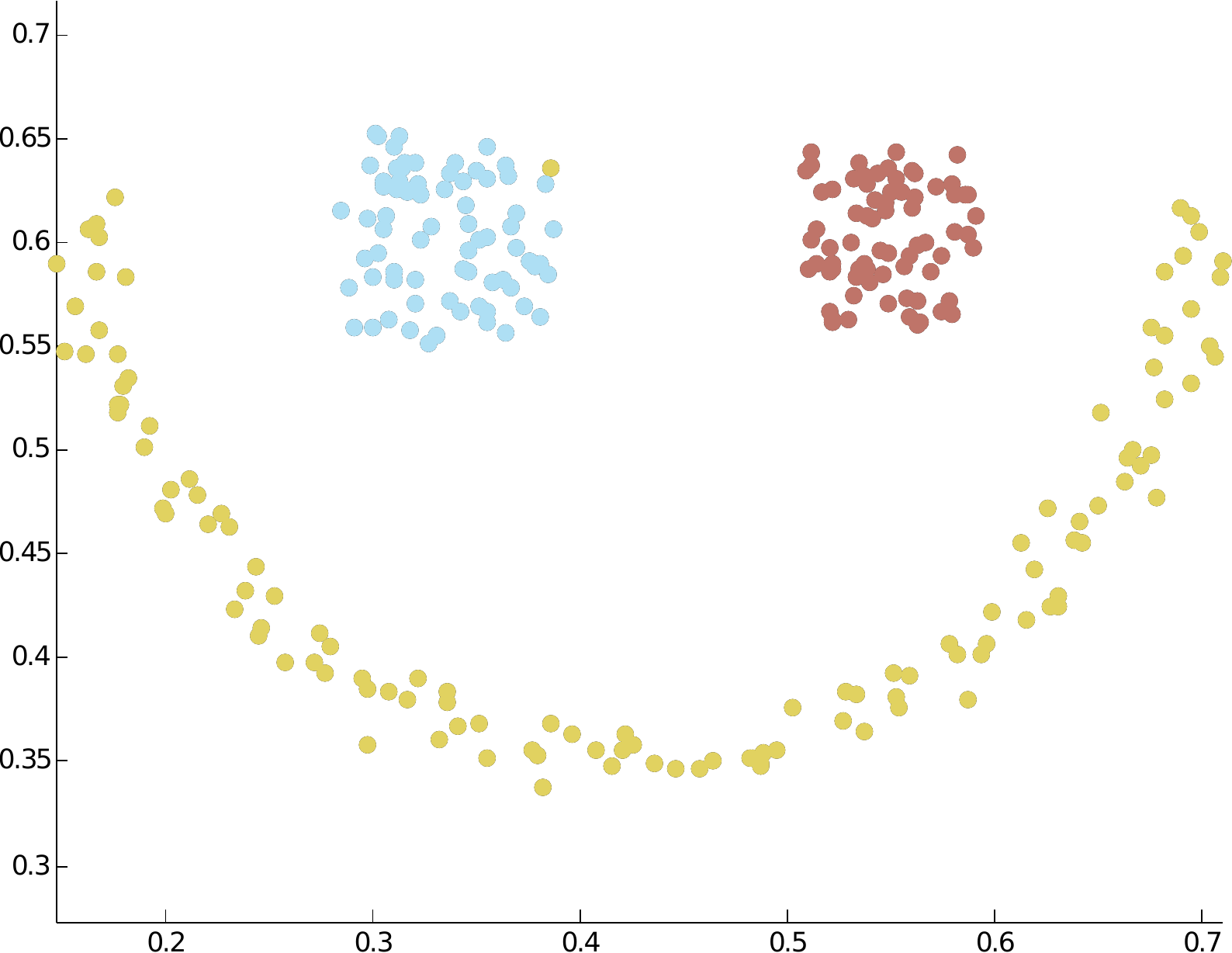}
			\fbox{\includegraphics[width=.36\textwidth]{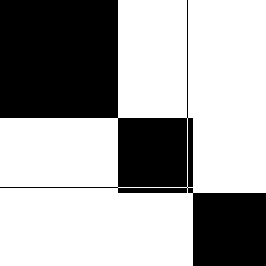}}\\
			\textsc{nmi}=0.981\\
			\textsc{f-m}=99.62
		}
		
		\caption{Bi-clustering}
		\label{fig:smiley_bc}
	\end{subfigure}
		
	\caption{Solving the consensus problem on a standard synthetic dataset~\cite{zelnik04}.
	\subref{fig:smiley_base}~Results obtained with the base algorithms.
	\subref{fig:smiley_spectral}~Clustering the co-occurrence matrix might yield good results but introduces new parameters, depending on the employed clustering algorithm. \subref{fig:smiley_snmf}~SNMF yields a poor result, even when the correct number of classes is specified. \subref{fig:smiley_jl}~J-linkage yields poor results. \subref{fig:smiley_bc}~A single point is misclassified with the proposed bi-clustering approach. NMI and F-M stand for normalized mutual information and F-measure, respectively.}	
	\label{fig:smiley}
	
\end{figure}

We compare in Figure~\ref{fig:smiley} our bi-clustering approach with clustering-based consensus methods on a standard synthetic dataset\footnote[1]{\url{http://www.vision.caltech.edu/lihi/Demos/SelfTuningClustering.html}}. Our first observation is that spectral clustering can obtain good results as a consensus algorithm, as long as the correct number of classes is used. Notice that this is sort of a self-defeating argument because we are introducing new (and critical) parameters for getting rid of other parameters. The same consideration is valid for SNMF~\cite{li2007}, although in this case we obtain a very poor solution even if the correct number of classes is used (matrix $\mat{B}$ is not really clustered, a thresholding of $\mat{H} \mat{D} \transpose{\mat{H}}$ being still needed, see problem~(\ref{eq:li2007snmf})). Another clustering algorithm, J-linkage~\cite{toldo08}, popular in parametric model estimation, also yields a poor solution. Our algorithm, without tuning any parameters yields a good solution and only misclassifies a single point (this occurs because 5 out of the 8 algorithms wrongly classify that specific point).

\begin{figure}[p]

	\begin{subtable}[b]{\textwidth}
		\centering
		
		\begin{footnotesize}
		\begin{tabular}{ l @{\hspace{6pt}} l *{4}{d{1.3} d{2.2}}}
			\toprule
			      &       & \multicolumn{2}{c}{Breast} & \multicolumn{2}{c}{Glass} & \multicolumn{2}{c}{Iris} & \multicolumn{2}{c}{Wine} \\
			\cmidrule(lr){3-4} \cmidrule(lr){5-6} \cmidrule(lr){7-8} \cmidrule(lr){9-10}
			      &       & \multicolumn{1}{c}{NMI} & \multicolumn{1}{c}{F-M} & \multicolumn{1}{c}{NMI} & \multicolumn{1}{c}{F-M} & \multicolumn{1}{c}{NMI} & \multicolumn{1}{c}{F-M} & \multicolumn{1}{c}{NMI} & \multicolumn{1}{c}{F-M} \\
			\midrule
			      
			A1 & ($K_{\textsc{gt}}-1$)-means
			& \multicolumn{1}{c}{--}      & \multicolumn{1}{c}{--}      & 0.368 & 54.74 & 0.657 & 76.35 & 0.478 & 66.59 \\
			A2 & ($K_{\textsc{gt}}$)-means
			& 0.733 & 95.73 & 0.315 & 49.51 & 0.758 & 89.18 & 0.876 & 96.60 \\
			A3 & ($K_{\textsc{gt}}+1$)-means
			& 0.708 & 93.98 & 0.358 & 50.21 & 0.722 & 82.43 & 0.791 & 89.34 \\
			A4 & ($K_{\textsc{gt}}+2$)-means
			& \multicolumn{1}{c}{--}      & \multicolumn{1}{c}{--}      & 0.378 & 50.30 & 0.694 & 76.29 & 0.724 & 79.84 \\
			
			A5 & SC ($1, K_{\textsc{gt}}-1$)
			& \multicolumn{1}{c}{--}      & \multicolumn{1}{c}{--}      & 0.289 & 46.99 & \multicolumn{1}{c}{--}      & \multicolumn{1}{c}{--}      & 0.470 & 64.94 \\
			A6 & SC ($1, K_{\textsc{gt}}$)
			& 0.791 & 96.93 & 0.348 & 47.22 & 0.742 & 88.53 & \multicolumn{1}{B{1.3}}{0.928} & \multicolumn{1}{B{2.2}}{98.31} \\
			A7 & SC ($1, K_{\textsc{gt}}+1$)
			& 0.705 & 93.02 & \multicolumn{1}{B{1.3}}{0.385} & \multicolumn{1}{B{2.2}}{49.19} & 0.623 & 77.81 & 0.720 & 82.53 \\
			A8 & SC ($1, K_{\textsc{gt}}+2$)
			& \multicolumn{1}{c}{--}      & \multicolumn{1}{c}{--}      & 0.360 & 43.23 & 0.736 & 79.11 & 0.620 & 74.50 \\
			A9 & SC ($1, K_{\textsc{gt}}-1$)
			& \multicolumn{1}{c}{--}      & \multicolumn{1}{c}{--}      & 0.315 & 47.48 & \multicolumn{1}{c}{--}      & \multicolumn{1}{c}{--}      & 0.470 & 64.94 \\
			A10 & SC ($1, K_{\textsc{gt}}$)
			& \multicolumn{1}{B{1.3}}{0.799} & \multicolumn{1}{B{2.2}}{97.08} & 0.348 & 47.22 & 0.735 & 87.82 & 0.911 & 97.74 \\
			A11 & SC ($1, K_{\textsc{gt}}+1$)
			& 0.709 & 92.46 & 0.352 & 45.25 & 0.665 & 79.16 & 0.731 & 83.08 \\
			A12 & SC ($1, K_{\textsc{gt}}+2$)
			& \multicolumn{1}{c}{--}      & \multicolumn{1}{c}{--}      & 0.351 & 42.66 & 0.706 & 78.17 & 0.602 & 75.56 \\
			A13 & SC ($1, K_{\textsc{gt}}-1$)
			& \multicolumn{1}{c}{--}      & \multicolumn{1}{c}{--}      & 0.315 & 46.86 & \multicolumn{1}{c}{--}      & \multicolumn{1}{c}{--}      & 0.470 & 64.94 \\
			A14 & SC ($1, K_{\textsc{gt}}$)
			& \multicolumn{1}{B{1.3}}{0.799} & \multicolumn{1}{B{2.2}}{97.08} & 0.348 & 47.22 & 0.708 & 86.53 & 0.911 & 97.74 \\
			A15 & SC ($1, K_{\textsc{gt}}+1$)
			& 0.691 & 91.97 & 0.350 & 44.75 & 0.613 & 78.20 & 0.731 & 83.08 \\
			A16 & SC ($1, K_{\textsc{gt}}+2$)
			& \multicolumn{1}{c}{--}      & \multicolumn{1}{c}{--}      & 0.351 & 42.66 & 0.661 & 74.59 & 0.602 & 75.56 \\
			
			\rowcolor{LightGreen}
			\multicolumn{2}{l}{Consensus}
			& \multicolumn{1}{B{1.3}}{0.799} & \multicolumn{1}{B{2.2}}{97.08} & 0.381 & 46.89 & \multicolumn{1}{B{1.3}}{0.770} & \multicolumn{1}{B{2.2}}{89.12} & 0.915 & 98.01 \\
			
			\bottomrule
		\end{tabular}%
		\end{footnotesize}
		
		\caption{Normalized mutual information (NMI) and F-measure (F-M) values for each individual algorithm and for the proposed consensus solution. SC ($\sigma, K$) stands for spectral clustering with kernel $\sigma$ and $K$ clusters. For $K$-means, the value between parenthesis indicates the actual value of $K$. $K_{\textsc{gt}}$ indicates the ground truth number of classes.}
		\label{fig:uci_table}
	\end{subtable}

	\begin{subfigure}[b]{\textwidth}
	
		\begin{small}
		\centerline{
		\hfill
		\shortstack{%
			Breast\\
			\includegraphics[width=.49\textwidth]{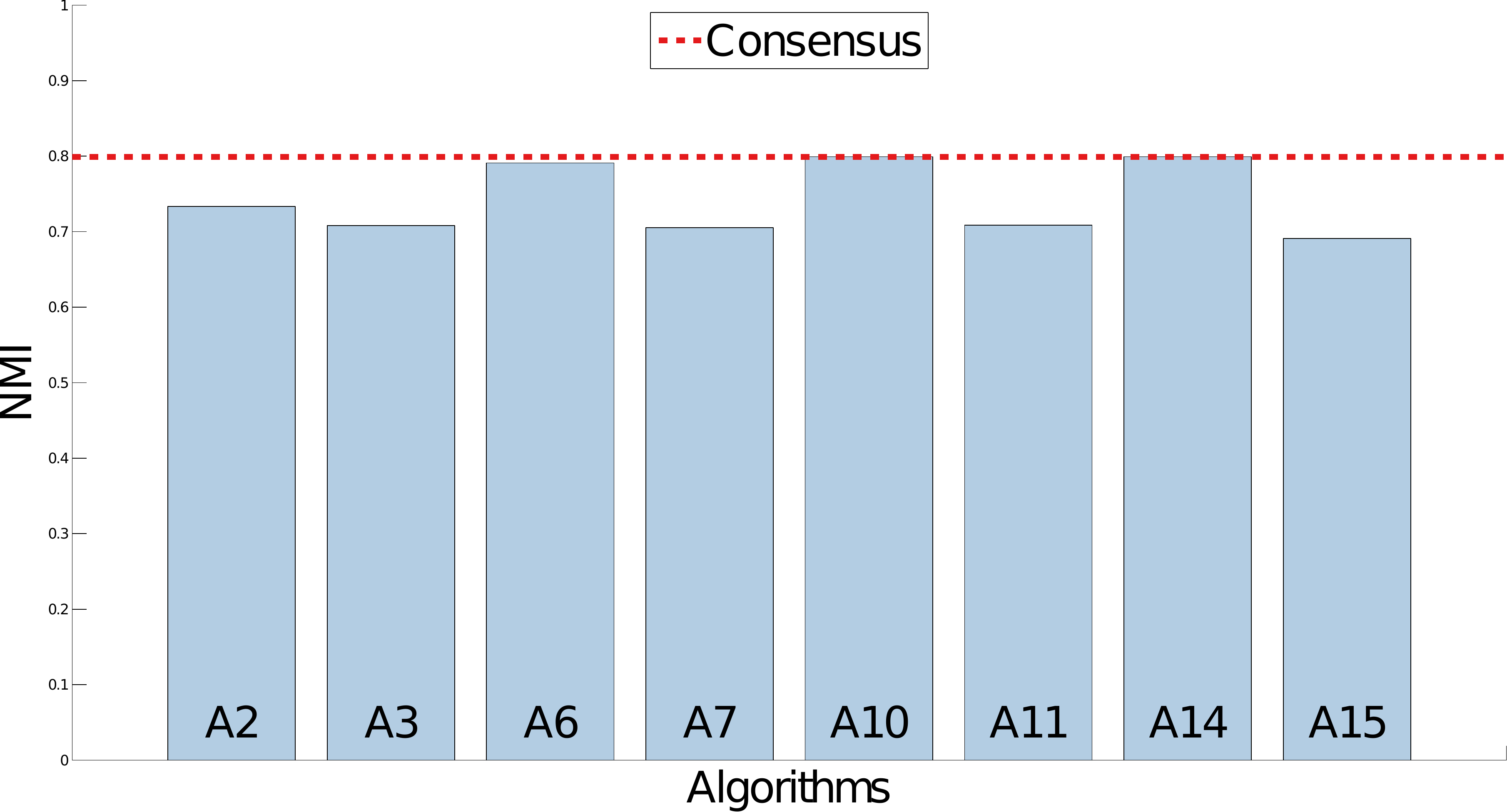}
		}
		\hfill
		\shortstack{%
			Glass\\
			\includegraphics[width=.49\textwidth]{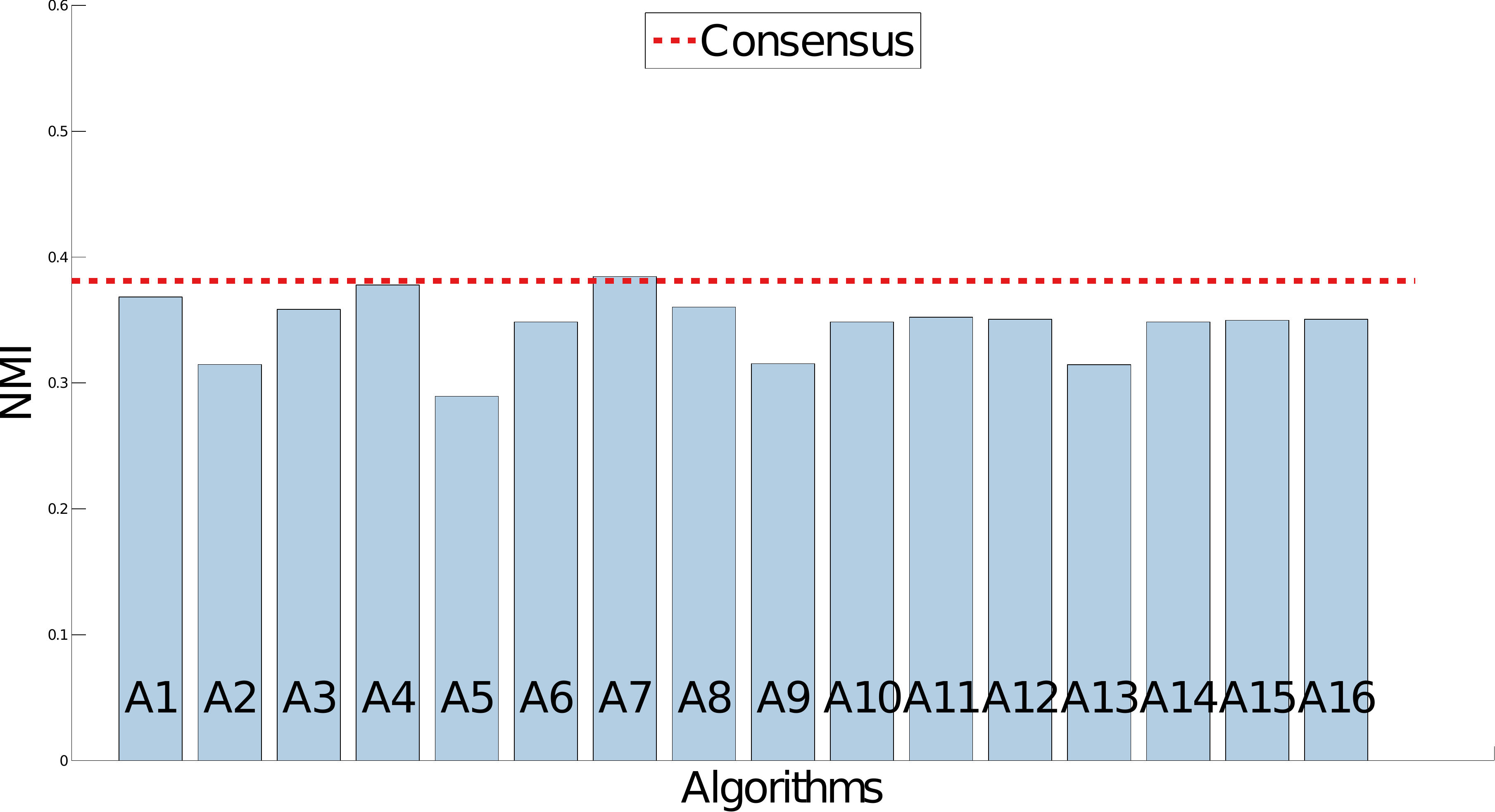}
		}
		\hfill
		}
		
		\vskip.5em
		
		\centerline{
		\hfill
		\shortstack{%
			Iris\\
			\includegraphics[width=.49\textwidth]{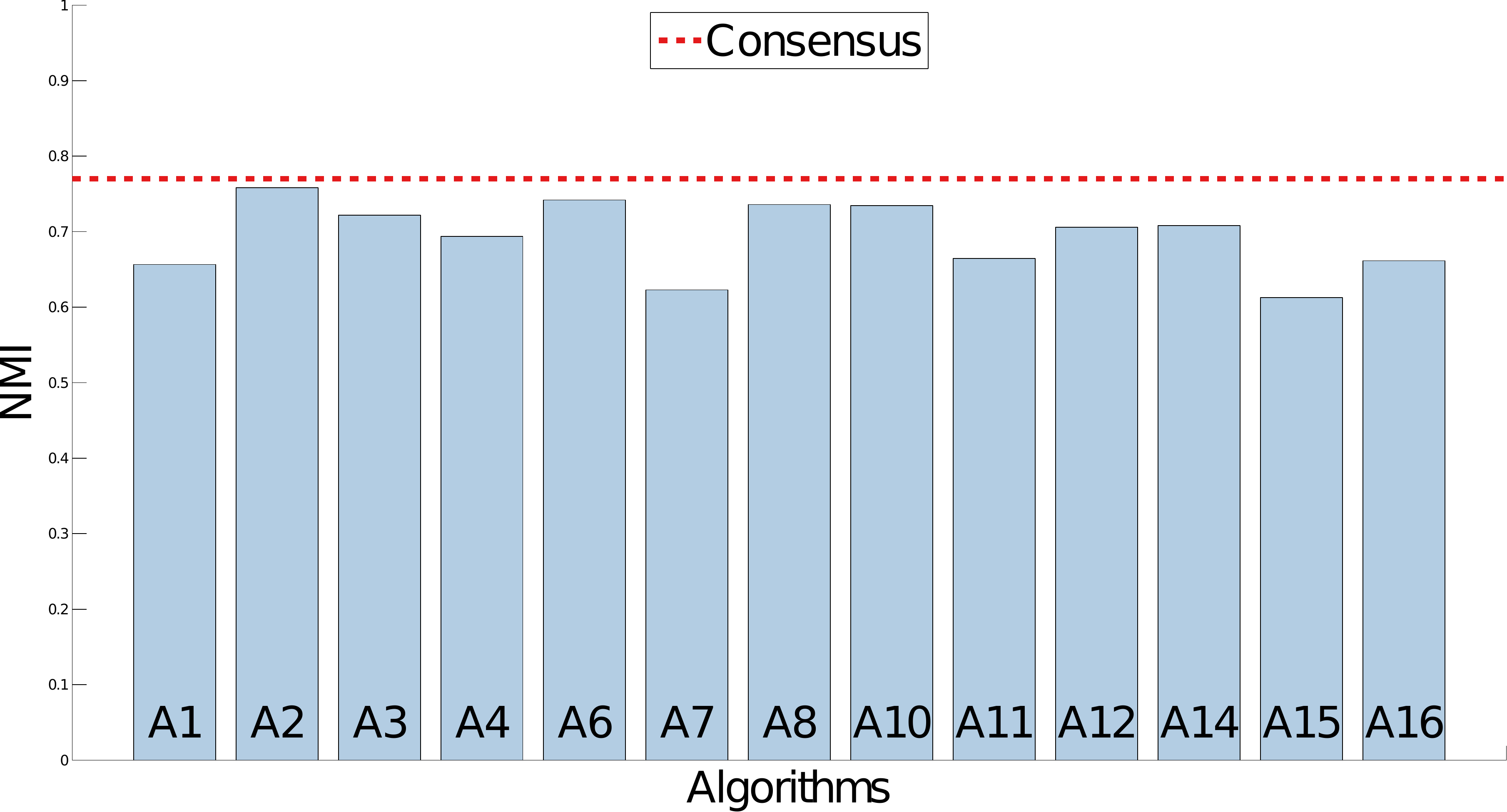}
		}
		\hfill
		\shortstack{%
			Wine\\
			\includegraphics[width=.49\textwidth]{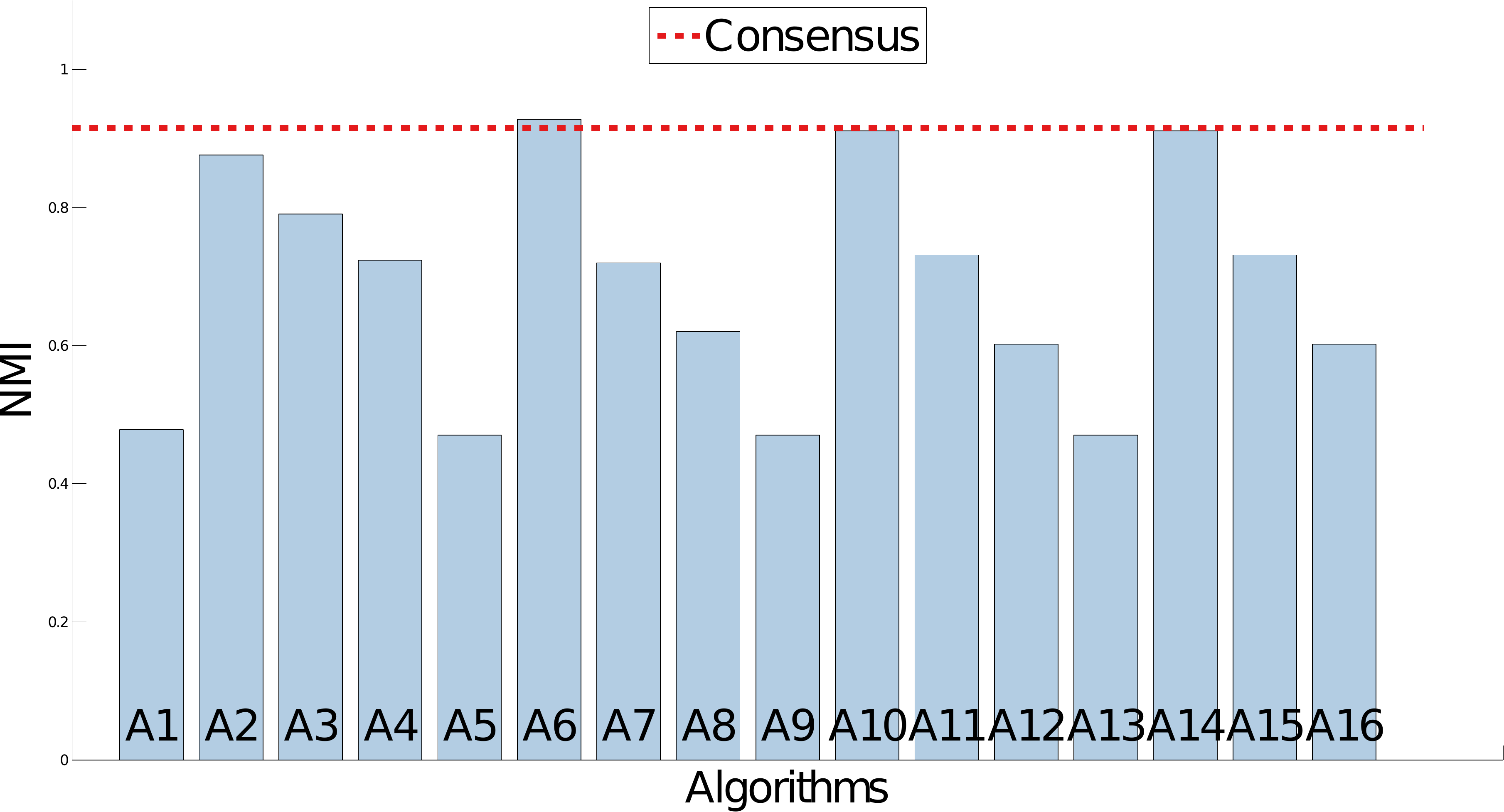}
		}
		\hfill
		}
		\end{small}
		
		\caption{Graph representation of Table~\subref{fig:uci_table}, highlighting the quality of the proposed consensus solution.}
	\end{subfigure}
	
	\caption{Consensus experiments on datasets from the UCI repository~\cite{uciRepository}. The proposed bi-clustering solution provides normalized mutual information (NMI) values that are universally on par with the best algorithms from the pool (which is dataset dependent and with carefully tuned parameters), without any information about the nature of the algorithms or their parameters.}
	\label{fig:clustering_UCI}
\end{figure}

We also present experiments on standard real datasets from the UCI repository~\cite{uciRepository} in Figure~\ref{fig:clustering_UCI}, where we compute the consensus solution of several instances of $K$-means and spectral clustering. In all four examples, we obtained results competitive with the best solutions in the pool  (the best being always different for different datasets, our proposed framework is universally good). In the Iris dataset, our consensus solution is better than all of the base solutions.
When the dataset contains only two classes, e.g., the Breast dataset,  it becomes much harder to optimize for the number of classes since in the pool of solutions there is no balance between under and over-clustered solutions: all of them either have the right number of clusters or are over-clustered. In this case, we obtained good results by only employing algorithms that yield two or three clusters.

\textbf{Subspace clustering.}
In many problems involving high-dimensional data, each class or category spans a low-dimensional subspace of the high-dimensional ambient space.
The data are a set $\set{X}$ of $m$ $d$-dimensional data points $\{ \vect{x}_i \in \Real^d \}_{i=1}^m$ that lie approximately in a union of low-dimensional subspaces.
Sparse subspace clustering aims at clustering these data points, finding the correct subspaces.
The data are represented as a matrix $\mat{D} \in \Real^{m \times d}$, whose columns are the vectors $\vect{x}_i$.
In the case of noisy data, the underlying algorithm is defined as~\cite{elhamifar2013}:
\begin{enumerate}
	\item Solve the sparse optimization problem
	\begin{equation}
	\min_{\mat{C}, \mat{Z}} \norm{\mat{C}}{1} + \tfrac{\kappa}{2 \mu_{\mat{D}}} \norm{\mat{Z}}{F}^2
	\quad
	\text{s.t.}
	\quad
	\begin{gathered}
	\mat{D} = \mat{D} \mat{C} + \mat{Z},\\
	\operatorname{diag} (\mat{C}) = \mat{0} ,
	\end{gathered}
	\label{eq:sparse_subspace_clustering}
	\end{equation}
	where $\mu_{\mat{D}} \in \Real^+$ is a normalization coefficient dependent on $\mat{D}$, and $\kappa \in \Real^+$ is a parameter of the algorithm that controls how much denoising we wish to apply.
	\item Normalize the columns $(\mat{C})_i$ of $\mat{C}$ as $(\mat{C})_i \gets (\mat{C})_i / \norm{(\mat{C})_i}{\infty}$.
	\item Form a similarity graph with $m$ nodes representing the data points, set the affinity matrix $\mat{W}$ of the graph to $\mat{W} = |\mat{C}| + |\transpose{\mat{C}}|$ (this is the element-wise absolute value operator, i.e., $(|\mat{B}|)_{ij} = |(\mat{B})_{ij}|$).
	\item Apply spectral clustering~\cite{ng01} to the similarity graph, the number $K$ of desired clusters/subspaces being a parameter of the algorithm.
\end{enumerate}

\begin{figure}

	\begin{subfigure}[t]{.59\textwidth}
		\centering
		\begin{small}
		\begin{tabular}{r *{6}{d{1.2}}}
			\toprule
			      & \multicolumn{3}{c}{Original features} & \multicolumn{3}{c}{Compressed features} \\
			\cmidrule(lr){2-4} \cmidrule(lr){5-7}
			\multicolumn{1}{c}{$\kappa$}     & \multicolumn{1}{c}{2}     & \multicolumn{1}{c}{3}     & \multicolumn{1}{c}{All}   & \multicolumn{1}{c}{2}     & \multicolumn{1}{c}{3}     & \multicolumn{1}{c}{All} \\
			\midrule
			100   & \multicolumn{1}{B{1.2}}{1.46}  & 5.86  & 2.45  & \multicolumn{1}{B{1.2}}{1.48}  & 5.85  & 2.47 \\
			200   & 1.61  & 5.50  & 2.49  & 1.62  & 5.65  & 2.53 \\
			300   & 1.60  & 4.78  & 2.31  & 1.58  & 5.30  & 2.42 \\
			400   & 1.53  & 4.84  & 2.27  & 1.62  & 5.37  & 2.46 \\
			500   & 1.61  & 4.55  & 2.27  & 1.66  & 4.52  & \multicolumn{1}{B{1.2}}{2.30} \\
			600   & 1.54  & 4.50  & 2.20  & 1.62  & 4.54  & 2.28 \\
			700   & 1.66  & 4.36  & 2.27  & 1.73  & 4.41  & 2.34 \\
			\rowcolor{LightYellow}
			800   & 1.53  & 4.40  & 2.18  & 1.83  & 4.40  & 2.41 \\
			900   & 1.95  & 5.43  & 2.73  & 1.79  & 5.56  & 2.64 \\
			1000  & 1.79  & 4.58  & 2.42  & 2.36  & 5.34  & 3.03 \\
			1100  & 2.16  & 5.25  & 2.86  & 2.15  & 5.24  & 2.85 \\
			1200  & 2.14  & 5.28  & 2.85  & 2.17  & 5.19  & 2.85 \\
			1300  & 2.20  & 5.27  & 2.90  & 2.53  & 5.12  & 3.12 \\
			1400  & 2.19  & 5.15  & 2.86  & 2.33  & 5.24  & 2.99 \\
			
			\rowcolor{LightGreen}
			Cons. & 1.56  & \multicolumn{1}{B{1.2}}{4.25}  & \multicolumn{1}{B{1.2}}{2.17}  & 1.78  & \multicolumn{1}{B{1.2}}{4.30}  & 2.35 \\
			\bottomrule
		\end{tabular}%
		\end{small}
		
		\caption{Misclassification errors. In orange we observe the value of $\kappa$ hand-selected in~\cite{elhamifar2013}. In green, we observe the result of computing the consensus solution of all the subspace clustering instances.}
		\label{fig:subspace_table}
	\end{subfigure}
	\hfill
	\begin{subfigure}[t]{.395\textwidth}
		\centering
		\raisebox{2em}{\begin{sideways}{\tiny Misclassification error}\end{sideways}}\hspace{1pt}%
		\includegraphics[width=.96\textwidth]{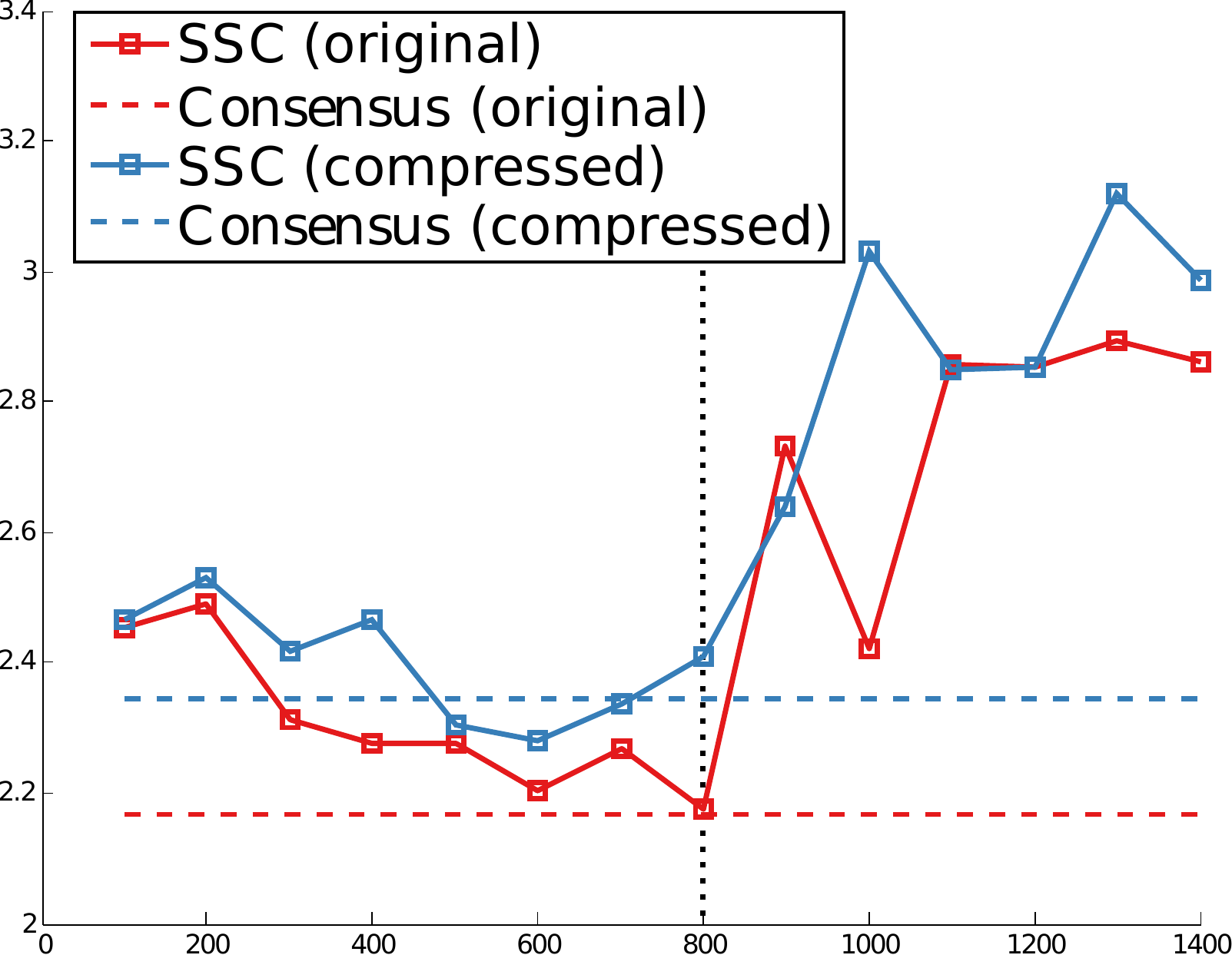}\\[-8pt]
		{\tiny $\kappa$}
		
		\caption{Graph representation of Table~\subref{fig:subspace_table} considering all instances in the dataset. The value $\kappa=800$ is marked with a vertical dotted line. Very competitive misclassification errors are obtained with our consensus solution, without any information about the nature of the algorithms or their parameters.}
		\label{fig:subspace_plot}
	\end{subfigure}
	
	\caption{Subspace clustering example for the Hopkins 155 dataset~\cite{hopkins155}, which consists of 155 video sequences of 2 or 3 motions, each corresponding to a low-dimensional subspace in each video. We use the original trajectories and the data projected into a lower-dimensional space using PCA. The base algorithms are different subspace clustering instances~\cite{elhamifar2013}, varying the regularization parameter $\kappa$ in~(\ref{eq:sparse_subspace_clustering}). We show the mean misclassification error (\%) for each of them and for our consensus solution.}
	\label{fig:subspace}
\end{figure}

We use subspace clustering for motion segmentation in videos: given feature points on multiple rigidly moving objects tracked in multiple frames of a video, the goal is to separate the feature trajectories according to the object's motions~\cite{elhamifar2013}.
We treat the case of subspace clustering as yet another grouping instance, therefore addressed by our proposed bi-clustering consensus approach. 
We run several instances of this algorithm with different values for the regularization parameter $\kappa$ and compute the consensus solution from them. (We also experimented with different numbers of classes, but sparse subspace clustering is very sensitive to using the wrong number of classes.) The results of this experiment are shown in Figure~\ref{fig:subspace}. In this case, following the methodology in~\cite{elhamifar2013}, we use misclassification error to analyze our results. Our first observation is that there is a range of $\kappa$ values that give correct results, thus enabling the use of consensus algorithms. The next observation is that very low misclassification errors are obtained with the consensus solution, sometimes even outperforming the best individual result. 

\begin{table}
	\caption{Bi-cluster sizes for several clustering experiments. Algorithm~\ref{algo:biclustering} returns $T$ bi-clusters. We compare the sizes of the $T$-th and $(T+1)$-th bi-clusters, i.e., the last returned and the first discarded, respectively. In the cases with missing numbers, the algorithm terminated because $\mat{A} = \emptyset$. These sizes help to understand why there is no need to carefully tune the thresholds $\tau_{\text{R}}, \tau_{\text{C}}$, considering the size gap between the $T$-th and $(T+1)$-th bi-clusters.}
	\label{tab:last_bicluster_sizes}

	\centering
	\begin{small}
	\begin{tabular}{l c c c c}
	\toprule
%	Figure & \multicolumn{4}{c}{Bi-cluster size} \\
%	\midrule
	& $\# \{ \set{R}_T \}$ & $\# \{ \set{C}_T \}$ & $\# \{ \set{R}_{T+1} \}$ & $\# \{ \set{C}_{T+1} \}$ \\
	\cmidrule(lr){2-3} \cmidrule(lr){4-5}
	Figure~\ref{fig:clustering_synthetic_gmm} & 90 & 7 & 2 & 2 \\
	Figure~\ref{fig:clustering_synthetic_mean-shift} & 295 & 5 & 2 & 1 \\
	Figure~\ref{fig:smiley} & 74 & 6 & - & - \\
	Figure~\ref{fig:clustering_UCI} (Breast) & 243 & 8 & - & - \\
	Figure~\ref{fig:clustering_UCI} (Glass) & 12 & 6 & 2 & 10 \\
	Figure~\ref{fig:clustering_UCI} (Iris) & 36 & 12 & - & - \\
	Figure~\ref{fig:clustering_UCI} (Wine) & 67 & 13 & 1 & 5 \\
	\bottomrule
	\end{tabular}
	\end{small}
\end{table}

\textbf{On the parameters $\tau_{\text{R}}, \tau_{\text{C}}$.}
One of the stoping criteria for Algorithm~\ref{algo:biclustering} is the size of the extracted bi-clusters. Each bi-cluster must have at least $\tau_{\text{R}}$ rows and $\tau_{\text{C}}$ columns. As mentioned above, for every clustering experiment we set $\tau_{\text{R}} = 6, \tau_{\text{C}} = 3$. Table~\ref{tab:last_bicluster_sizes} compares the size of the $T$-th and $(T+1)$-th bi-clusters (recall that Algorithm~\ref{algo:biclustering} only returns $T$ bi-clusters, hence discarding the $(T+1)$-th bi-cluster) for several clustering experiments. The size drop is in all cases so dramatic that $\tau_{\text{R}}, \tau_{\text{C}}$ become extremely easy to set.

% !TEX root = main.tex
\section{Consensus community detection in networks}
\label{sec:communities}

Networks are frequently used to describe many real-life scenarios were units interact with each other (e.g., see~\cite{newman04,aldecoa10} and references therein). A seemingly common property to many networks is the \emph{community structure}: networks can be divided into (in general non-overlapping) groups such that intra-group connections are denser than inter-group ones. Finding and analyzing these communities sheds light on important characteristics of the networks and the data they represent. However, the best way to establish the community structure is still disputed. Addressing this is the topic of this section.

Let $G = (V, E, \psi)$ be the graph to analyze, where $V$ is the set of $m$ nodes, $E$ is the set of edges, and $\psi : E \rightarrow \Real^+$ is a weighting function on the edges (in the following we indistinguishably use the terms graph and network). Generically, we consider that a community-detection algorithm provides a set $\set{C}$ of candidate communities ($\set{C} \subset \mathbb{P} (V)$, where $\mathbb{P} (V)$ is the power set of $V$).

\subsection{Experimental Results}
\label{sec:communities_results}

For the experiments we use the following base algorithms for community detection:
Louvain~\cite{blondel08louvain},
Infomap~\cite{rosvall2008}, and
Spectral Clustering (SC-($K$))~\cite{ng01}, where $K$ is the number of detected clusters/communities. For assessing the quality of a solution when ground truth is available, we again use normalized mutual information (NMI).
Unless specified, we use uniform weights  in $\mat{A}$.

\textbf{Several algorithms, same network.}
The most classical consensus scenario is when we have the result of several detection algorithms and wish to combine them into a better result.

\begin{figure}
    \centering
    
    \begin{small}
    \begin{tabular}{ @{\hspace{0pt}} *{4}{>{\centering\arraybackslash}m{.17\columnwidth} @{\hspace{3pt}}} >{\centering\arraybackslash}m{.17\columnwidth} @{\hspace{0pt}} }

        \shortstack{Preference\\matrix $\mat{A}$} &
        Bi-clustering &
        \shortstack{L1-NMF\\\begin{scriptsize}$(q=2)$\end{scriptsize}} &
        \shortstack{L1-NMF\\\begin{scriptsize}$(q=3)$\end{scriptsize}} &
        \shortstack{L1-NMF\\\begin{scriptsize}$(q=4)$\end{scriptsize}} \\

        \includegraphics[width=.17\columnwidth]{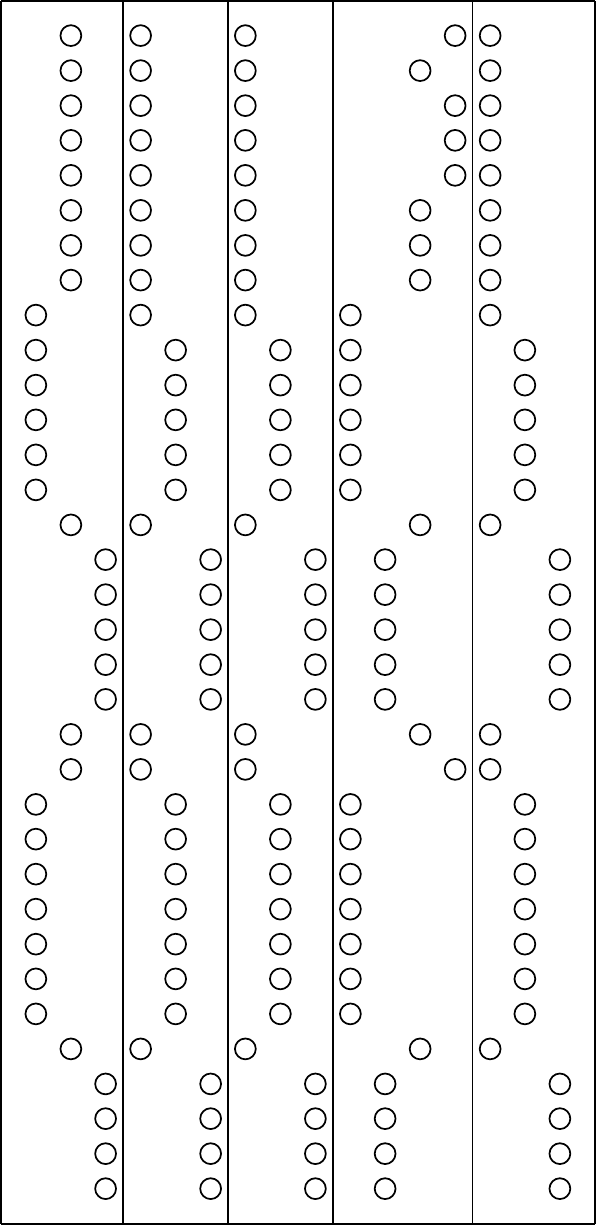} &
        \includegraphics[width=.17\columnwidth]{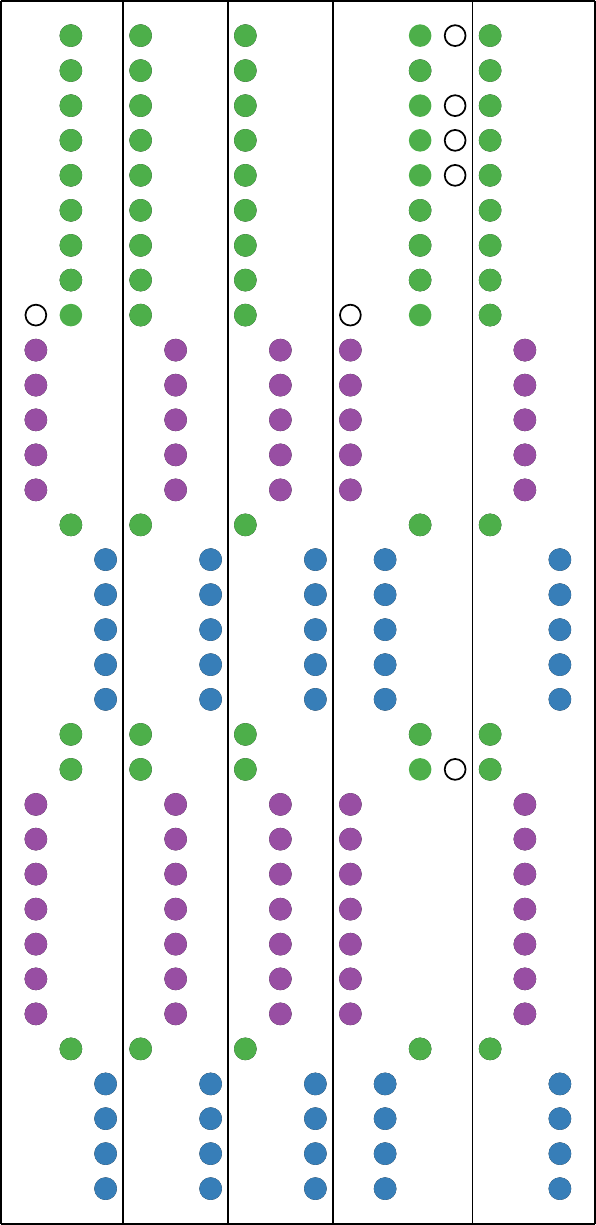} &
        \includegraphics[width=.17\columnwidth]{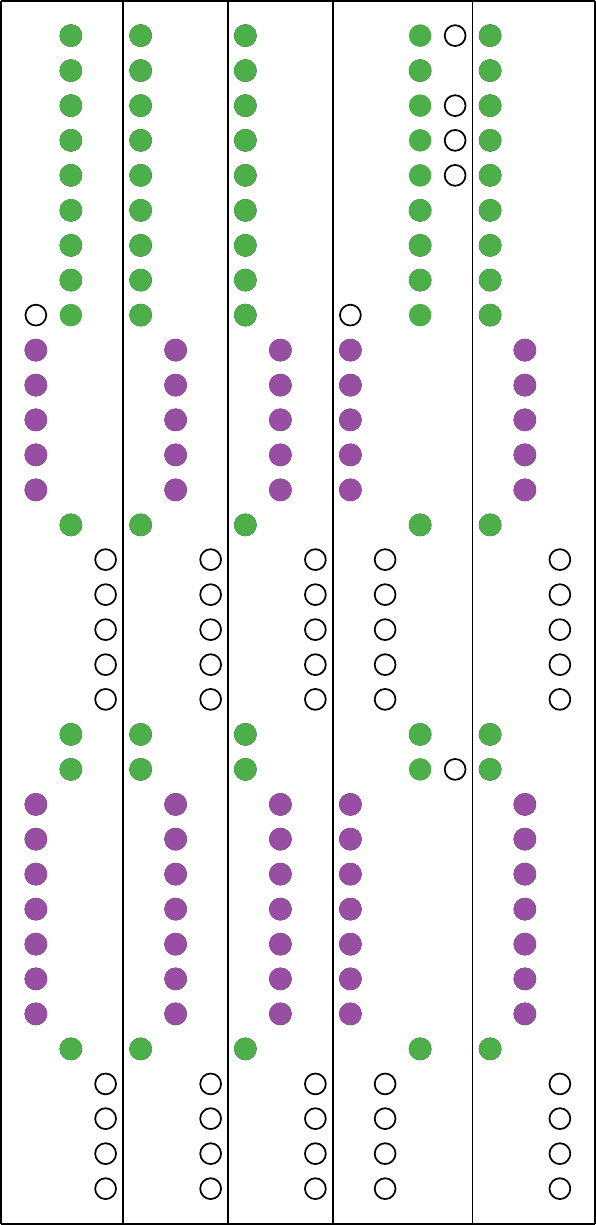} &
        \includegraphics[width=.17\columnwidth]{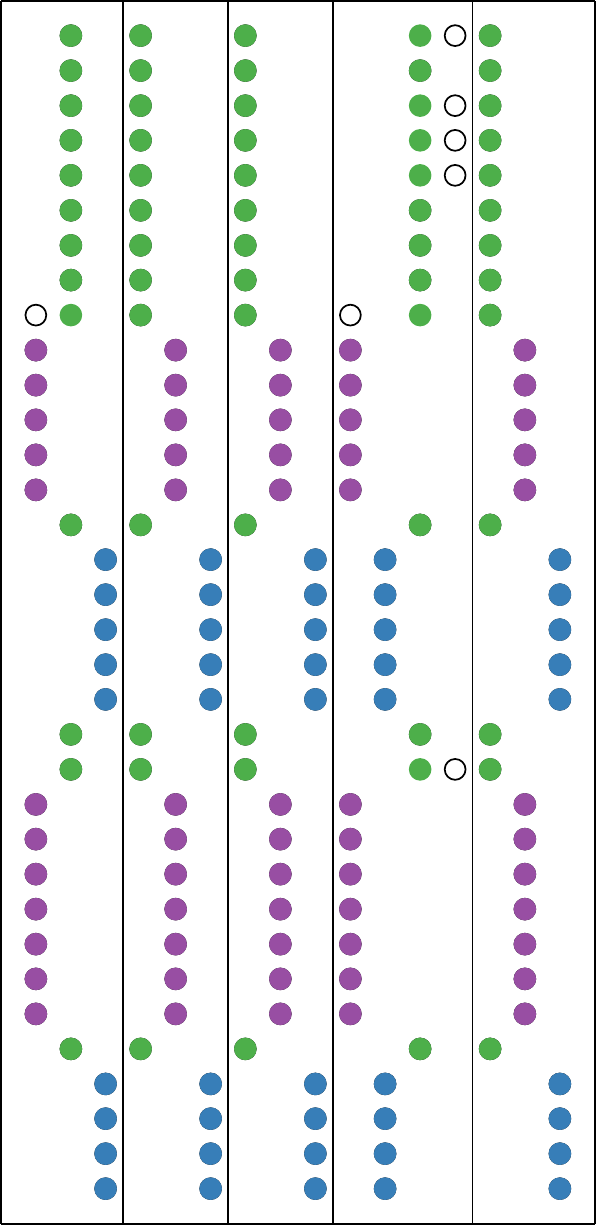} &
        \includegraphics[width=.17\columnwidth]{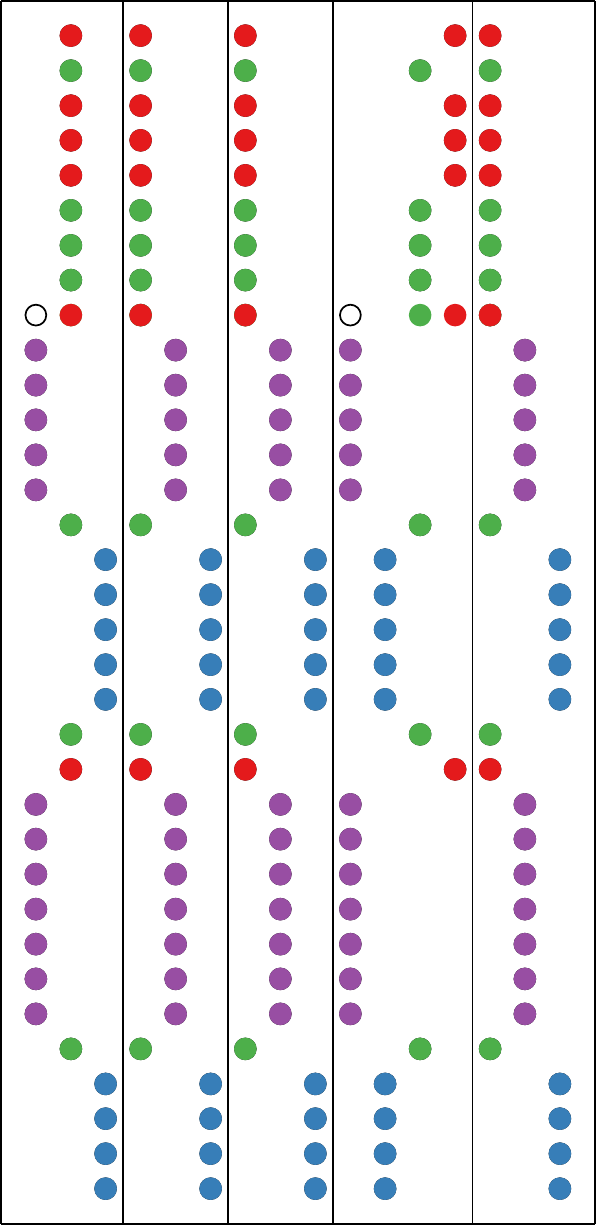} \\
    \end{tabular}
    \end{small}
    
    \caption{The proposed iterative bi-clustering approach finds the correct number of bi-clusters (3) on the Aegean34 network~\cite{evans11}, with 34 nodes (a different color is assigned to each pair $(\set{R}_t; \set{Q}_t)$, see Algorithm~\ref{algo:biclustering}). L1-NMF, see problem~(\ref{eq:nmf}), with $q=2$ undersegments $\mat{A}$ (some nodes are not assigned to any community, i.e., no color, despite a lot of consistency between the base algorithms), and with $q=4$ oversegments $\mat{A}$ (a single algorithm splitting a community is enough to create a new ``artificial'' consensus community, see the red entries).}
    \label{fig:greedyVSnmf}

\end{figure}

In Figure~\ref{fig:greedyVSnmf}, we first explore the differences between the proposed iterative rank-one L1-NMF approach and L1-NMF with $q \neq 1$. We use the Aegean34 network~\cite{evans11}, its small size makes easy to visualize the preference matrix $\mat{A}$. The network models the interactions between Middle Bronze Age (MBA) Aegean archaeological sites. Our algorithm recovers the correct number of bi-clusters, a critical parameter in classical NMF approaches.

\begin{figure*}
    \centering
    
	\hfill
    	\begin{minipage}[c]{.36\textwidth}
	        \centering
	         \begin{small}
	        Preference matrix $\transpose{\mat{A}}$\\[4pt]
	        \includegraphics[width=\textwidth]{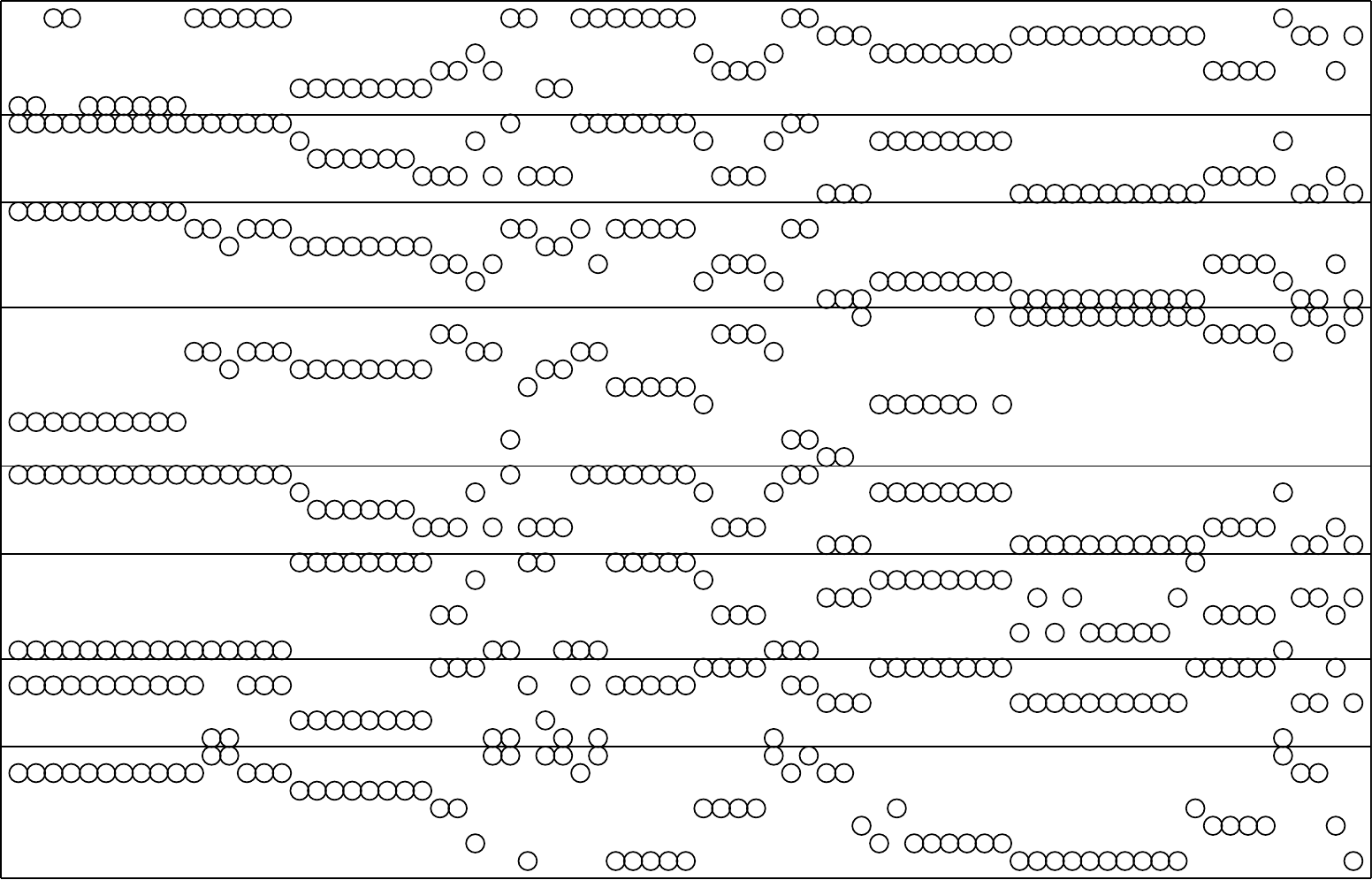}\\
	        Consensus \\[4pt]
	        \includegraphics[width=\textwidth]{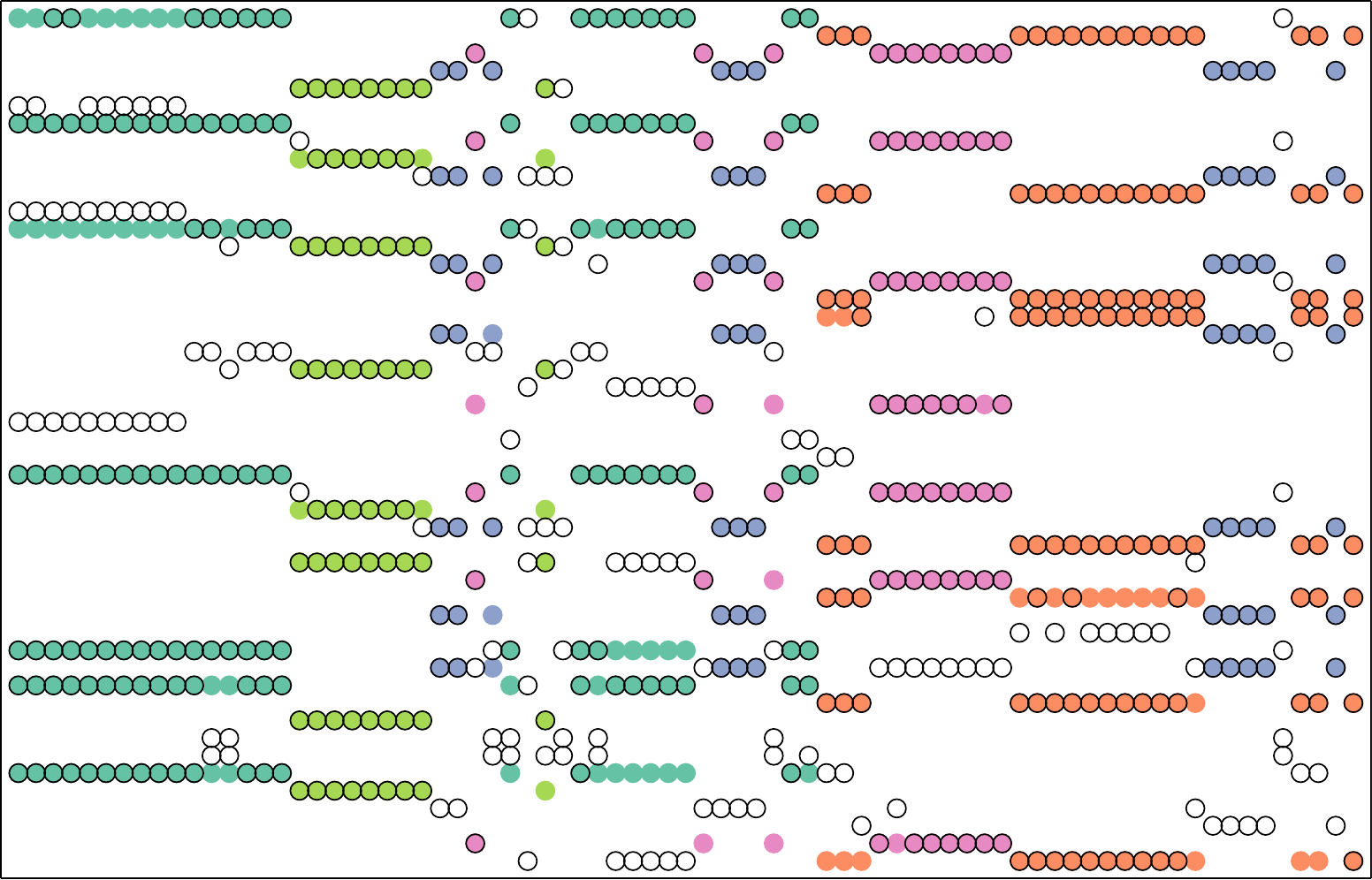}
	         \end{small}
    	\end{minipage}
	\hfill
    	\begin{minipage}[c]{.6\textwidth}
	        \centering
	        \begin{small}
	        \begin{tabular}{ @{\hspace{0pt}} c @{\hspace{1pt}} c @{\hspace{0pt}} }
	            Louvain &
	            Infomap \\
	            
	            \includegraphics[width=.49\textwidth]{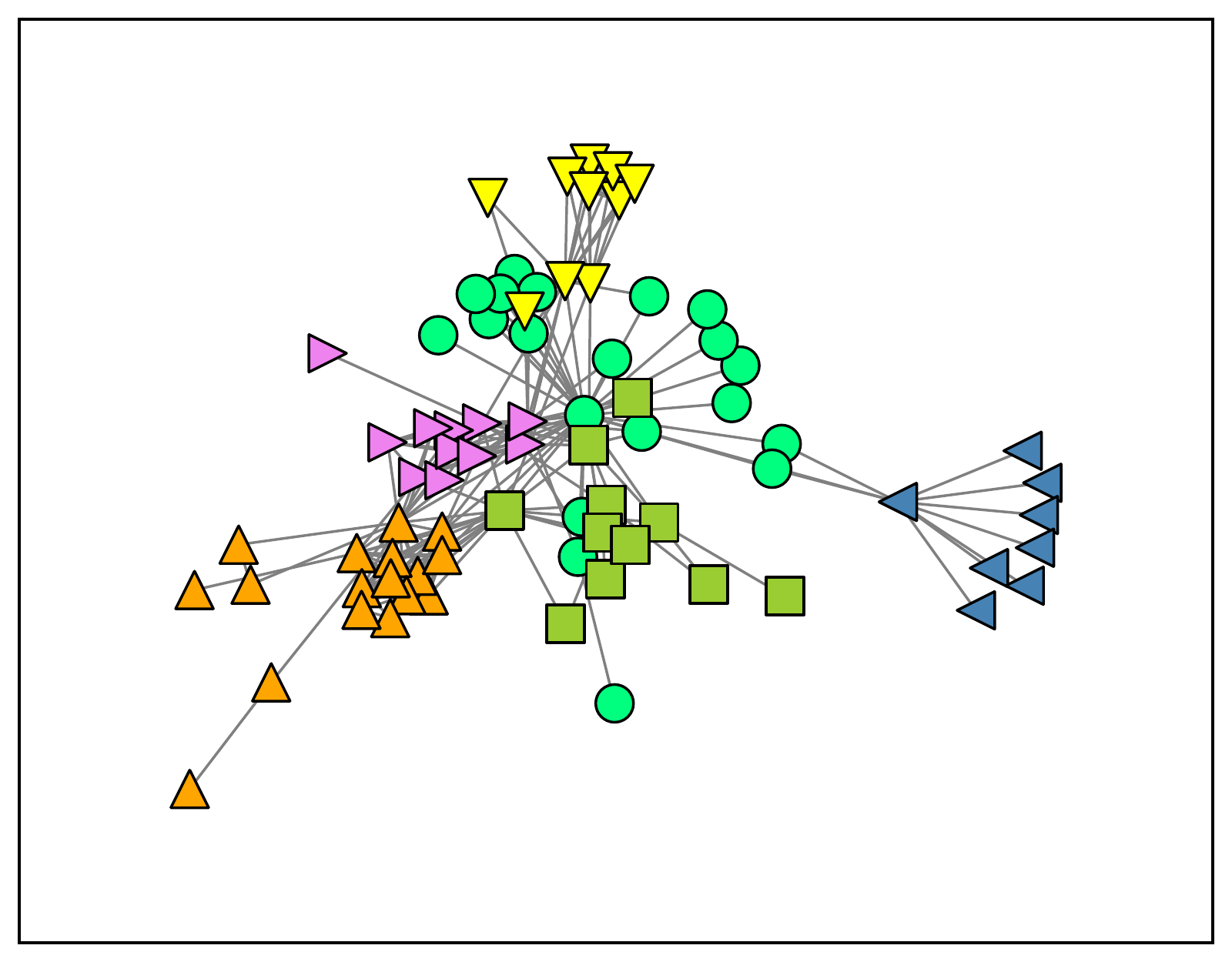} &
	            \includegraphics[width=.49\textwidth]{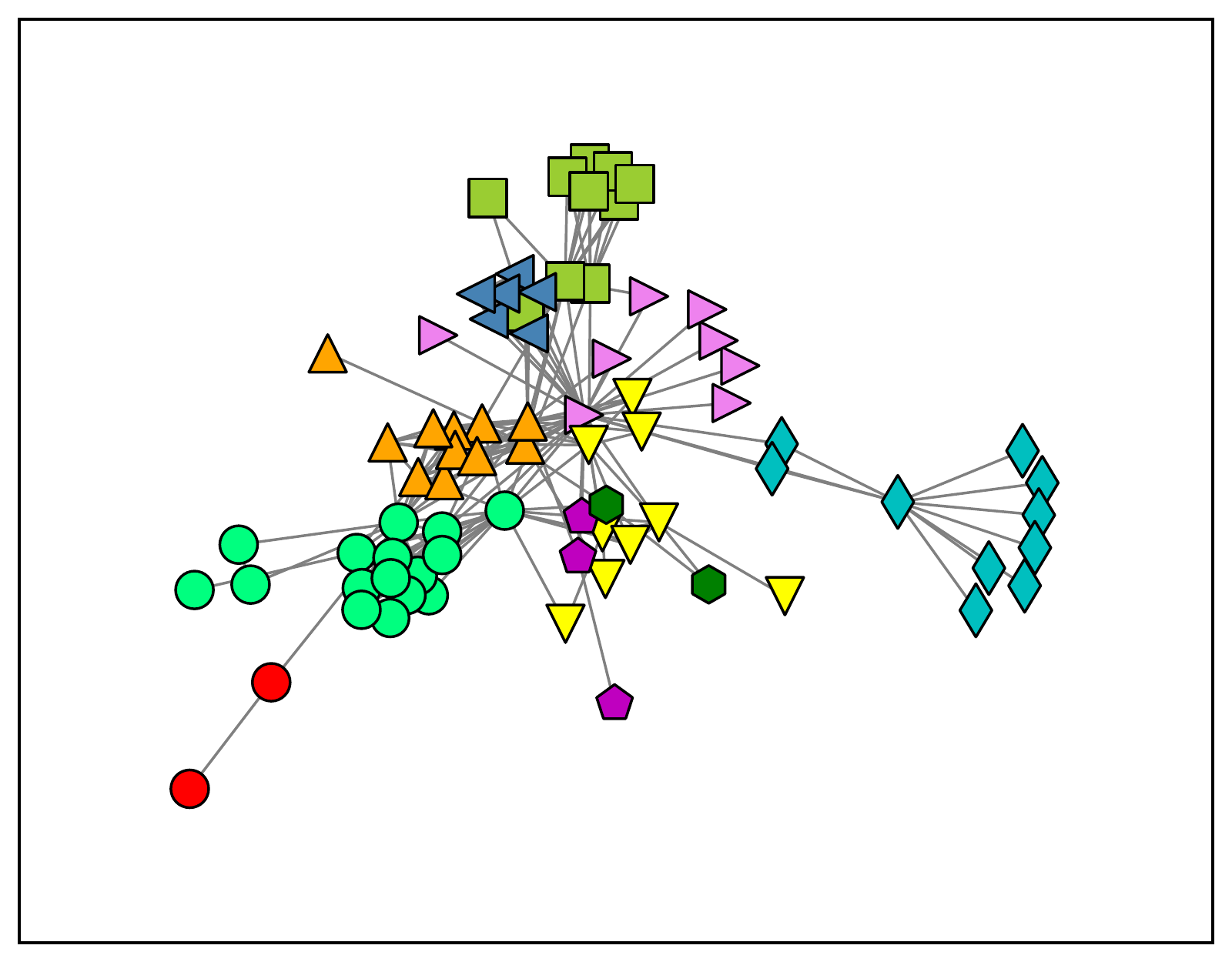} \\
	
	            Spectral clustering &
	            Consensus \\
	            
	            \includegraphics[width=.49\textwidth]{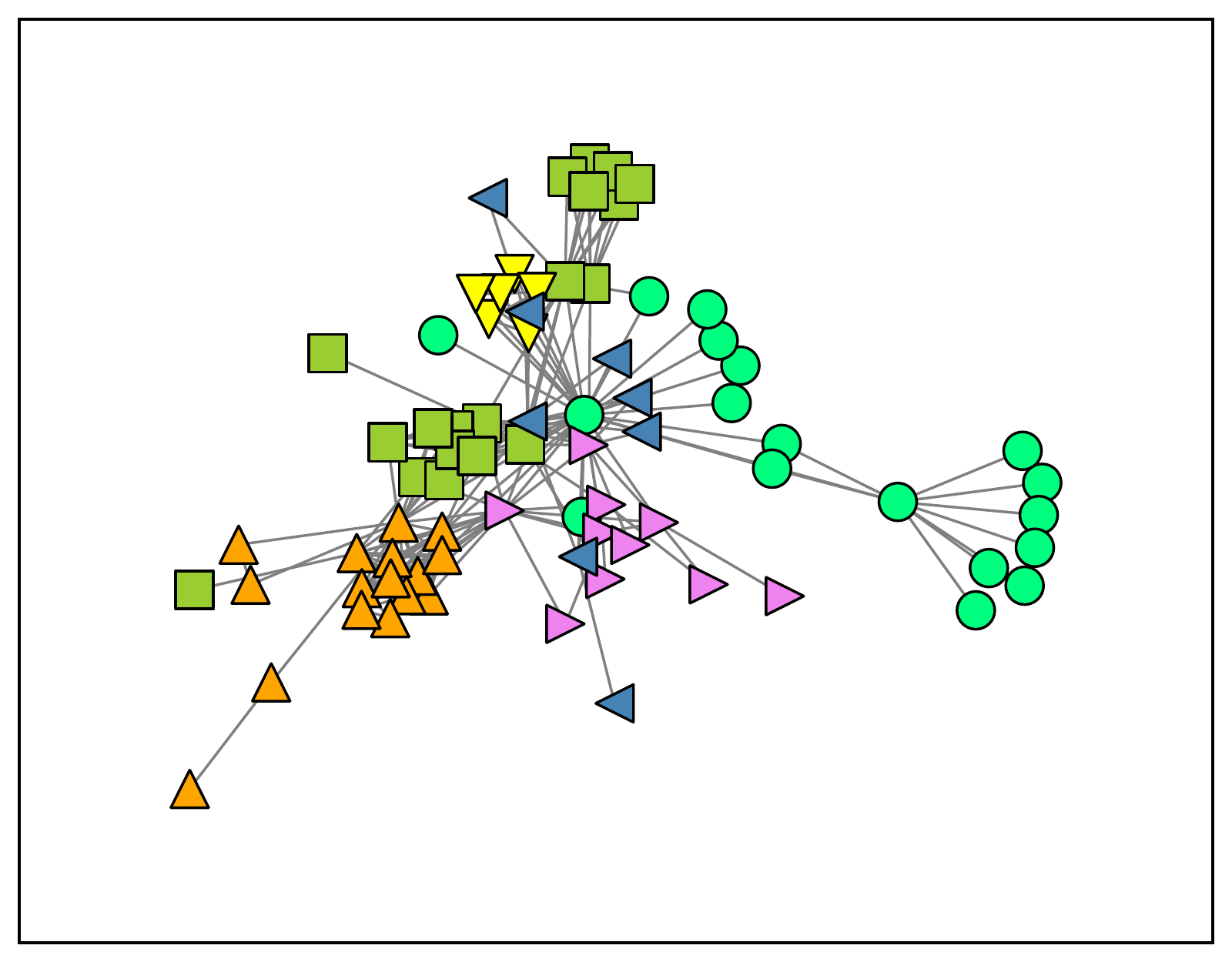} &
	            \includegraphics[width=.49\textwidth]{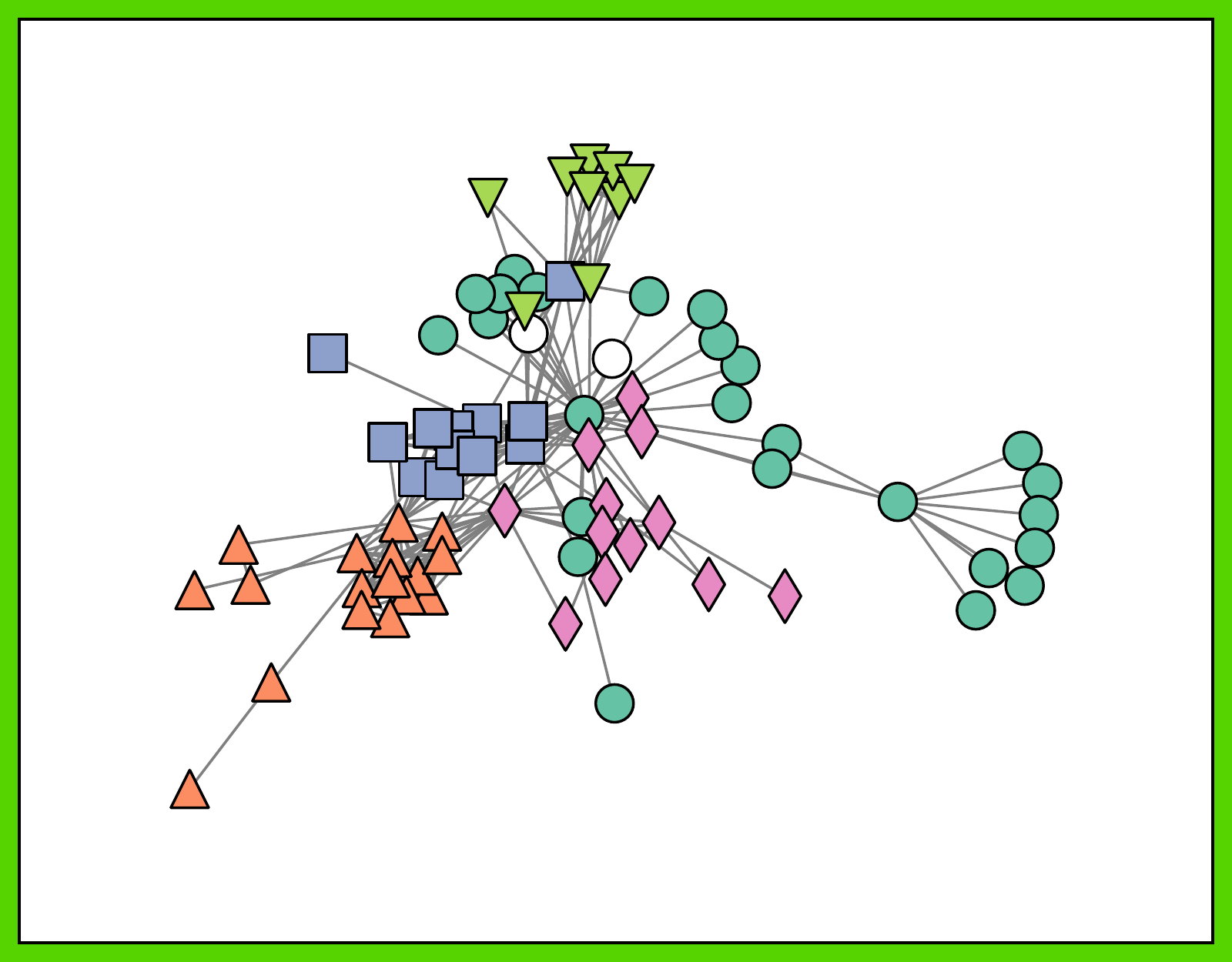} \\
	        \end{tabular}
	        \end{small}
        \end{minipage}
        \hfill

        \caption{Running different standard algorithms on  the ``Les Mis\'erables'' network with 77 nodes~\cite{knuth93}. We show three individual results and the consensus solution. The bi-clustering result does not carbon-copy any of the individual solutions (the horizontal bands in $\transpose{\mat{A}}$), but creates a new one. Also notice on the colored preference matrix (bottom left) how the algorithm ``corrects'' individual solutions (a different color is assigned to each pair $(\set{R}_t, \set{Q}_t)$, see Algorithm~\ref{algo:biclustering}). The proposed consensus algorithm detects two nodes as singletons (in white in the bottom right graph); it is interesting to notice that the base algorithms all differ on how to treat these nodes and assign them to different communities.}
        \label{fig:lesmis}

\end{figure*}

In Figure~\ref{fig:lesmis} we observe in detail how the bi-clustering algorithm selects entries of $\mat{A}$ to create a new solution, preferring regularities in the matrix, while disregarding peculiarities of individual solutions (``inpainted'' nodes do not have a black circle around them in the bottom left graph in Figure~\ref{fig:lesmis}). An important feature is that the proposed algorithm does not blindly select the best solution, but composes a consensual solution from the provided candidates.

In Table~\ref{tab:benchmark} we show our results using a generator of synthetic networks~\cite{lancichinetti2008benchmark}.
For this experiment, we compute the modularity $\operatorname{Mod}(\set{C}_k)$~\cite{newman04} of the solution $\set{C}_k$ provided by the $k$-th base algorithm, and use a smooth increasing nonlinear function of $\operatorname{Mod}(\set{C}_k)$ as the weight for each community $C \in \set{C}_k$. In general, there is no community detection algorithm that ``rules them all'' for every network; however our algorithm consistently performs well in all examples.

\begin{table*}

    \caption{Results with synthetic networks ($m$ is the number of nodes and $\delta$ the average node degree), produced with a standard benchmark generator~\cite{lancichinetti2008benchmark}. There is no single algorithm that produces the best solution for every network; however, the consensus solution is always competitive with the best base solution (in bold). To ``help'' spectral clustering, $K$ was set to the number of ground truth communities. NMI and Mod. stand for normalized mutual information and modularity, respectively.}
    \label{tab:benchmark}

	\centering    
	\begin{small}
	\begin{tabularx}{\textwidth}{ c X *{6}{d{1.3}} >{\columncolor{LightGreen}}d{1.3} }
	\toprule
	&& \multicolumn{1}{c}{\multirow{2}{*}{Louvain}} & \multicolumn{1}{c}{\multirow{2}{*}{Infomap}} & \multicolumn{4}{c}{Spectral clustering} & \multicolumn{1}{c}{\multirow{2}{*}{Consensus}} \\
	\cmidrule(lr){5-8}
	&&&& \multicolumn{1}{c}{$K$} & \multicolumn{1}{c}{$K-1$} & \multicolumn{1}{c}{$K+1$} & \multicolumn{1}{c}{$K+2$} \\
	\midrule
	$m=10^2$ & Mod. &
	\multicolumn{1}{B{1.3}}{0.494} & 0.481 & 0.335 & 0.307 & 0.429 & 0.361 & 0.476 \\
	$\delta=5$ & NMI &
	0.565 & 0.514 & 0.333 & 0.350 & 0.517 & 0.379 & \multicolumn{1}{>{\columncolor{LightGreen}}B{1.3}}{0.584} \\
	\midrule
	$m=10^2$ & Mod. &
	\multicolumn{1}{B{1.3}}{0.436} & \multicolumn{1}{B{1.3}}{0.436} & \multicolumn{1}{B{1.3}}{0.436} & 0.286 & 0.378 & 0.321 & \multicolumn{1}{>{\columncolor{LightGreen}}B{1.3}}{0.436} \\
	$\delta=15$ & NMI &
	\multicolumn{1}{B{1.3}}{1.000} & \multicolumn{1}{B{1.3}}{1.000} & \multicolumn{1}{B{1.3}}{1.000} & 0.612 & 0.851 & 0.774 & \multicolumn{1}{>{\columncolor{LightGreen}}B{1.3}}{1.000} \\
	\midrule
	$m=10^3$ & Mod. &
	\multicolumn{1}{B{1.3}}{0.757} & 0.753 & 0.612 & 0.651 & 0.618 & 0.601 & 0.750 \\
	$\delta=5$ & NMI &
	0.867 & \multicolumn{1}{B{1.3}}{0.886} & 0.782 & 0.813 & 0.806 & 0.793 & 0.885 \\
	\midrule
	$m=10^3$ & Mod. &
	\multicolumn{1}{B{1.3}}{0.759} & 0.759 & 0.665 & 0.662 & 0.662 & 0.634 & 0.759 \\
	$\delta=10$ & NMI &
	0.995 & \multicolumn{1}{B{1.3}}{1.000} & 0.907 & 0.892 & 0.910 & 0.890 & \multicolumn{1}{>{\columncolor{LightGreen}}B{1.3}}{1.000} \\
	\midrule
	$m=10^4$ & Mod. &
	\multicolumn{1}{B{1.3}}{0.765} & \multicolumn{1}{B{1.3}}{0.765} & 0.753 & 0.760 & 0.721 & 0.699 & \multicolumn{1}{>{\columncolor{LightGreen}}B{1.3}}{0.765} \\
	$\delta=300$ & NMI &
	\multicolumn{1}{B{1.3}}{1.000} & \multicolumn{1}{B{1.3}}{1.000} & 0.955 & 0.980 & 0.958 & 0.960 & \multicolumn{1}{>{\columncolor{LightGreen}}B{1.3}}{1.000} \\
	\midrule
	$m=10^4$ & Mod. &
	\multicolumn{1}{B{1.3}}{0.794} & \multicolumn{1}{B{1.3}}{0.794} & 0.783 & 0.785 & 0.791 & 0.784 & 0.794 \\
	$\delta=50$ & NMI &
	\multicolumn{1}{B{1.3}}{1.000} & \multicolumn{1}{B{1.3}}{1.000} & 0.978 & 0.973 & 0.969 & 0.981 & 0.994 \\
	\midrule
	\multicolumn{2}{c}{Average NMI} &
	0.904 & 0.900 & 0.826 & 0.770 & 0.835 & 0.796 & \multicolumn{1}{>{\columncolor{LightGreen}}B{1.3}}{0.911} \\
	\bottomrule
	\end{tabularx}
	\end{small}

\end{table*}

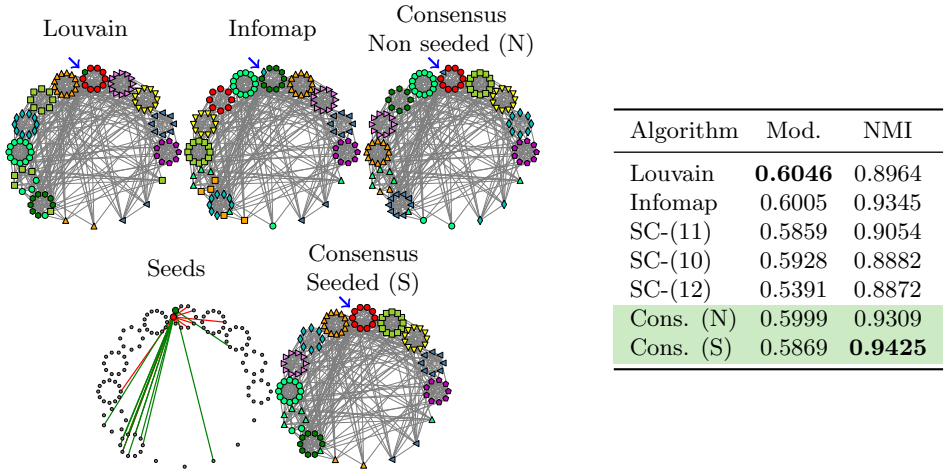
\begin{figure*}

    \begin{small}
    \begin{minipage}[c]{.6\textwidth}
    \centering
    \begin{tabular}{@{\hspace{0pt}} *{4}{c@{\hspace{1pt}}} c@{\hspace{0pt}} }
    \multirow{2}{*}{Louvain} &
    \multirow{2}{*}{Infomap} &
    Consensus \\
    
    &
    &
    Non seeded (N) \\[2pt]
    
    \setlength{\unitlength}{1pt}
    
    \thicklines
    
    \begin{tikzpicture}[overlay]
    \draw [blue, thick, ->] (27pt,65pt) -- ++(4pt,-4pt);
    \draw [blue, thick, ->] (95pt,65pt) -- ++(4pt,-4pt);
    \draw [blue, thick, ->] (163pt,65pt) -- ++(4pt,-4pt);
    \end{tikzpicture}

    \includegraphics[width=.3\textwidth]{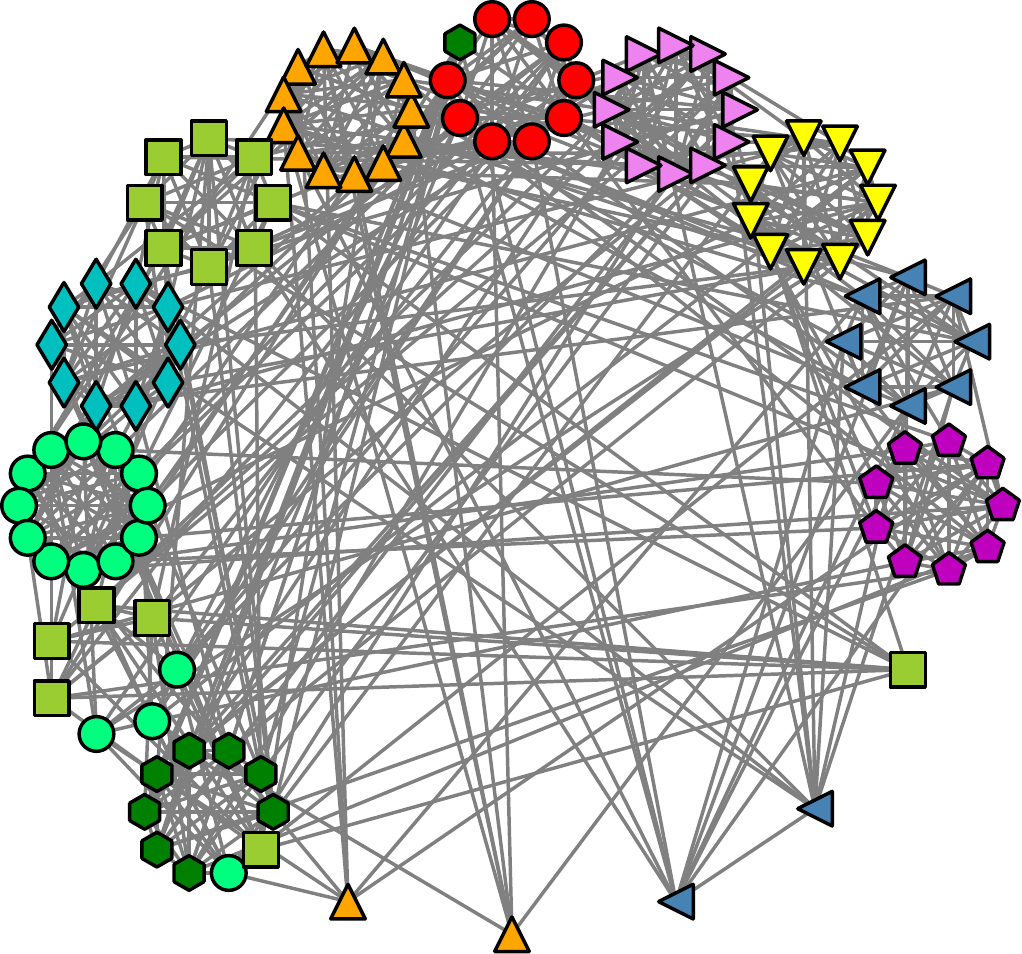} &
    \includegraphics[width=.3\textwidth]{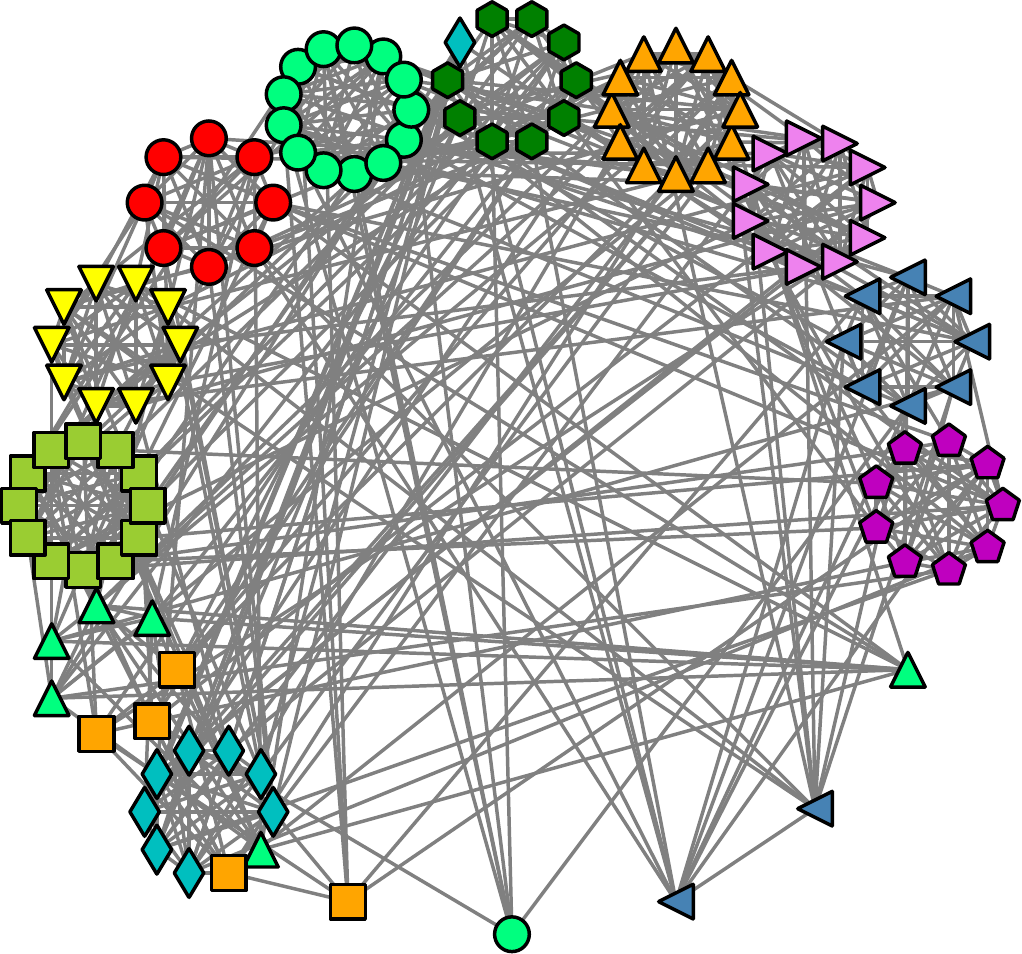} &
    \includegraphics[width=.3\textwidth]{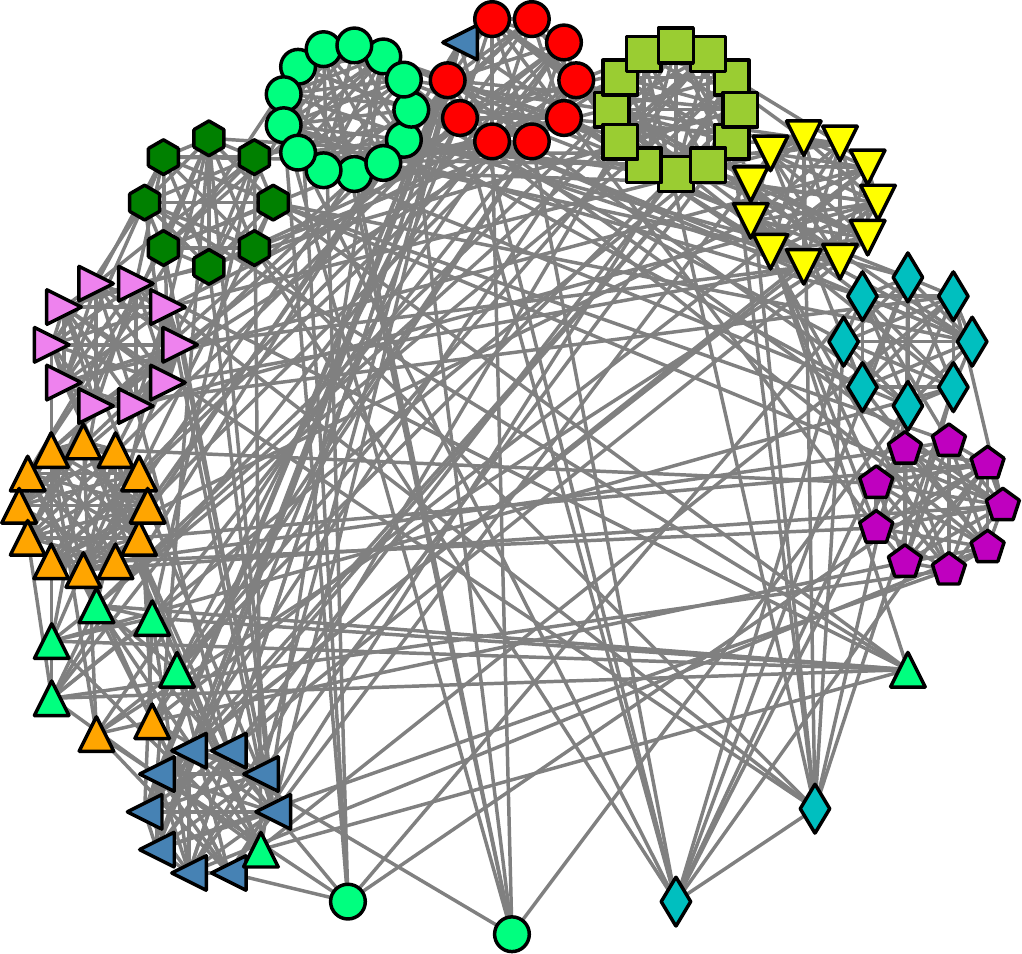} \\
    \end{tabular}
    
    \begin{tabular}{ @{\hspace{0pt}} c @{\hspace{1pt}} c @{\hspace{0pt}} }
    \multirow{2}{*}{Seeds} &
    Consensus  \\
    
    &
    Seeded (S) \\[2pt]
    
    \setlength{\unitlength}{1pt}
    
    \thicklines
    
    \begin{tikzpicture}[overlay]
    \draw [blue, thick, ->] (95pt,65pt) -- ++(4pt,-4pt);
    \end{tikzpicture}

    \includegraphics[width=.3\textwidth]{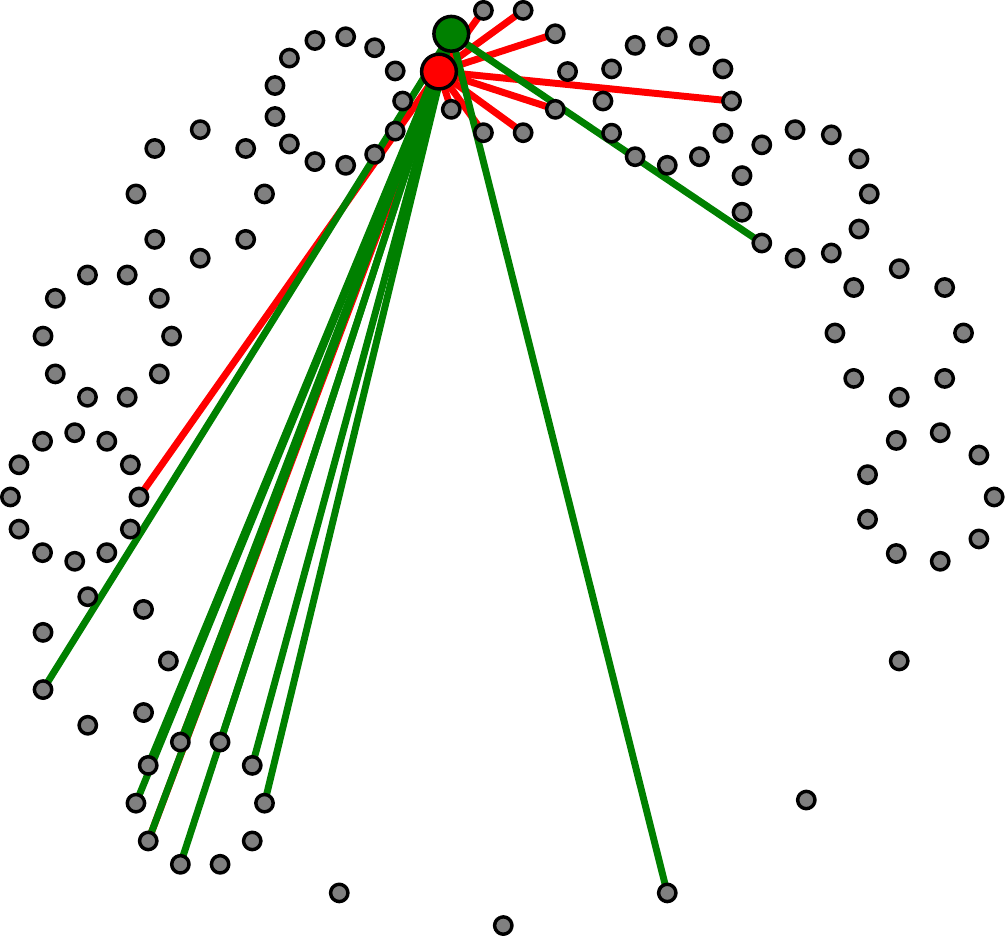} &
    \includegraphics[width=.3\textwidth]{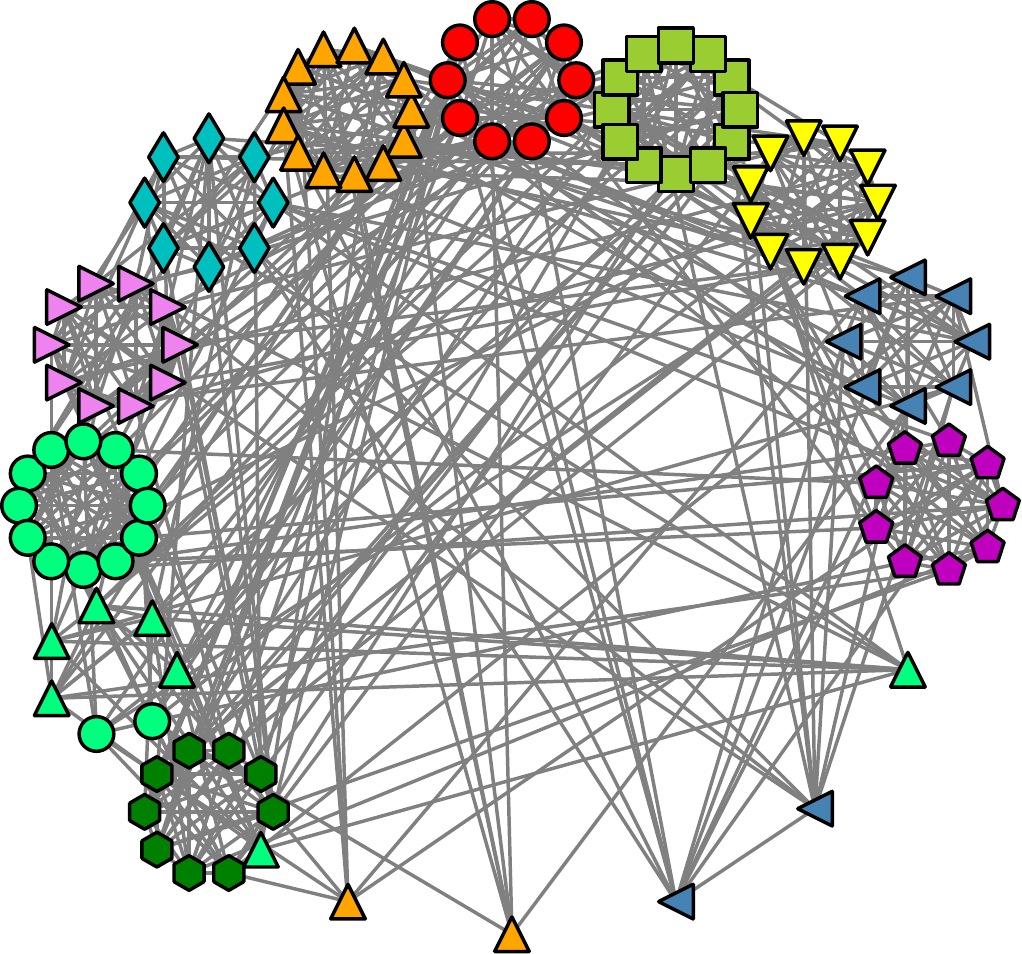} \\
    \end{tabular}
    \end{minipage}\hfill
    \begin{tabular}{l @{\hspace{6pt}} c @{\hspace{6pt}} c}
        \toprule
        Algorithm & Mod. & NMI \\
        \midrule
        Louvain & \textbf{0.6046} & 0.8964\\
        Infomap & 0.6005 & 0.9345 \\
        SC-$(11)$ & 0.5859 & 0.9054 \\
        SC-$(10)$ & 0.5928 & 0.8882 \\
        SC-$(12)$ & 0.5391 & 0.8872 \\
        \rowcolor{LightGreen}
        Cons. (N) & 0.5999 & 0.9309\\
        \rowcolor{LightGreen}
        Cons. (S) & 0.5869 & \textbf{0.9425}\\
        \bottomrule
    \end{tabular}
    \end{small}
    
    \caption{College football network with 115 nodes~\cite{girvan2002community}, representing the matches between different teams. All base methods separate the node (marked with a blue arrow) from its division. By looking at its edges (in green in the 4th graph), we see that this can indeed be correct given the network. Using a little extra information, i.e., forcing the red and green nodes in the 4th graph to be in the same community, corrects this effect. As before, NMI and Mod.~stand for normalized mutual information and modularity, respectively.}
    \label{fig:football}
\end{figure*}

\textbf{Using seeds.} What happens when there is not enough information in the network to recover the ``correct'' structure? The College football network~\cite{girvan2002community} presents a very interesting such example. The network represents the matches played between teams in a season. The teams are organized in divisions, and teams should play more matches with teams from the same division than from different ones. Hence, divisions are considered as ground truth communities for this network. When we run different community detection algorithms on this network, we can observe that one of the divisions is not well recovered by any of them, one of its teams being assigned to a different community, see the blue arrows in Figure~\ref{fig:football}. But in fact, when we observe this team's matches, it did not play against any of the teams in his division! We can add a tiny bit of a priori information by manually adding seeds to $\mat{A}$, i.e., by forcing some nodes (in Figure~\ref{fig:football} the red and green nodes in the 4th graph) to be on the same community. For this, we just modify the corresponding rows of $\mat{A}$ by replacing them by their disjunction (logical or). This simple seeding mechanism is able to correct the ``original mistake.''

\textbf{One algorithm, different networks.}
The proposed approach also allows to combine the results of community structure algorithms that analyze different aspects of a given network (e.g., a network with different modalities or evolving over time).

\begin{figure*}
    \centering
    \begin{small}
    \begin{tabular}{ @{\hspace{0pt}} c @{\hspace{0pt}}c @{\hspace{0pt}} c @{\hspace{0pt}} }
        Consensus &
        Louvain (gender) &
        Louvain (study field) \\
        \includegraphics[width=.33\textwidth]{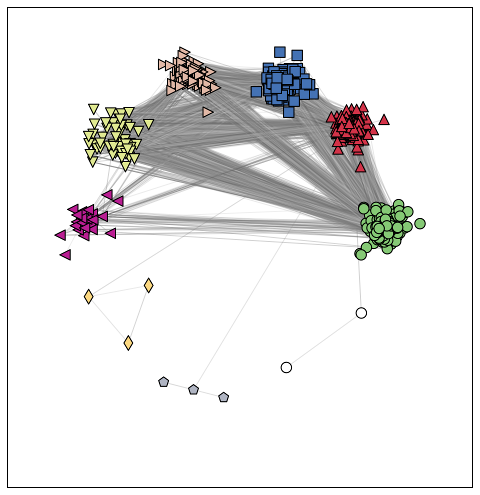} &
        \includegraphics[width=.33\textwidth]{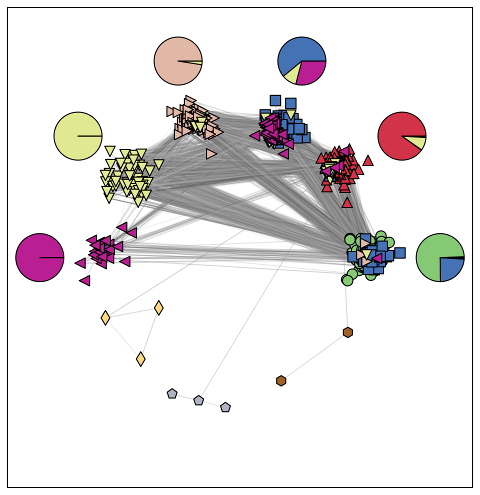} &
        \includegraphics[width=.33\textwidth]{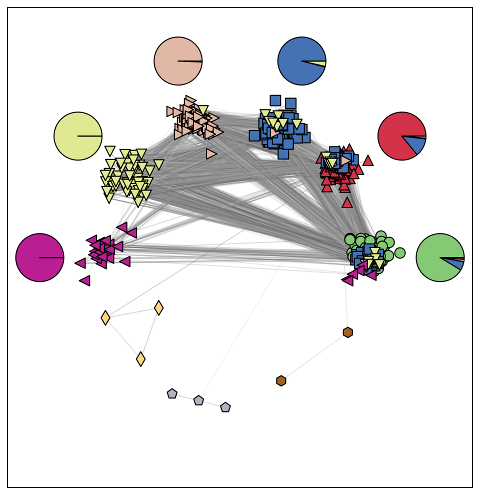} \\
    \end{tabular}
    \end{small}
    \caption{Network of Facebook links between 2006 Duke graduates with 1424 nodes (part of the Facebook 100 dataset \cite{facebook100}). We build two different modalities by assigning weights to the edges according to different node (student) features: field of study and gender. We find a consensus between the two different results of the Louvain algorithm. The pie charts represent the distribution of the nodes of each modality's communities with respect to the consensus.}
    \label{fig:duke}

\end{figure*}

The Facebook 100 dataset presents such an example. The edges represent Facebook friendship but we can also observe several node attributes (e.g., gender, major, minor, dorm, year of graduation). In our particular example, we focus on the 2006 Duke graduates. We build two modalities of this network, by assigning different weights to the edges. In the first one, we use gender information, assigning a weight of 1 if an edge links students of different sex and of 2 otherwise. In the second one, we use study field information, assigning a weight of 1 if the students do not share major nor minor, of 2 if they share major or minor, and of 3 if they share major and minor. We run the Louvain algorithm independently on these two networks and obviously obtain two different community structures, see Figure~\ref{fig:duke}. By running our consensus algorithm on these two results, we produce a solution that aggregates information from both modalities.
The proposed approach allows to perform a coherent cross-modality analysis, something not possible with traditional community detection algorithms. We can therefore expand the analysis to multimodal networks, using standard algorithms (such as Louvain) and without the need to develop new algorithms.

Another interesting example occurs when the network connectivity changes over time or when different modalities exhibit different edge sets. 
This is important for example when the graph is obtained through inference, because differences and/or errors in the inference process might yield different connectivities.
We simulate such an example by taking a network and building 10 perturbed copies, randomly reassigning a subset of its edges. We then run Infomap on each copy and compare the result in terms of NMI with the Infomap result on the original network (Table~\ref{tab:noise}). When a small portion of edges is perturbed, the best individual solution is still good, because there is a non-negligible chance that one of the perturbations does not alter the community structure of the original network. This is no longer true as the number of perturbed edges increases. Our algorithm is able to balance out the peculiarities of the perturbed solutions, obtaining a solution much closer to the original one and hence more resilient to perturbations.

\begin{table}
    \caption{Results of perturbing the edge set of the US politics books network with 105 nodes (\url{http://www.orgnet.com/divided.html}): $\rho\, |E|$ edges are exchanged at random. We provide average NMI values across 1000 trials, using 10 perturbed networks per trial. The result of Infomap in the original network is considered the ground truth. The consensus solution outperforms all individual ones, with the performance gap increasing as more edges are perturbed.}
    \label{tab:noise}

    \centering
    \begin{tabular}{d{1.2}cc>{\columncolor{LightGreen}}c}
%        \toprule
        \hline
        \multicolumn{1}{c}{$\rho$} & Best   & Median   & \multicolumn{1}{c}{Consensus} \\
%        \midrule
        \hline
        0.1 & 0.9224 & 0.8388 & \textbf{0.9258} \\
        0.25 & 0.8004 & 0.6944 & \textbf{0.8551} \\
%                & std   & 0.0406 & 0.0254 & 0.0379 \\
        0.3   & 0.7428 & 0.6315 & \textbf{0.8244} \\
%                & std   & 0.0447 & 0.0274 & 0.0394 \\
        0.35   & 0.6801 & 0.5581 & \textbf{0.7933} \\
%        & std   & 0.047 & 0.0292 & 0.0417 \\
%        \bottomrule
        \hline
    \end{tabular}
    
\end{table}
% !TEX root = main.tex
\section{Multiple parametric model estimation}
\label{sec:mpme}

This section addresses the problem of fitting multiple instances of a parametric model to data corrupted by noise and outliers, formally connecting it with bi-clustering and consensus.
In our context, the data is a set $\set{X}$ of $m$ geometric objects, described by
$\set{X} = \left( \bigcup_i \set{X}_i \right) \cup \set{O}$, with $(\forall i) \set{X}_i \cap \set{O} = \emptyset$, and $\set{O}$ being once again outliers as detailed next.
The objects in each subset $\set{X}_i$, which might intersect, are (noisy) measurements that can be described with a parametric model $\mu(\theta^{(i)})$, where $\theta^{(i)}$ is a parameter vector. In the following, we say that the objects in $\set{X}_i$ are inliers to the model $\mu(\theta^{(i)})$. We also generally refer to a set of objects as inliers, in the sense that it exists a statistically meaningful model that describes them. The objects in $\set{O}$ cannot be described with any of the models $\mu(\theta^{(i)})$, and we refer to them as outliers. 

Let us define more formally these intuitive concepts.
A model $\mu$ is defined as the zero level set of a smooth parametric function $f_\mu (x; \theta)$,
\begin{equation}
    \mu(\theta) = \{ \vect{x} \in \Real^d ,\ f_\mu (\vect{x}; \theta) = 0 \} ,
    \label{eq:model}
\end{equation}
where $\theta$ is a parameter vector.
We define the error associated with the datum $\vect{x} \in \set{X}$ with respect to the model $\mu(\theta)$ as
\begin{equation}
    \operatorname{e}_\mu (\vect{x}, \theta) = \min_{\vect{x}' \in \mu(\theta)} \operatorname{dist} (\vect{x}, \vect{x}') ,
    \label{eq:pointModelDistance}
\end{equation}
where $\operatorname{dist}$ is an appropriate distance function.
Using this error metric, we define the Consensus Set (CS) of a model as
\begin{equation}
    \set{C}(\theta) = \{ \vect{x} \in \set{X} ,\ \operatorname{e}_\mu (\vect{x}, \theta) \leq \delta \} ,
    \label{eq:consensusSet}
\end{equation}
where $\delta$ is a threshold that accounts for the measurement noise.

The goal of this section is to find the set of inliers-model pairs $\{ (\set{X}_i, \theta^{(i)}) \}$ from the observed $\set{X}$ such that $\set{X}_i=\set{C}(\theta^{(i)})$. This problem is, by itself, ill-posed. It is standard in the literature to implicitly or explicitly impose a penalty on the number of recovered pairs to render it tractable. We also adopt such a design choice.
Notice that once $\set{X}_i$ is found, the corresponding $\theta^{(i)}$ can be estimated for example by simple least-squares regression, i.e.,
\begin{equation}
    \hat{\theta}^{(i)} = \argmin_{\theta}  \sum_{\vect{x} \in \set{X}_i} \big[ \operatorname{e}_\mu (\vect{x}, \theta) \big]^2 .
    \label{eq:leastSquares}
\end{equation}
This is an important but difficult problem, as standard robust estimators, like RANSAC (RANdom SAmple Consensus)~\cite{choi2009,ransac}, are designed to extract a single model.  
Let us then begin by formally explaining how does the RANSAC machinery work, to further illustrate the value and perspective of our proposed multi-model formulation.

Let us denote by $b$ the minimum number of elements necessary to uniquely characterize a given parametric model, e.g., $b=2$ for lines and $b=3$ for circles. For example, if we want to discover alignments in a 2D point cloud, the elements are 2D points, models $\mu$ are lines, and $b=2$, since a line is defined by two points. A set of $b$ objects is called a minimal sample set (MSS).
RANSAC randomly samples $n$ MSSs, each generating a model hypothesis. The number $n$ is an overestimation of the number of trials needed to obtain a certain number of ``good'' models~\cite{ransac,toldo08,zuliani2005}.
Then RANSAC computes the CS of each model hypothesis using Equation~(\refeq{eq:consensusSet}). Algorithm~\ref{algo:ransac} describes the standard RANSAC procedure.

Applying RANSAC sequentially, removing the inliers from the dataset as each model instance is detected, has been proposed as a solution for multi-model estimation, e.g.,~\cite{rabin2010macransac}. However, this approach is known to be suboptimal~\cite{zuliani2005}. The multiRANSAC algorithm~\cite{zuliani2005} provides a more effective alternative, although the number of models must be known a priori, imposing a very limiting constraint in many applications.
An alternative approach consists of finding modes in a parameter space. The overall idea is that we can map the data into the parameter space through random sampling, and then seek the modes of the distribution by discretizing the distribution, i.e., using the Randomized Hough Transform~\cite{xu1990}, or by using non-parametric density estimation techniques, like the mean-shift clustering algorithm~\cite{subbarao09}. These, however, are not intrinsically robust techniques, even if they can be robustified with outliers rejection heuristics~\cite{toldo08}. Moreover, the choice of the parametrization is critical, among other important shortcomings~\cite{toldo08}.
The computational cost of these techniques can be very high as well.

\begin{algorithm2e}[t]
	\SetKwInput{Input}{input}
	
	\Input{set of objects $\set{X}$, parametric function $f_\mu$.}
	
	Set $b$ to the minimum number of elements necessary to uniquely characterize model $\mu$, see Equation~(\ref{eq:model})\;
	
	\ForEach{$k \in \{ 1 \dots n \}$}{
		Sample at random a minimal sample set (MSS) $\set{X}_{\textsc{mms}}$ of size $b$ from $\set{X}$\; \nllabel{algo:ransac_mss}
		Estimate $\theta^{(k)}$ from $\set{X}_{\textsc{mms}}$ by solving $(\forall \vect{x} \in \set{X}_{\textsc{mms}})\ f_\mu (\vect{x}; \theta) = 0$\;
		Check that $\mu(\theta^{(k)})$ is not a degenerate model, otherwise go to line~\ref{algo:ransac_mss}\;
		Compute $\set{C} \left( \theta^{(k)} \right)$\;
	}
	$k_{\text{max}} = \argmax_{k} \left| \set{C} \left( \theta^{(k)} \right) \right|$\;
	Estimate $\theta^{(k_{\text{max}})}$ from $\set{C} \left( \theta^{(k_{\text{max}})} \right)$ using least-squares\;
	\Return $\left( \set{C} \left( \theta^{(k_{\text{max}})} \right), \theta^{(k_{\text{max}})} \right)$\;
	
	\caption{RANSAC}
	\label{algo:ransac}
\end{algorithm2e}

All these techniques share a common high-level conceptual approach to model estimation: in order to solve the problem, objects are clustered. In this work, we propose an alternative formulation that involves as before \emph{bi-clustering} the objects \emph{and} the models generated with, e.g., RANSAC.
First of all, the proposed modeling does not impose non-intersecting subsets $\set{X}_i$.
Secondly, it exploits consistencies that naturally arise during the RANSAC execution, while explicitly avoiding spurious inconsistencies.
This new formulation conceptually changes the way that the data produced by the popular RANSAC, or related model-candidate generation techniques, is analyzed.

\subsection{Multiple model estimation by analyzing the preference matrix}
\label{sec:mme}

The information generated throughout the RANSAC iterations, i.e., the link between objects and model hypotheses, can be represented with the same $m \times n$ preference matrix $\mat{A}$ considered before, whose rows and columns represent objects and models, respectively; the element $(\mat{A})_{ij} = 1$ if the $i$-th object is in the CS of the $j$-th model, and $0$ otherwise.

Traditionally and as in clustering, in algorithms like RANSAC, the preference matrix is (often implicitly) analyzed column-by-column or row-by-row.
For example, Toldo and Fusiello~\cite{toldo08} proposed to cluster the objects in $\set{X}$ using the rows of $\mat{A}$ as feature vectors, obtaining a powerful state-of-the-art clustering-based technique for multiple model estimation, called J-linkage.
For this, they use a tailored agglomerative clustering algorithm. As all agglomerative clustering algorithms, J-linkage proceeds in a bottom-up manner: starting from all singletons, each iteration of the algorithm merges the two clusters with the smallest distance.
J-linkage uses the Jaccard distance~\cite{toldo08}, and the features are updated during the merging process. Each cluster's feature is computed as the intersection of the features of its objects (i.e., the logical conjunction of the corresponding rows of $\mat{A}$).

Although a clustering of objects, using as features the set of sampled models they belong to, might lead to good results in certain applications, it does not fully address the problem at hand. The relationship of the objects with the sampled models is the actual focus of interest (notice again the block-diagonal structure of $\mat{A}$ in Figure~\ref{fig:preferenceMatrix}).
A pattern-discovery algorithm is needed in this relationship space.
We are thus interested in finding clusters in the product set $\set{X} \times \set{M}$, where $\set{M} = \{ \set{C}(\theta^{(i)}) \}_{1 \leq i \leq n}$ is the set of sampled models. As already mentioned in Section~\ref{sec:biclustering}, such a problem is known in the literature as bi-clustering.

We thus propose to address the problem of model estimation by bi-clustering the matrix $\mat{A}$, using the algorithm presented in Section~\ref{sec:biclustering}.
A further conceptual advantage is that an object is allowed to belong to multiple bi-clusters, because we are partitioning the product set $\set{X} \times \set{M}$ instead of $\set{X}$. This kind of situation occurs very frequently, as observed in Figure~\ref{fig:preferenceMatrix}: if the lines (models) intersect (left), then they share points (objects); this is translated as elements outside the block-diagonal structure of $\mat{A}$ (right). RANSAC-related techniques arbitrarily assign shared objects to a given model.

\subsection{Cleansing the preference matrix}
\label{sec:nfa}

The standard random sampling approach to multiple model estimation generates many good model instances (composed of inliers), but also generates many bad models (composed mostly of outliers). In general, the number of bad models exceeds by far the number of good ones. It is not worth devoting computational effort in the analysis of these columns of $\mat{A}$.
Any pattern-discovery technique, such as the bi-clustering approach presented in the previous section, would benefit from having a simple, efficient, and statistically meaningful method for discarding the bad models.
These models will typically contain only a handful of objects. The question is how do we determine the minimum size of a good consensus set?
This important computational contribution is addressed next, following the a contrario testing mechanism presented in depth in~\cite{desolneux08}.

Let us assume that we have a set of $m$ (random) objects.
We are interested in computing the probability that a model $\mu(\theta)$ has an associated consensus set $\set{C}(\theta)$ of at least $k=|\set{C}(\theta)|$ objects.
Under the simplifying assumption that all objects are i.i.d., the probability of such an event is
$\bintail{m}{k}{p}$,
where $\mathcal{B}$ is the binomial tail and $p=p(\mu(\theta),\delta)$ is the probability that a random object belongs to the consensus set $\set{C}(\theta)$, built with noise level $\delta$. We will later provide details on how to compute $p$ for several different scenarios (Appendix~\ref{sec:probaComputation}).

Let then $\mu(\theta)$ be a model and $\set{C}(\theta)$ its associated consensus set, obtained with precision parameter $\delta$.
The model $\mu(\theta)$ is said to be \meps-meaningful if (NFA stands for Number of False Alarms)
\begin{equation}
    \operatorname{NFA}(\mu(\theta)) = N_\text{tests} \ \bintail{m}{|\set{C}(\theta)|}{p} < \eps,
    \label{eq:nfa}
\end{equation}
where as mentioned above $p = p(\mu(\theta),\delta)$ takes different forms for different models, and $N_\text{tests} = \binom{m}{b}$ represents the total number of considered models. Recall that $b$ is the minimum number of elements necessary to uniquely characterize a given parametric model.
It is easy to prove, by the linearity of expectation, that the expected number of \meps-meaningful models in a finite set of random models is smaller than \meps.
Alternatively, $N_\text{tests}$ can be empirically set by analyzing a training dataset~\cite{burrus2009pr}, providing a tighter bound for the expectation.

Equation~(\ref{eq:nfa}) provides a formal probabilistic method for testing if a model is likely to happen at random or not. From a statistical viewpoint, the method goes back to multiple hypothesis testing.
Following an a contrario reasoning~\cite{desolneux08}, we decide whether the event of interest has occurred if it has a very low probability of occurring by chance in the above defined random (background) model. In other words, a model $\mu(\theta)$ is \meps-meaningful if $|\set{C}(\theta)|$ is sufficiently large to have $\operatorname{NFA}(\mu(\theta)) < \eps$. Only \meps-meaningful models are kept in $\mat{A}$.

Notice that we are in a sense using the a contrario validation procedure backwards: instead of using it to detect good models, we use it to eliminate bad models. We do not need a very sharp event-detection procedure in order to only keep good models; we only need a statistical test to eliminate the vast majority of clearly poor models.
Hence, the value of \meps is not critical, our model being inherently robust to poor models, as shown in Section~\ref{sec:biclustering}. As any statistical test that controls false positives, our a contrario tests do not provide a good control of false negatives (i.e., missed detections). We can thus easily relax the value of $\eps=1$, allowing for some quantity of poor models in our cleansed preference matrix, but being sure that we do not miss good ones.

As a result of this statistical validation procedure, the preference matrix $\mat{A}$ is considerably shrunk. Many unuseful columns are eliminated (be observed that only about 10\% of the original columns are kept in our experiments). Due to this elimination, some rows might also become zero-valued and can also be eliminated. This shrunk preference matrix is fed to the bi-clustering algorithm (Algorithm~\ref{algo:biclustering}), gaining in stability of the results as well as in speed.

\subsection{An algorithm for multiple parametric model estimation}
When we presented Algorithm~\ref{algo:biclustering}, we claimed that tighter bounds could be computed for $\tau_R$ in the case of parametric models. The tests used for cleansing the preference matrix are readily available for this task. For each bi-cluster, we can compute its NFA, where the number of elements in the group if the number of rows in the bi-cluster. Finally, we only keep those bi-clusters that are \meps-meaningful. The resulting complete algorithm is as follows:

\begin{enumerate}
	\item Compute the preference matrix $\mat{A}$ by randomly sampling minimal sampled sets and computing their corresponding consensus sets (Section~\ref{sec:mme}).
	\item Cleanse $\mat{A}$ by discarding the columns that do not satisfy the statistical test presented in Section~\ref{sec:nfa}. For these tests, we relax the value of \meps.
	\item Bi-cluster the cleansed version of $\mat{A}$ using Algorithm~\ref{algo:biclustering}.
	\item Discard the bi-clusters that are not \meps-meaningful, see Section~\ref{sec:nfa}. For these tests, we fix $\eps = 1$. We also ask that each bi-cluster is meaningful when we only consider the elements that do not belong to other bi-clusters. This last step ensures that each bi-cluster contains some points that only belong to it and is a mild version of the exclusion principle~\cite{desolneux08}.
\end{enumerate}
We next present many experimental results and different applications that make use of this general algorithm for multiple model estimation.

\subsection{Experimental results}

Each minimal sample set (MSS) is built using non-uniform random sampling~\cite{zuliani2005}, such that if an object $\vect{x}$ has already been chosen, $\vect{y} \neq \vect{x}$ is selected with probability
\begin{equation}
    \Pr (\vect{y} | \vect{x}) = 
        z^{-1} \exp \left( - \sigma^{-2} \operatorname{dist} \left( \vect{x},\vect{y} \right)^2 \right) ,
        \label{eq:distanceDistribution}
\end{equation}
where $z$ is a normalization constant and $\sigma$ is a parameter of the algorithm. Depending on the application at hand, the distance $\operatorname{dist}$ takes different forms (specified in Appendix~\ref{sec:probaComputation}).

\begin{figure*}[p]
    \centering
    
    \begin{subfigure}[b]{\textwidth}
    	\centering
	\begin{small}
	\begin{tabular}{ @{\hspace{0pt}} m{.01\textwidth} @{\hspace{2pt}} m{.238\textwidth} @{\hspace{4pt}} m{.238\textwidth} @{\hspace{4pt}} m{.238\textwidth} @{\hspace{4pt}} m{.238\textwidth} @{\hspace{0pt}} }
	
		&
		\multicolumn{2}{c}{J-linkage result} &
		\multicolumn{2}{c}{Bi-clustering result} \\
		\cmidrule(lr){2-3}
		\cmidrule(lr){4-5}
		& \multicolumn{1}{c}{Point assignments} & \multicolumn{1}{c}{Cluster sizes}
		& \multicolumn{1}{c}{Point assignments} & \multicolumn{1}{c}{Model assignments} \\
	
		A &
		\includegraphics[width=.238\textwidth]{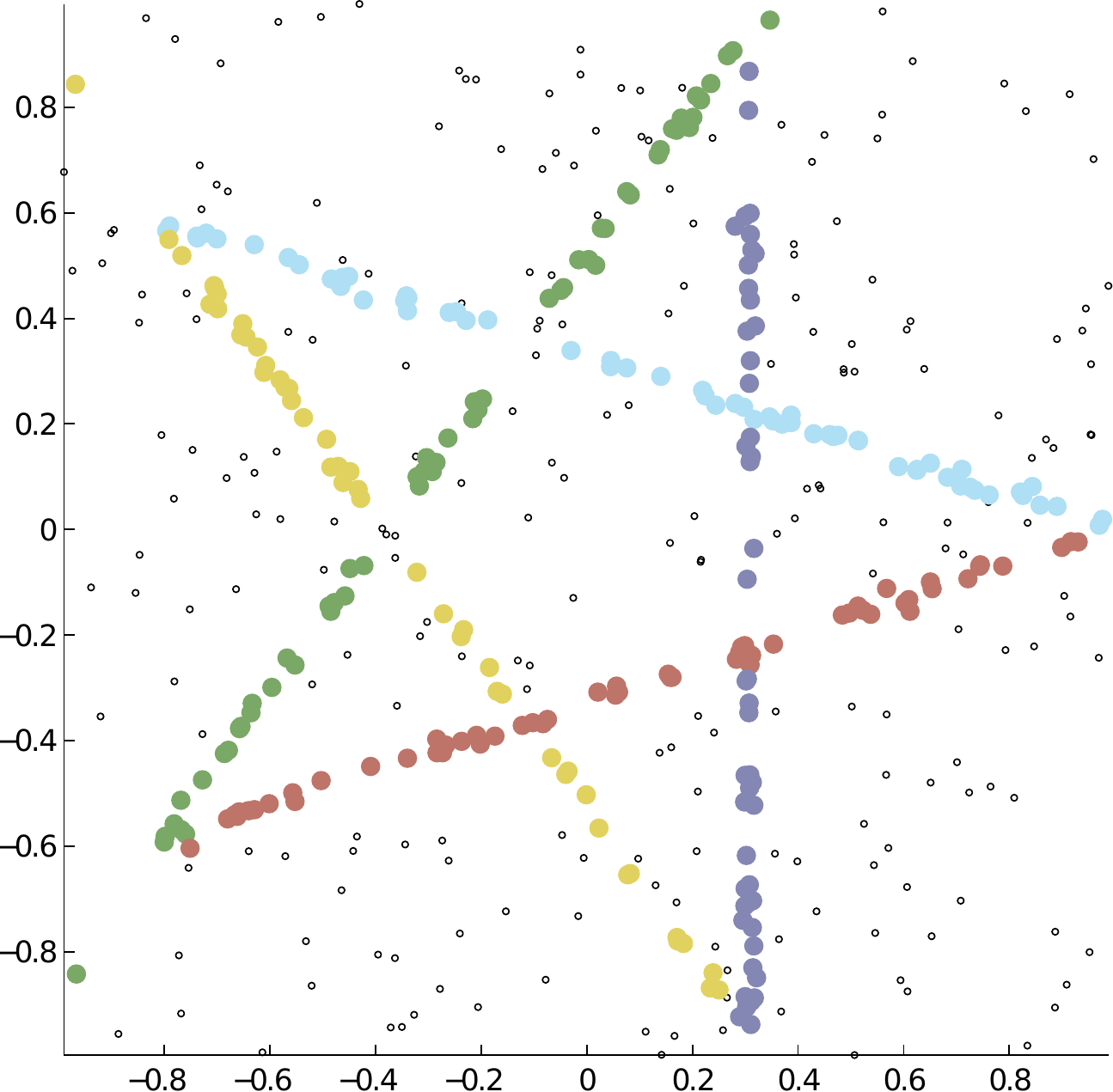} &
		\includegraphics[width=.238\textwidth]{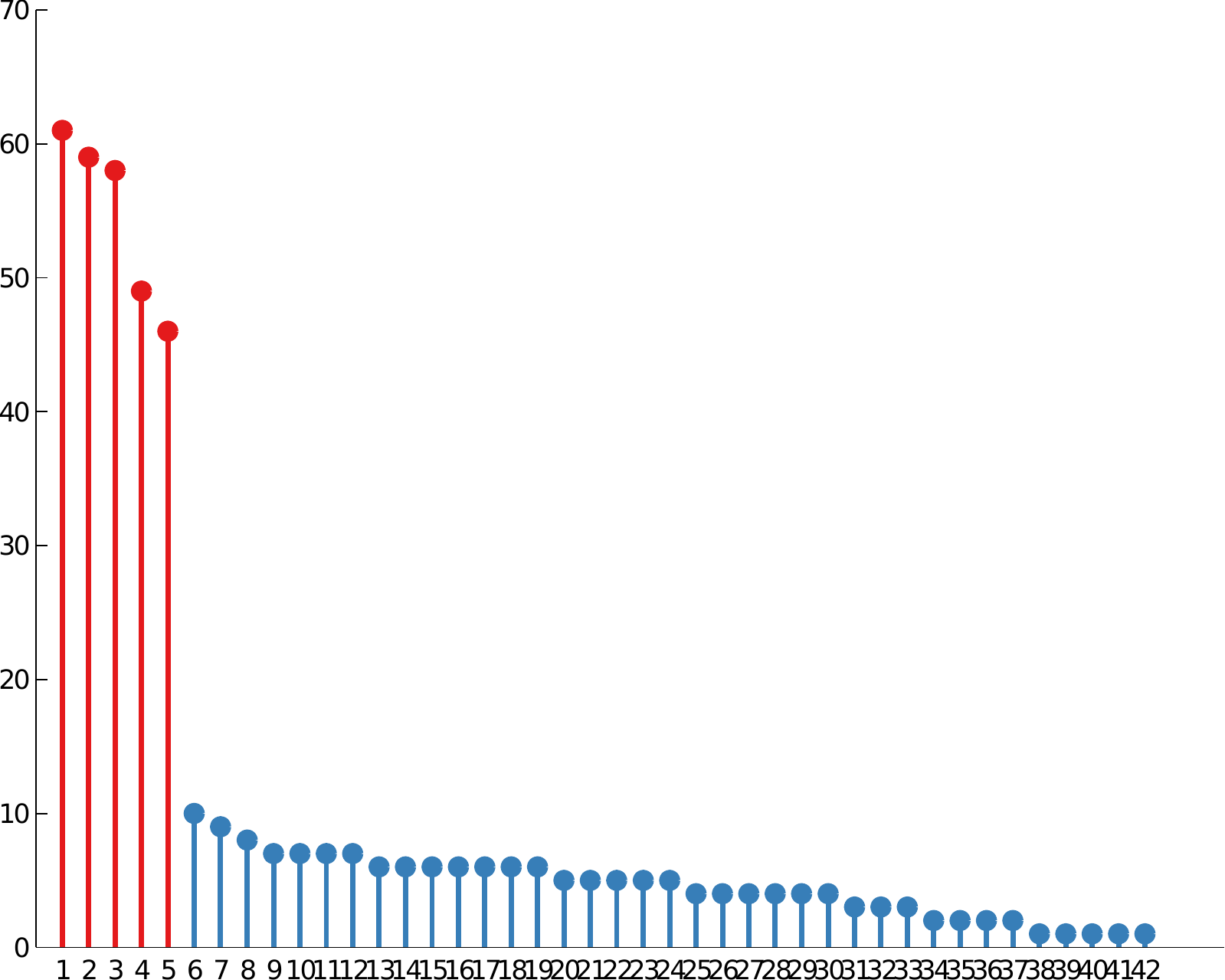} &
		\includegraphics[width=.238\textwidth]{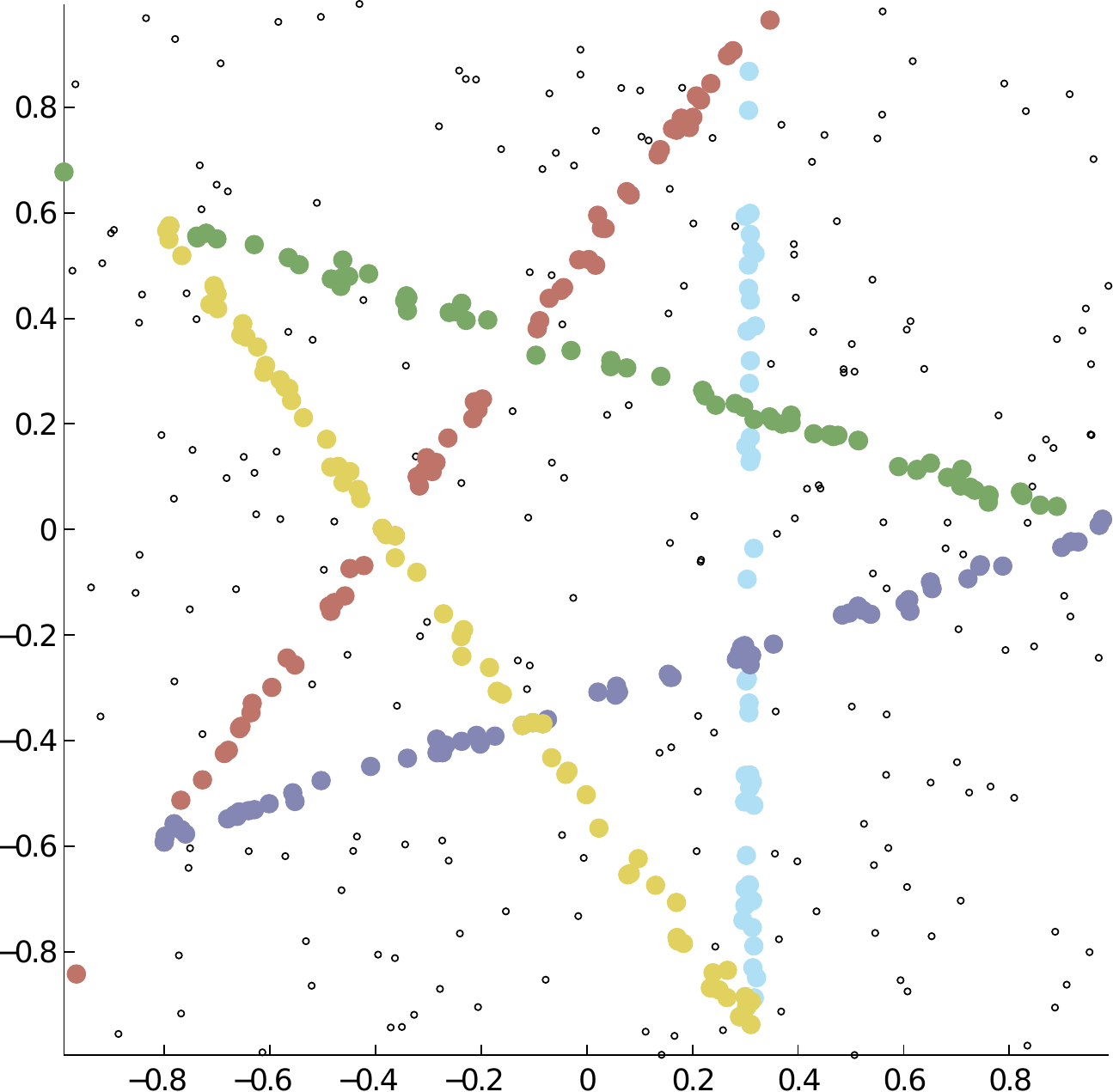} &
		\includegraphics[width=.238\textwidth]{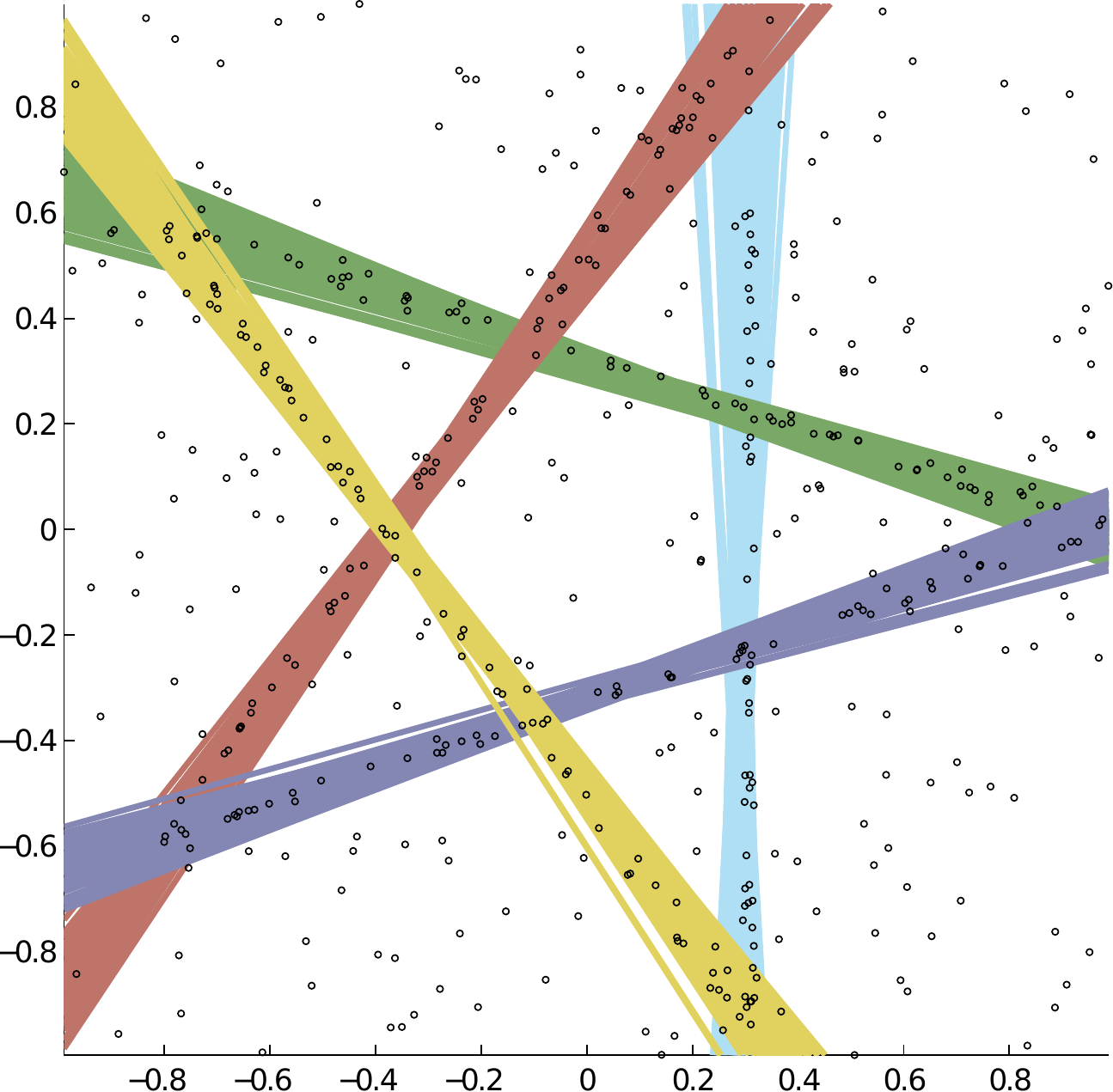} \\
	
		B &
		\includegraphics[width=.238\textwidth]{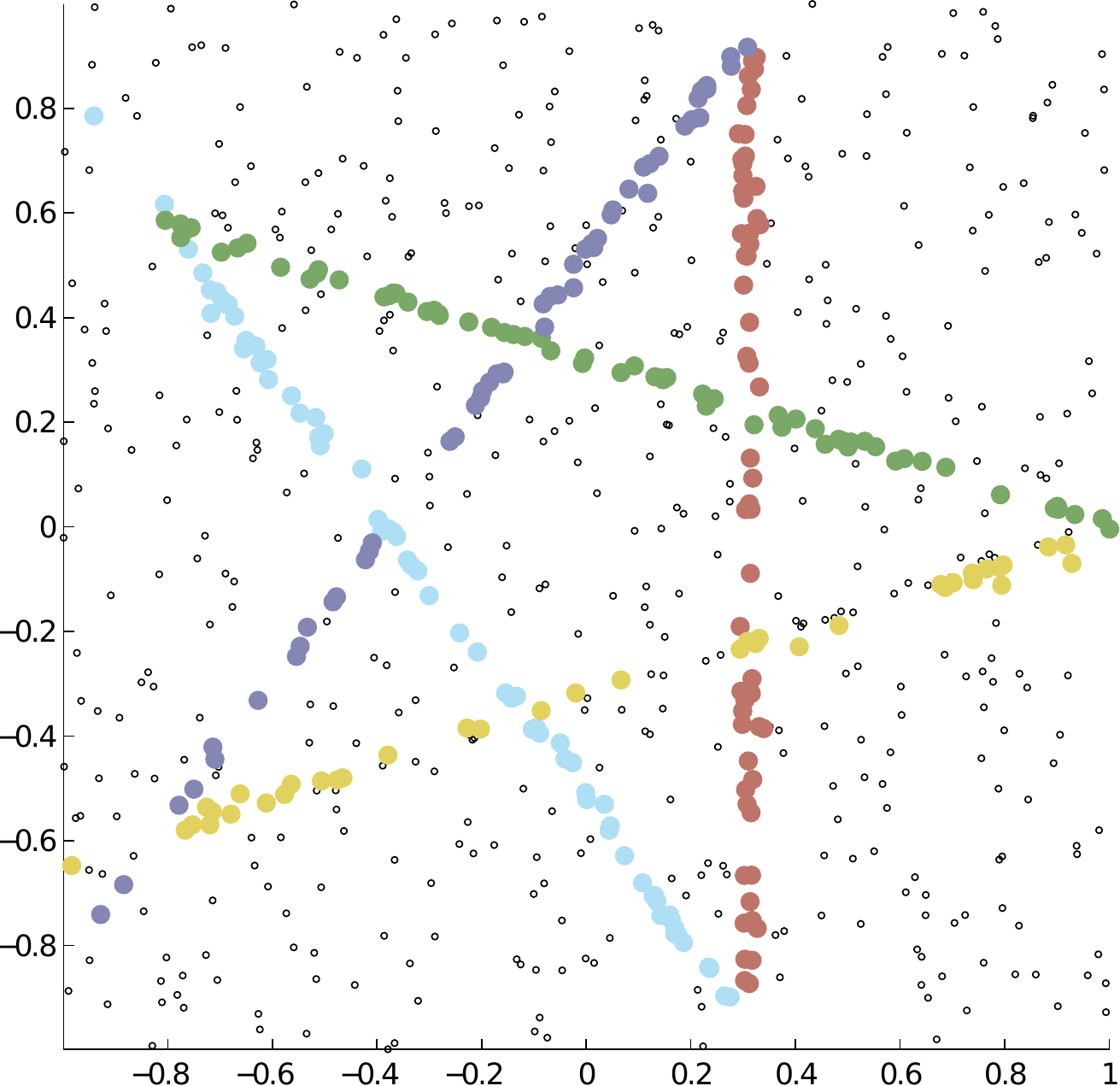} &
		\includegraphics[width=.238\textwidth]{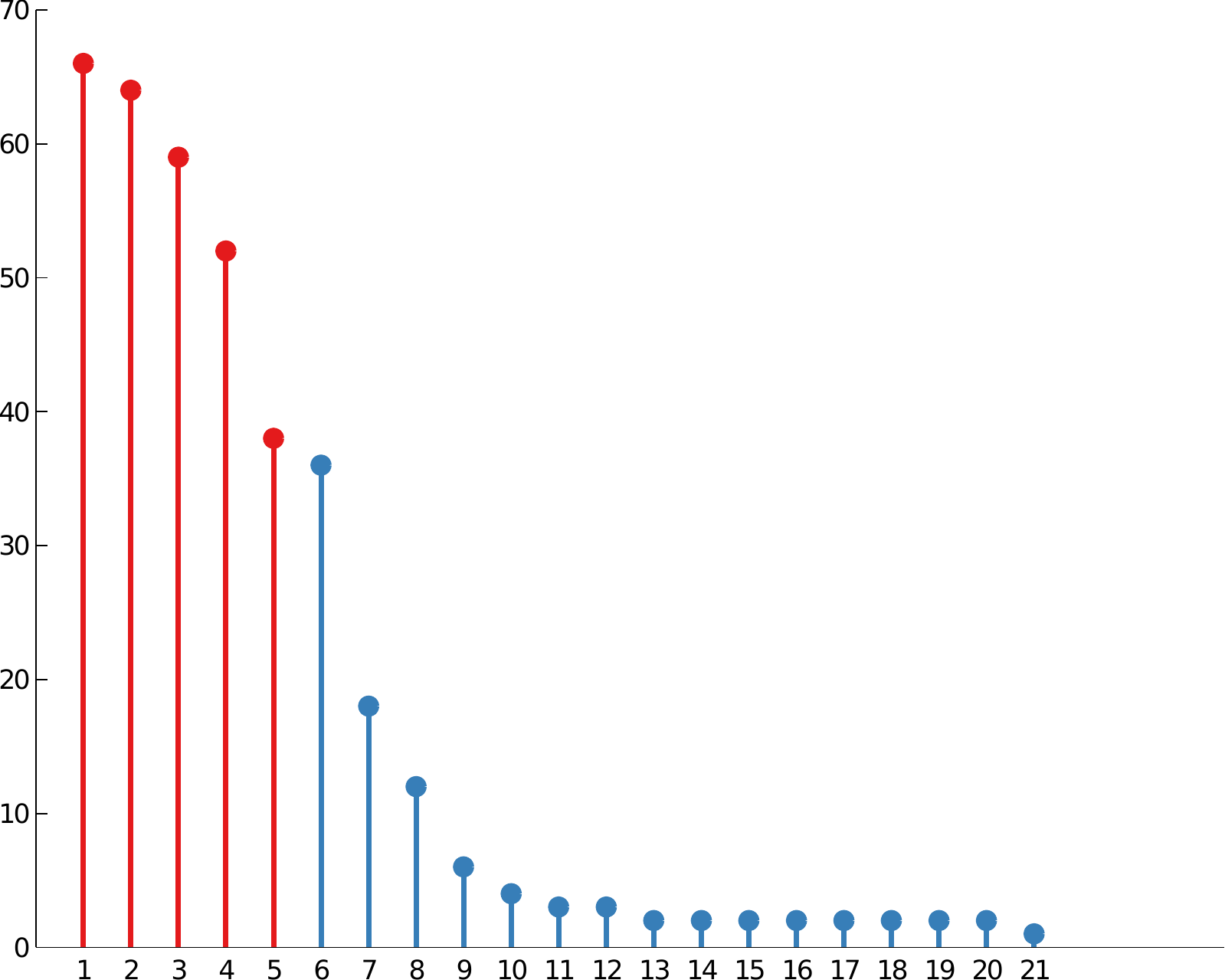} &
		\includegraphics[width=.238\textwidth]{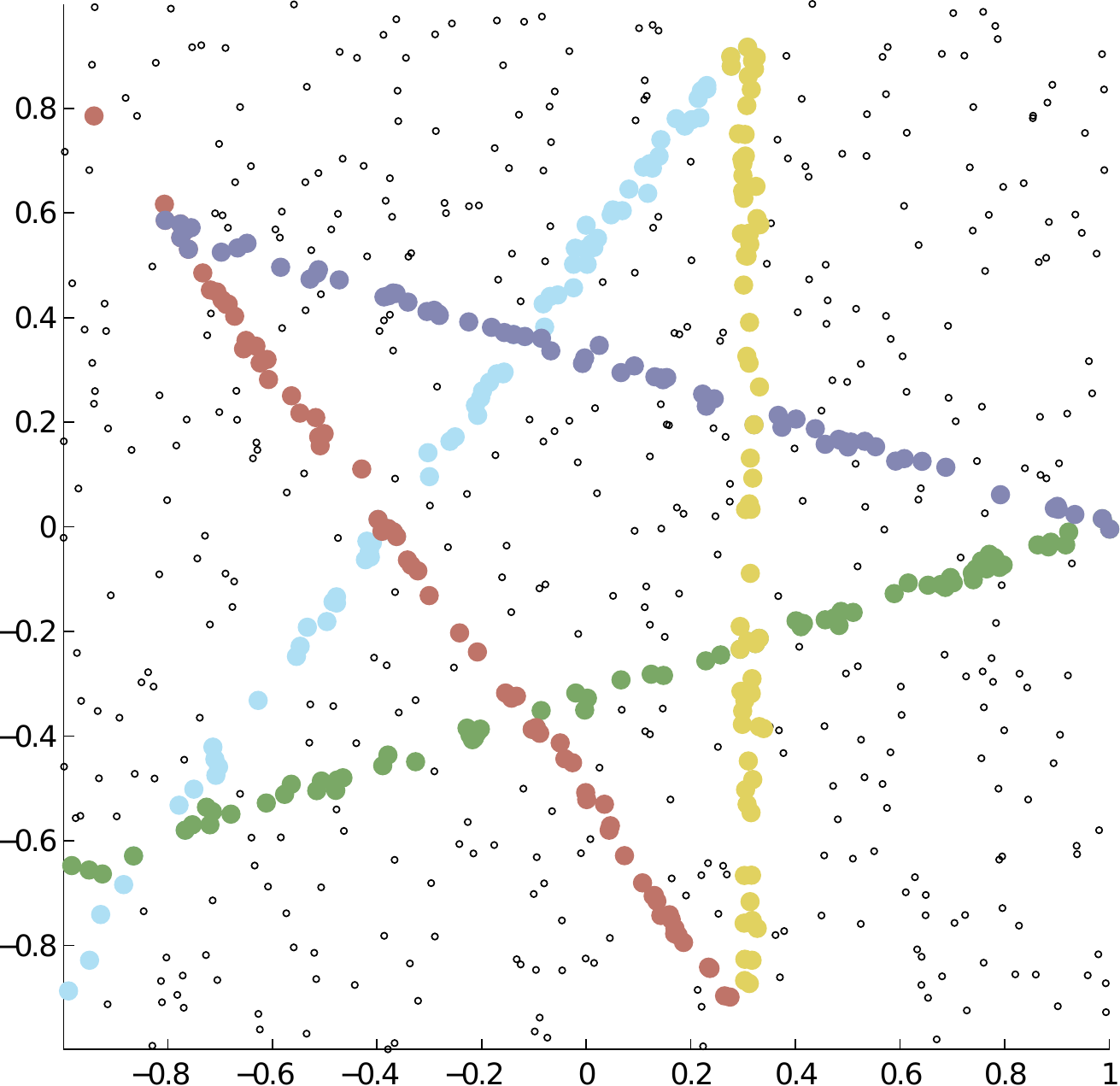} &
		\includegraphics[width=.238\textwidth]{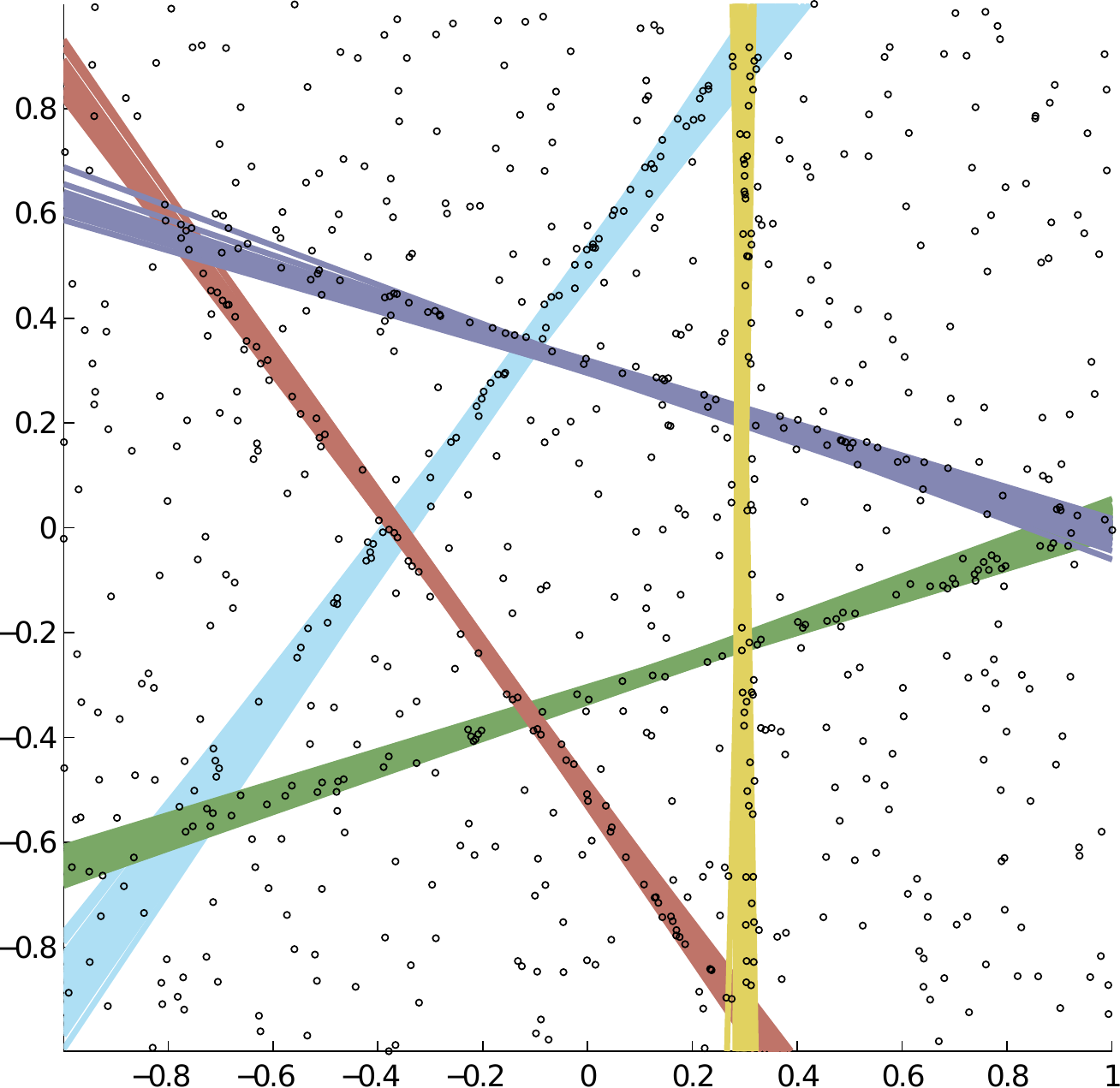} \\

		C &
		\includegraphics[width=.238\textwidth]{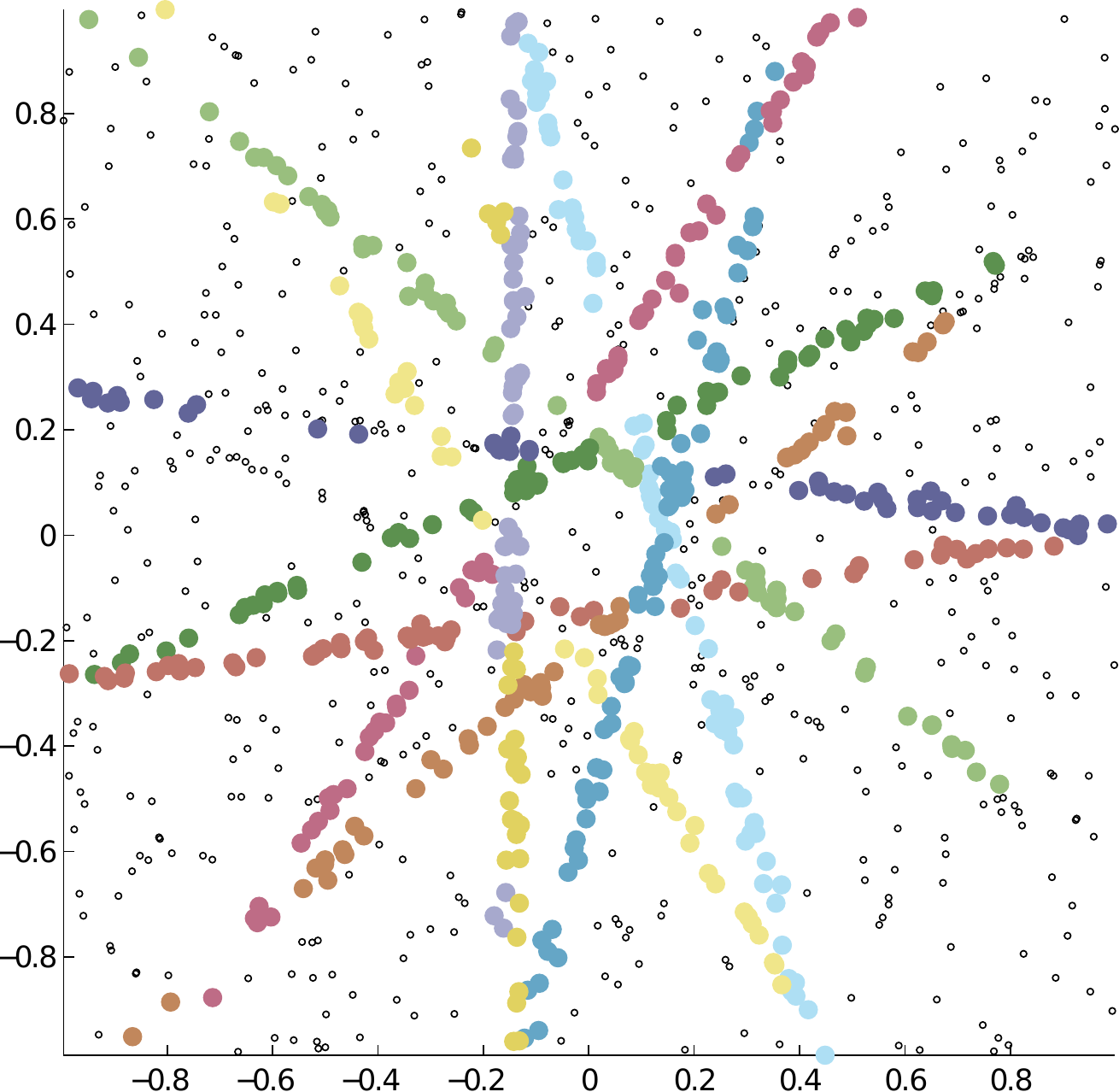} &
		\includegraphics[width=.238\textwidth]{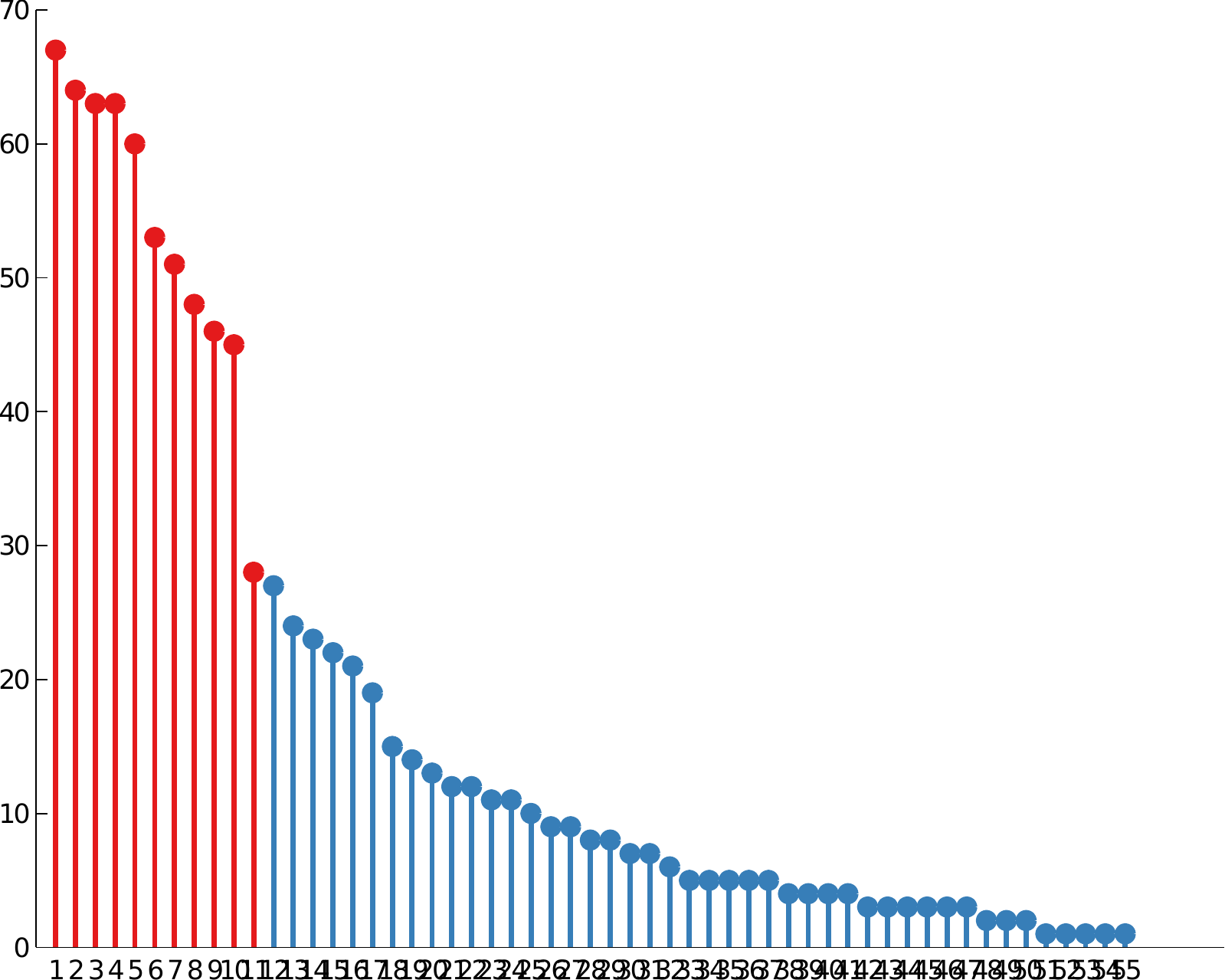} &
		\includegraphics[width=.238\textwidth]{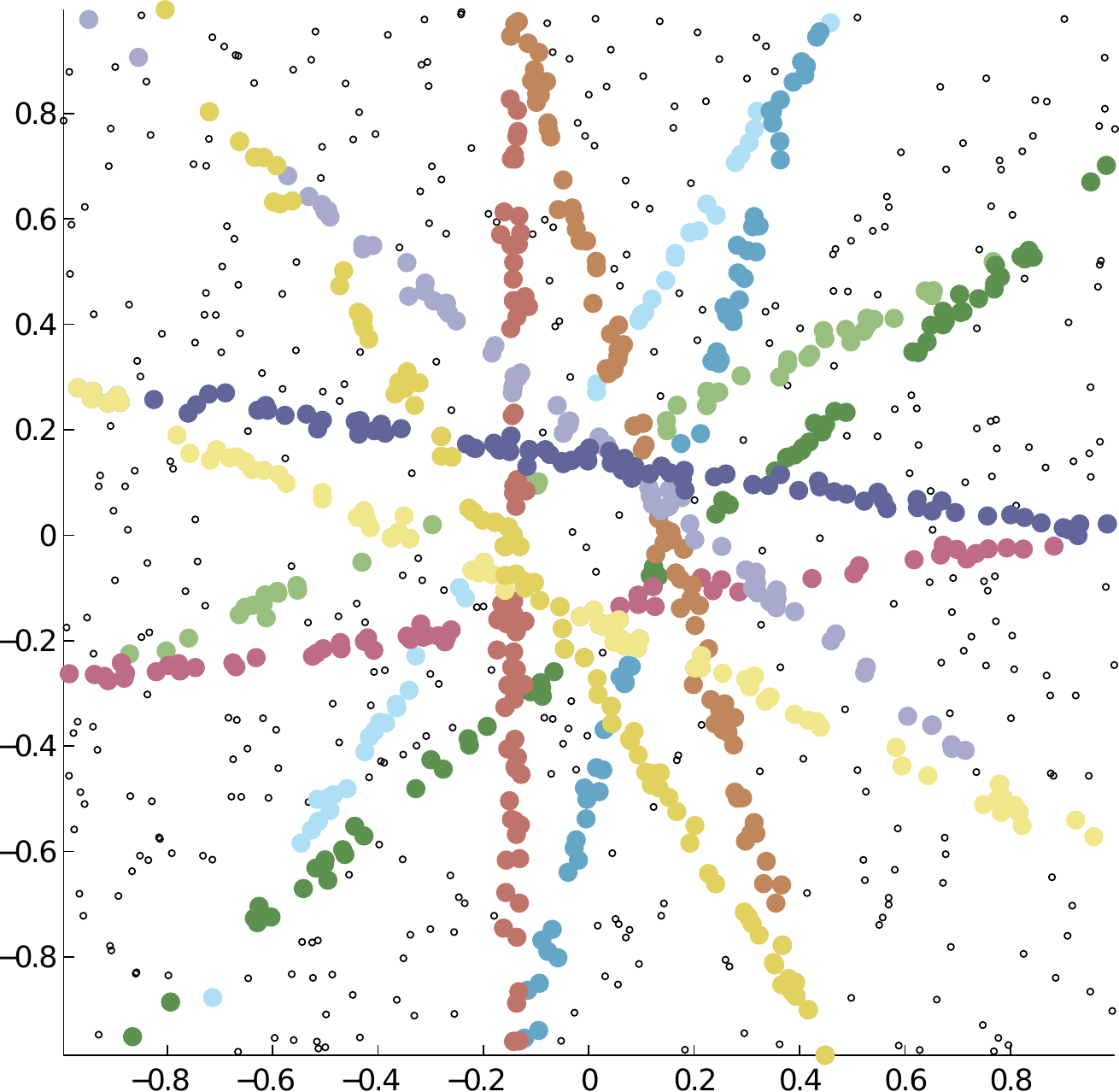} &
		\includegraphics[width=.238\textwidth]{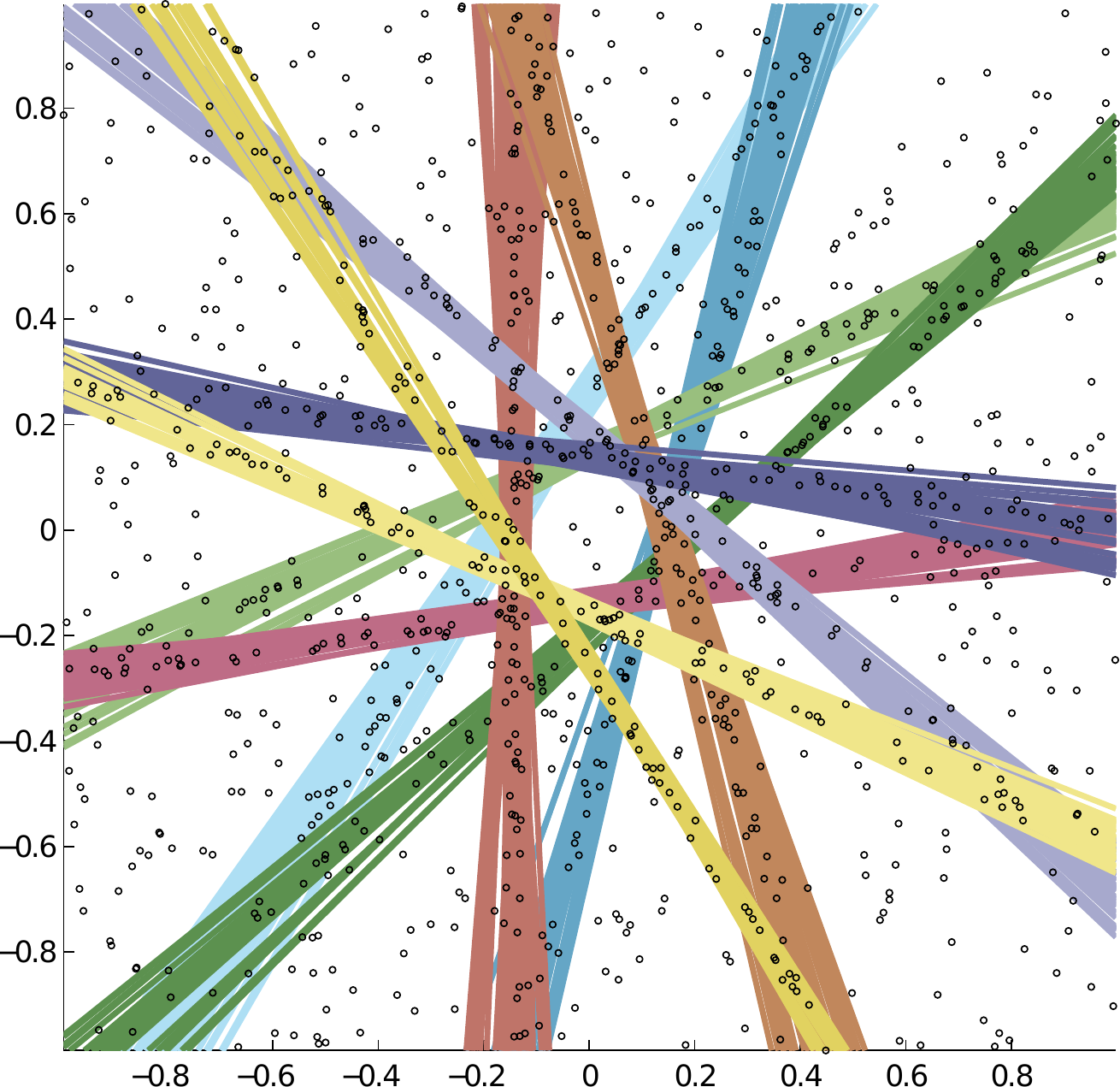} \\

		D &
		\includegraphics[width=.238\textwidth]{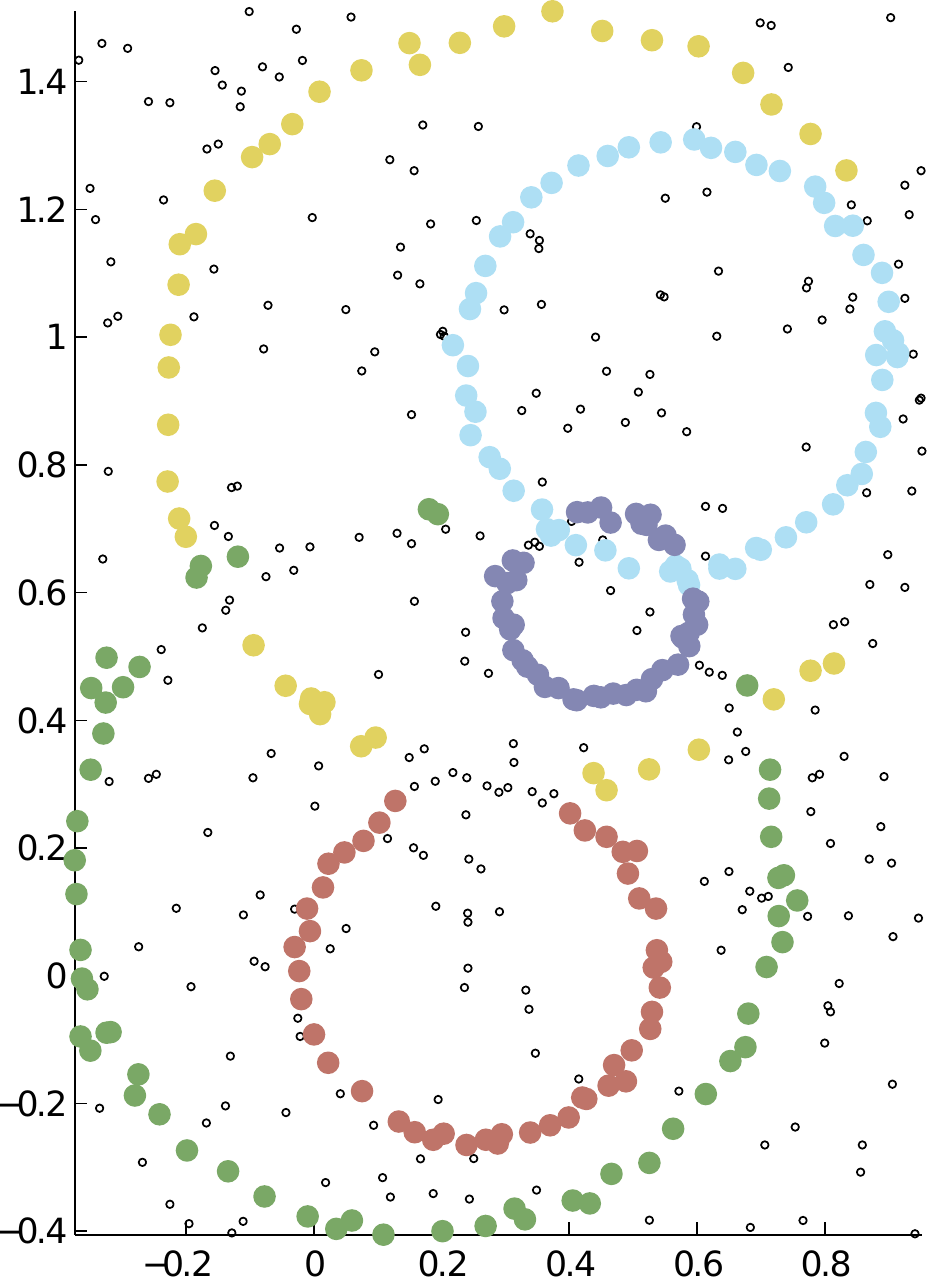} &
		\includegraphics[width=.238\textwidth]{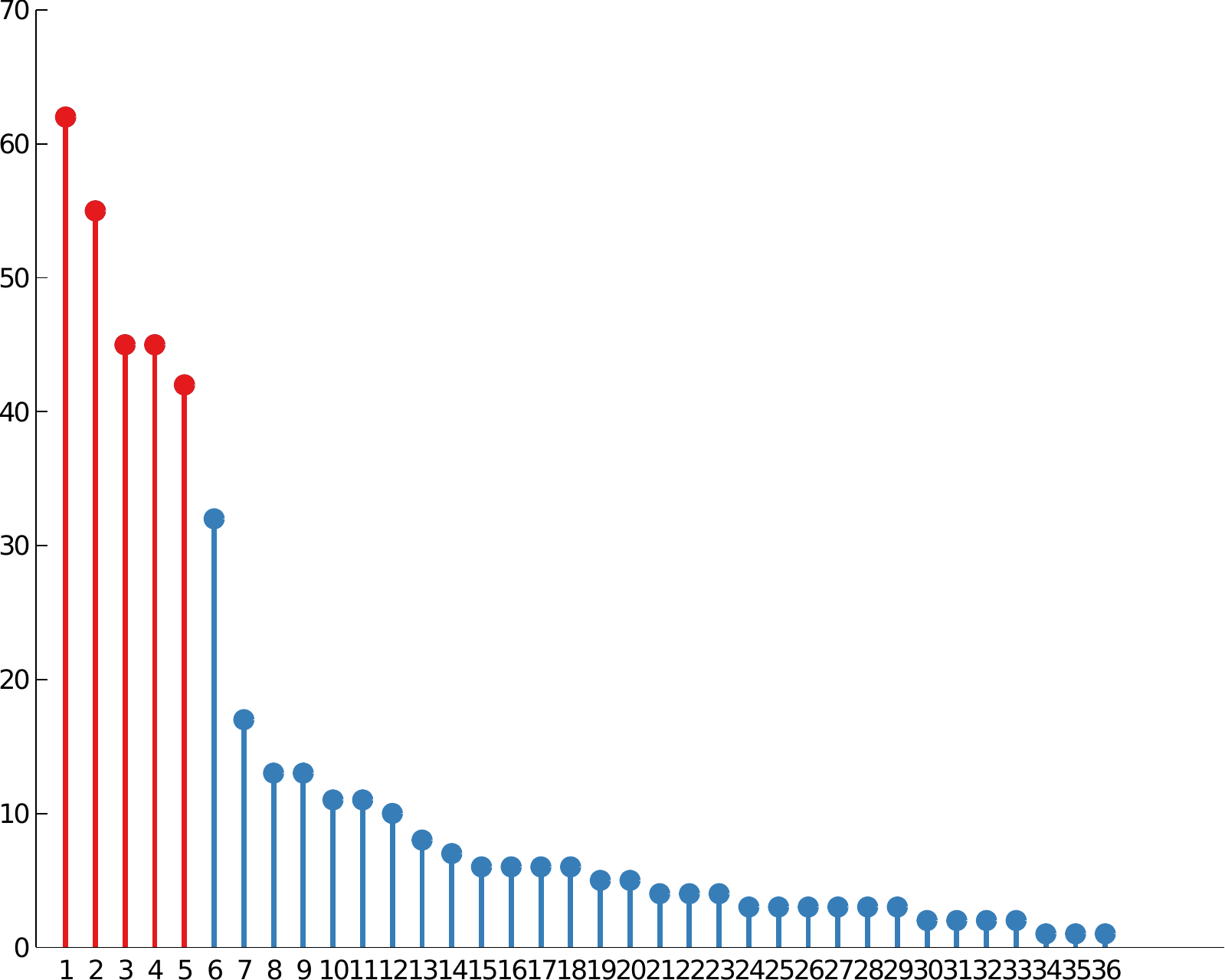} &        
		\includegraphics[width=.238\textwidth]{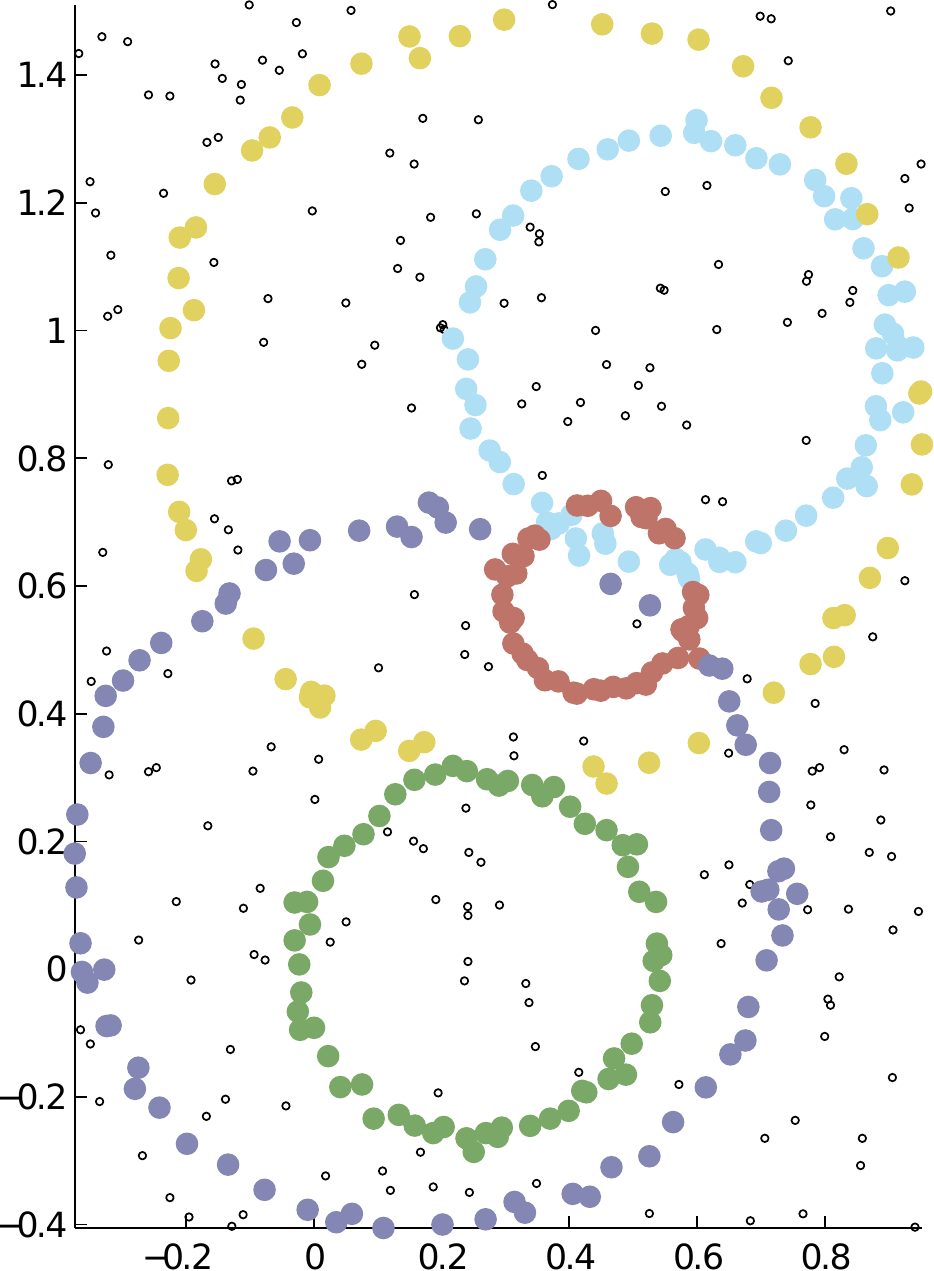} &
		\includegraphics[width=.238\textwidth]{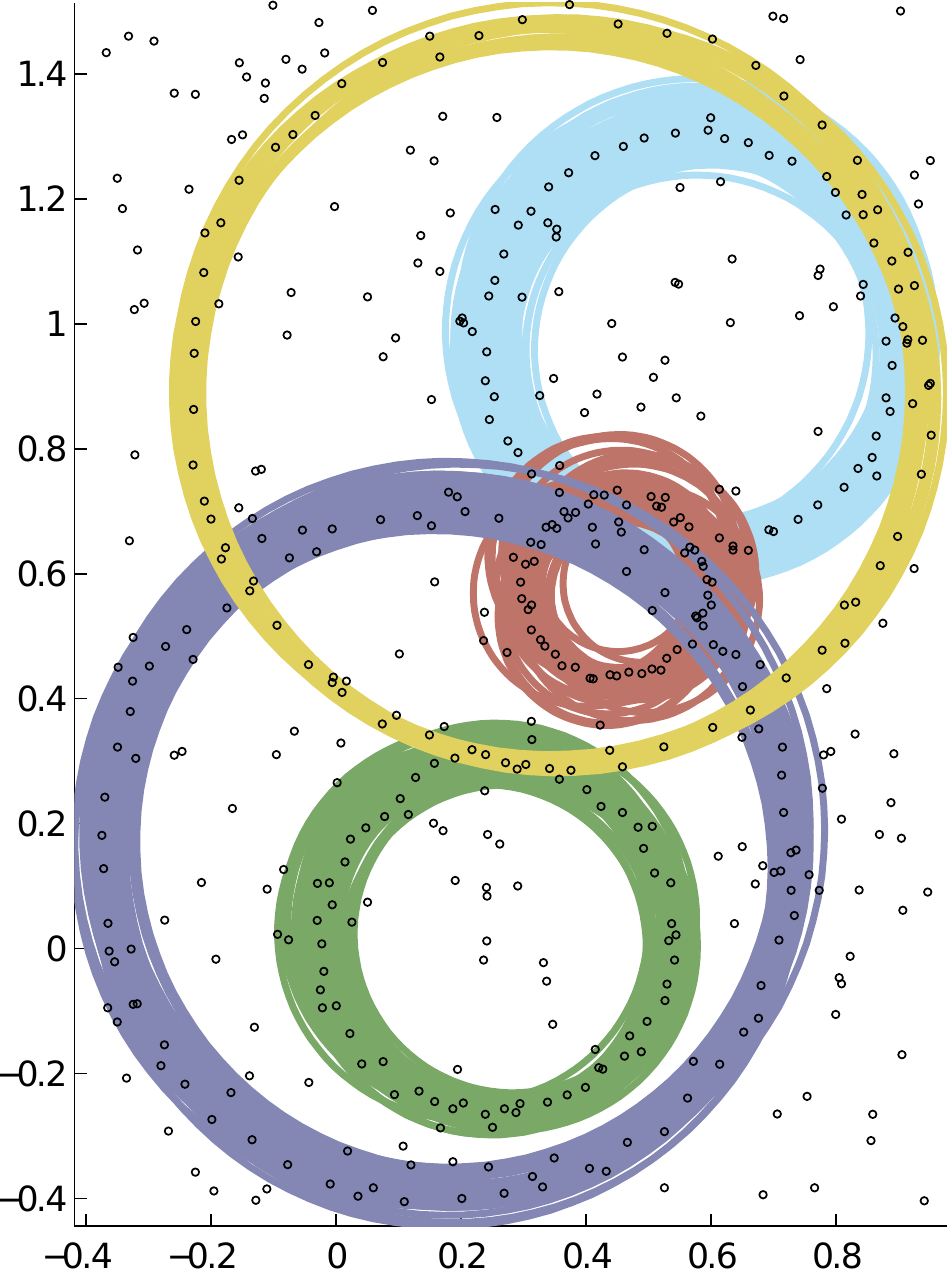} \\        
		
	\end{tabular}
	\end{small}
	
	\caption{J-linkage does not perform accurate point-model assignments, notice the missing points on the detected lines and circles. The size of the J-linkage clusters is not a robust criterion for selecting the final clusters. The proposed approach correctly retrieves the lines and circles.}
	\label{fig:mpme_synthetic_figures}
    \end{subfigure}
    
        \begin{subfigure}[b]{\textwidth}
        \centering
	\begin{tabular}{@{\hspace{0pt}} c d{2.1} d{3.1} @{\hspace{0pt}}}
		\toprule
		\multicolumn{1}{c}{Dataset} &
		\multicolumn{1}{c}{J-linkage} & \multicolumn{1}{c}{Bi-clustering} \\
		\midrule
		A & 96 & 100.0 \\ % Star5_S00075_O50
		B & 87.2 & 100.0 \\ % Star5_S00075_O75
		C & 82.2 & 99.6 \\ % Star11_S00075_O50
		D & 81.6 & 99.6 \\ % Circles5_S00075_O50
		\bottomrule
	\end{tabular}
    	\caption{Comparison of the recall (\%) for the above examples (in these settings precision is distorted by the outliers and is not completely meaningful). This shows that the bi-cluster sizes are very close to the ground truth sizes (slightly larger, again because of the outliers).}
	\label{fig:mpme_synthetic_recall}
	\end{subfigure}

    \caption{Several synthetic examples where the proposed approach yields considerable improvements over J-linkage~\cite{toldo08}, both on the quality and the stability of the results.}
    \label{fig:mpme_synthetic}
\end{figure*}

\begin{figure}[t]
    \centering
    \begin{small}
    \begin{tabular}{@{\hspace{0pt}}c@{\hspace{4pt}}c@{\hspace{0pt}}}
        Residual Histogram Analysis~\cite{zhang2006ransac} &
        Sequential RANSAC \\
        
        \includegraphics[height=.3\textwidth]{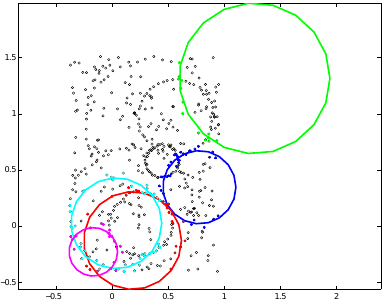} &
        \includegraphics[height=.3\textwidth]{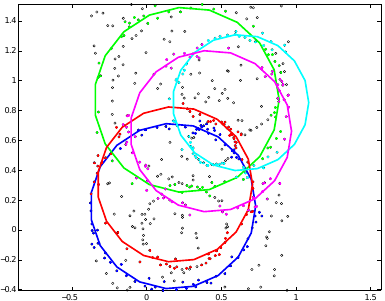} \\

        MultiRANSAC~\cite{zuliani2005} &
        Mean-Shift~\cite{subbarao09} \\

        \includegraphics[height=.3\textwidth]{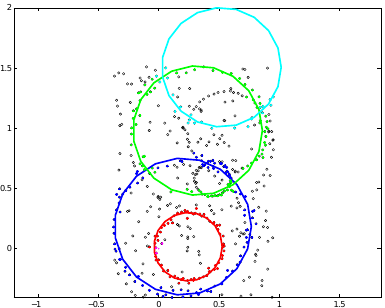} &
        \includegraphics[height=.3\textwidth]{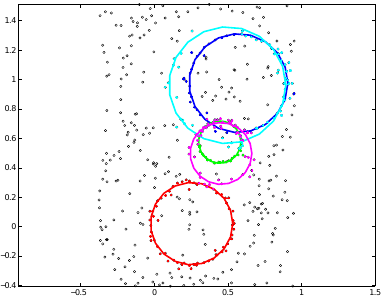} \\
    \end{tabular}
    \end{small}
    
    \caption{Comparisons with several multiple model estimation algorithms on the example in Figure~\ref{fig:mpme_synthetic} (reproduced from~\cite{toldo08}). Contrarily to the proposed approach, these techniques both miss models and detect false ones.}
    \label{fig:otherTechniques}
\end{figure}

We provide several standard 2D examples~\cite{toldo08} with lines and circles in Figure~\ref{fig:mpme_synthetic}.
For the line examples, we set $n=5000$, $\sigma=0.5$, $\delta=0.03$. For circles, we set $n=20000$, $\sigma=0.5$, $\delta=0.03$.
In all examples, the proposed bi-clustering approach does a much better job at recovering the data structure than J-linkage and other popular techniques, see Figure~\ref{fig:otherTechniques}. J-linkage, considered a state-of-the-art algorithm,  has a general tendency to find clusters with fewer points than expected (there are many `undercomplete' lines and circles). This is further emphasized by comparing the recall of both methods, see Table~\ref{fig:mpme_synthetic_recall}.

In all these examples, the bi-clustering algorithm automatically finds the number of clusters.
J-linkage uses the size of the obtained clusters to decide whether to keep them or to discard them.
In Figure~\ref{fig:mpme_synthetic_figures}, it can be easily seen that this method is not stable, since the decay in the size does not indicate the proper cut-point.

We also observe in Figure~\ref{fig:mpme_synthetic} that the proposed approach can correctly recover overlapping models. This is an intrinsic limitation of J-linkage and most multiple model estimation techniques, which are generally based on partitioning (clustering) the set of objects (data).

\begin{figure}[t]
    \centering
    
    \begin{subfigure}[b]{\textwidth}
    	\centering
    	\begin{tabular}{ @{\hspace{0pt}} *{3}{m{.23\columnwidth}} @{\hspace{0pt}} }
        \includegraphics[width=.23\columnwidth]{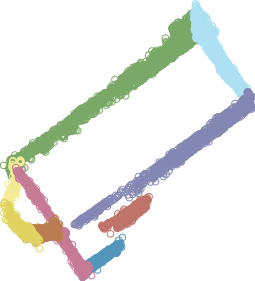} &
        \includegraphics[width=.23\columnwidth]{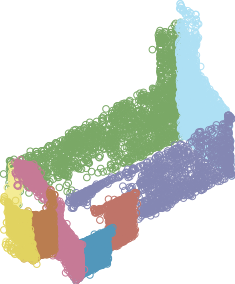} &
        \includegraphics[width=.23\columnwidth]{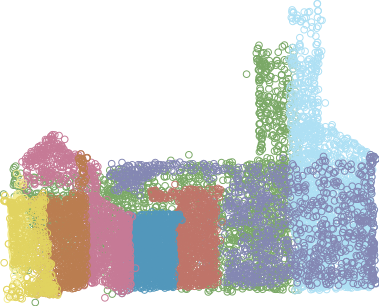} \\

        \includegraphics[width=.23\columnwidth]{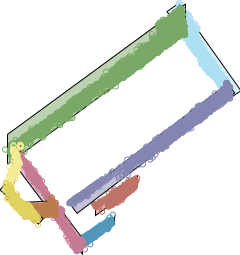} &
        \includegraphics[width=.23\columnwidth]{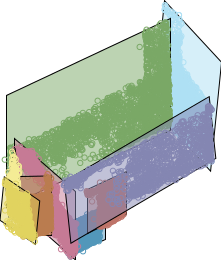} &
        \includegraphics[width=.23\columnwidth]{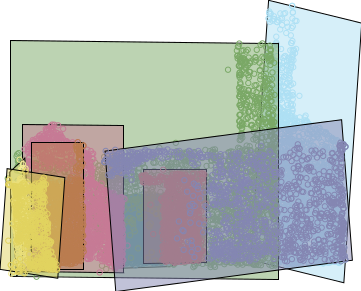} \\
        \end{tabular}
        
        \caption{Three 3D views of the 9 bi-clusters obtained with the proposed method. On the bottom row, we also display the fitted planes}
	\label{fig:planes3D}
    \end{subfigure}
    \\
    \begin{subfigure}[b]{\textwidth}
    	\centering
%    \begin{tabular}{ccc}
        \shortstack{
            \includegraphics[width=.25\columnwidth]{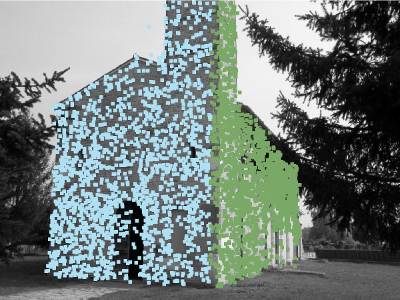}
            \includegraphics[width=.25\columnwidth]{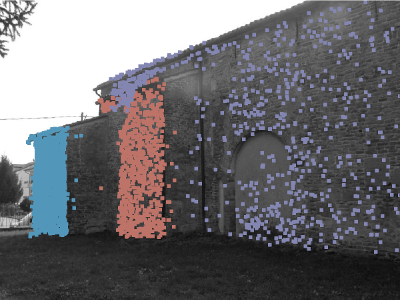}
            \includegraphics[width=.25\columnwidth]{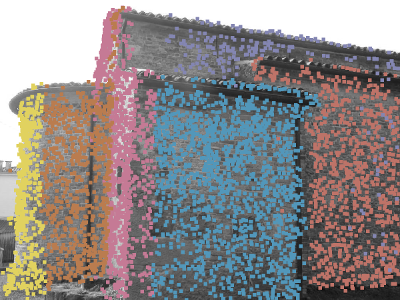}
        }
%    \end{tabular}

	\caption{Three different views of the 3D points projected back to the original images.}
	\label{fig:planesImages}
    \end{subfigure}

    \caption{Example with 3D planes on the Pozzoveggiani dataset. The 3D points are obtained from a set of images, such as the ones in~\protect\subref{fig:planesImages}, using multi-view 3D reconstruction. We correctly recover the building structure by detecting 3D planes.}
    \label{fig:planes}
\end{figure}

Figure~\ref{fig:planes} presents an example of a real application with 3D planes from the Pozzovegianni dataset.\footnote{\url{http://www.diegm.uniud.it/fusiello/demo/samantha/\#pozzo}}
The 3D points are obtained from different images of a building with a sparse multi-view 3D reconstruction algorithm. Our algorithm recovers the 3D planes in the scene to properly reconstruct the building structure (for this example, $n=5000$, $\sigma=0.5$, $\delta=0.5$).

\begin{figure}[t]

	\centering
	\begin{small}
	\begin{tabular}{ccc}
		\multirow{2}{*}{Original line segments} & \multicolumn{2}{c}{Consensus (without final cleansing)} \\
		\cmidrule(lr){2-3}
		& Segment assignments & Model assignments \\
		
		\includegraphics[width=.27\textwidth]{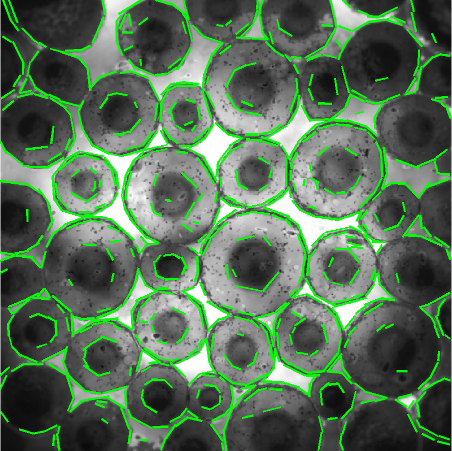} &
		\includegraphics[width=.27\textwidth]{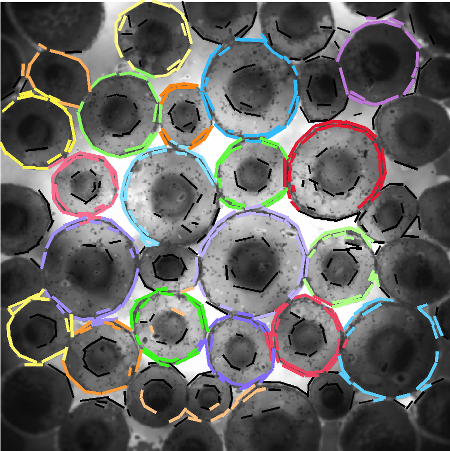}
		\begin{tikzpicture}[overlay]
			\draw [green, ultra thick, ->] (-106pt,97pt) -- ++(8pt,-8pt);
			\draw [green, ultra thick, ->] (-63pt,-3pt) -- ++(0pt,8pt);
		\end{tikzpicture}		
		&
		\includegraphics[width=.284\textwidth]{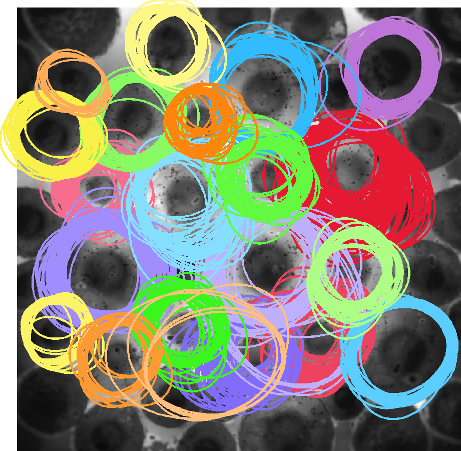}
		\begin{tikzpicture}[overlay]
			\draw [green, ultra thick, ->] (-106pt,97pt) -- ++(8pt,-8pt);
			\draw [green, ultra thick, ->] (-63pt,-3pt) -- ++(0pt,8pt);
		\end{tikzpicture}
		\\[6pt]

		\multirow{2}{*}{\shortstack{Line segments selected\\during sampling}} & \multicolumn{2}{c}{Consensus (with final cleansing)} \\
		\cmidrule(lr){2-3}
		& Segment assignments & Model assignments \\

		\includegraphics[width=.27\textwidth]{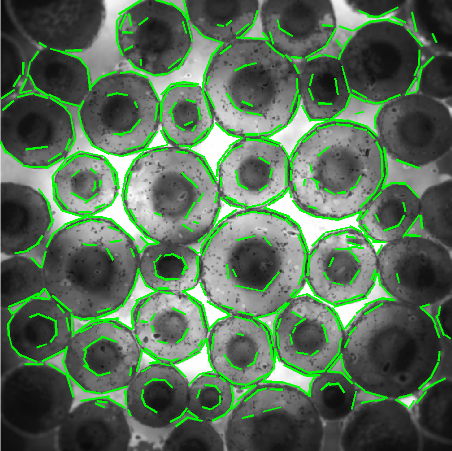} &
		\includegraphics[width=.27\textwidth]{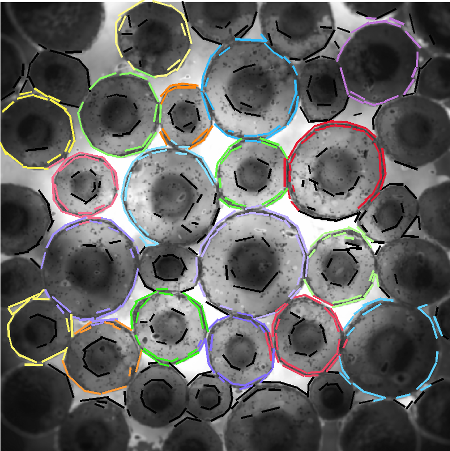} &
		\includegraphics[width=.284\textwidth]{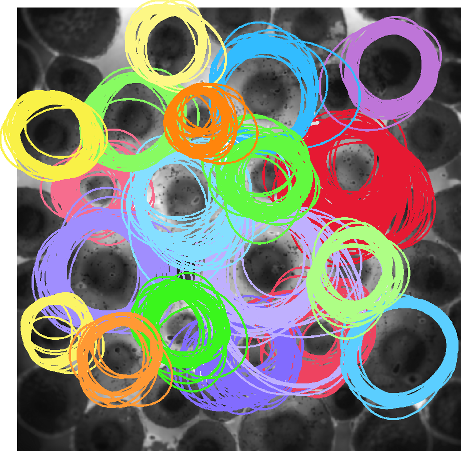} \\
	\end{tabular}
	
	\vskip.5em
	
	\begin{tabular}{ccc}
		\multicolumn{3}{c}{J-linkage result} \\
		\cmidrule(lr){1-3}

		All segment assignments &
		Wrong assignments &
		Cluster sizes \\
		\includegraphics[width=.27\textwidth]{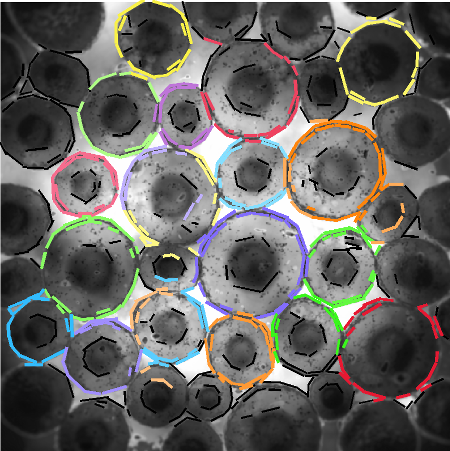} &
		\includegraphics[width=.27\textwidth]{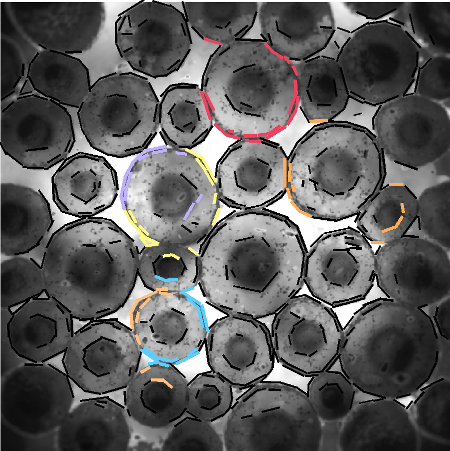} &
		\includegraphics[width=.33\textwidth]{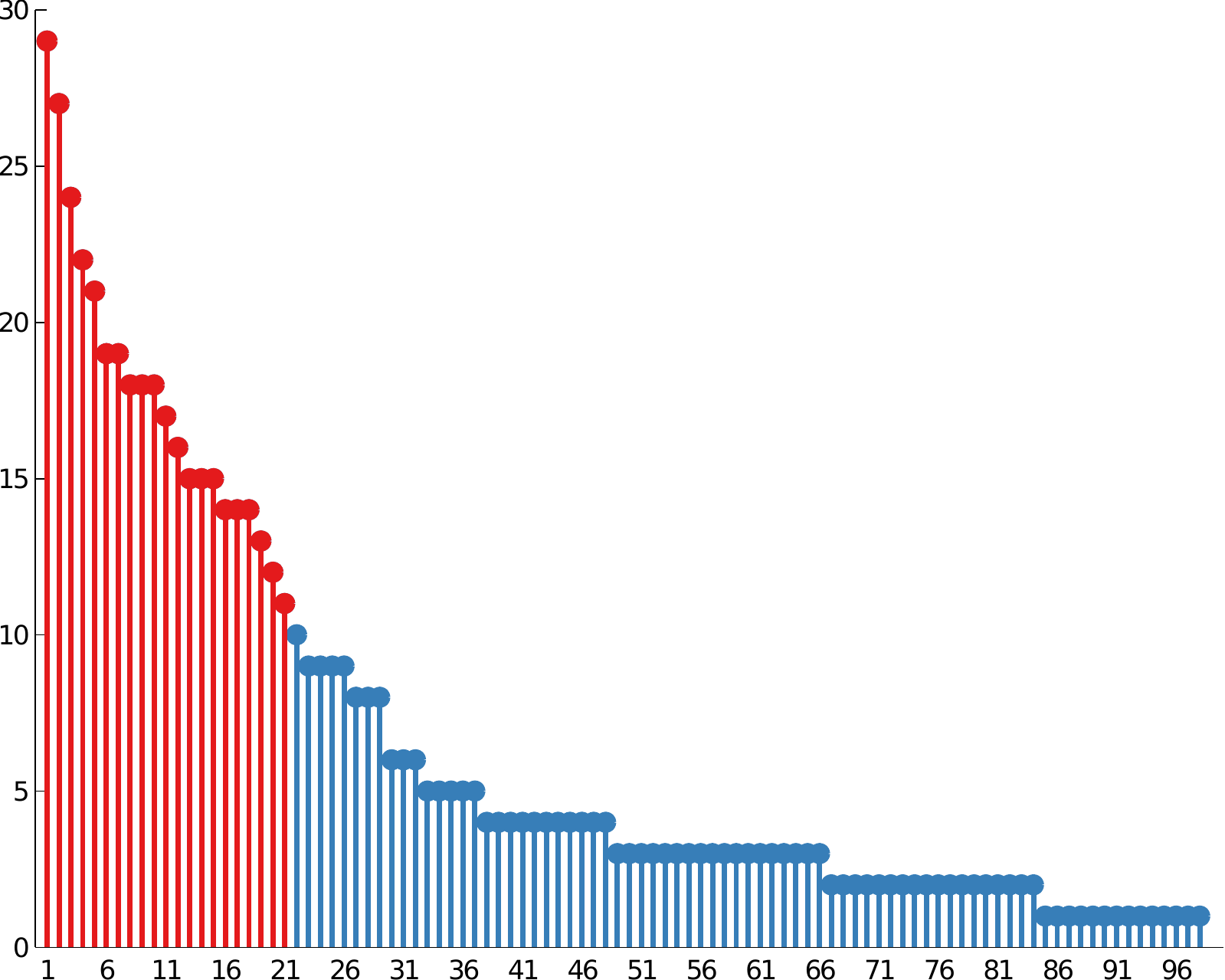} \\
	\end{tabular}
	
	\end{small}
	
	\caption{Ellipse (cell) detection example. We use line segments~\cite{grompone10} as base elements for our method. We are able to reliably detect most ellipses without a specific and highly tuned method. Notice that the final cleansing process eliminates two detected ellipses, one in the top left corner and one on the bottom of the image (indicated with green arrows in the first row). These removals make sense from a perceptual point of view. J-linkage (bottom row) does not yield good results, returning wrong clusters and exhibiting an arbitrary cut-off point (notice the smooth decay in the cluster sizes).}
	\label{fig:ellipses}
\end{figure}
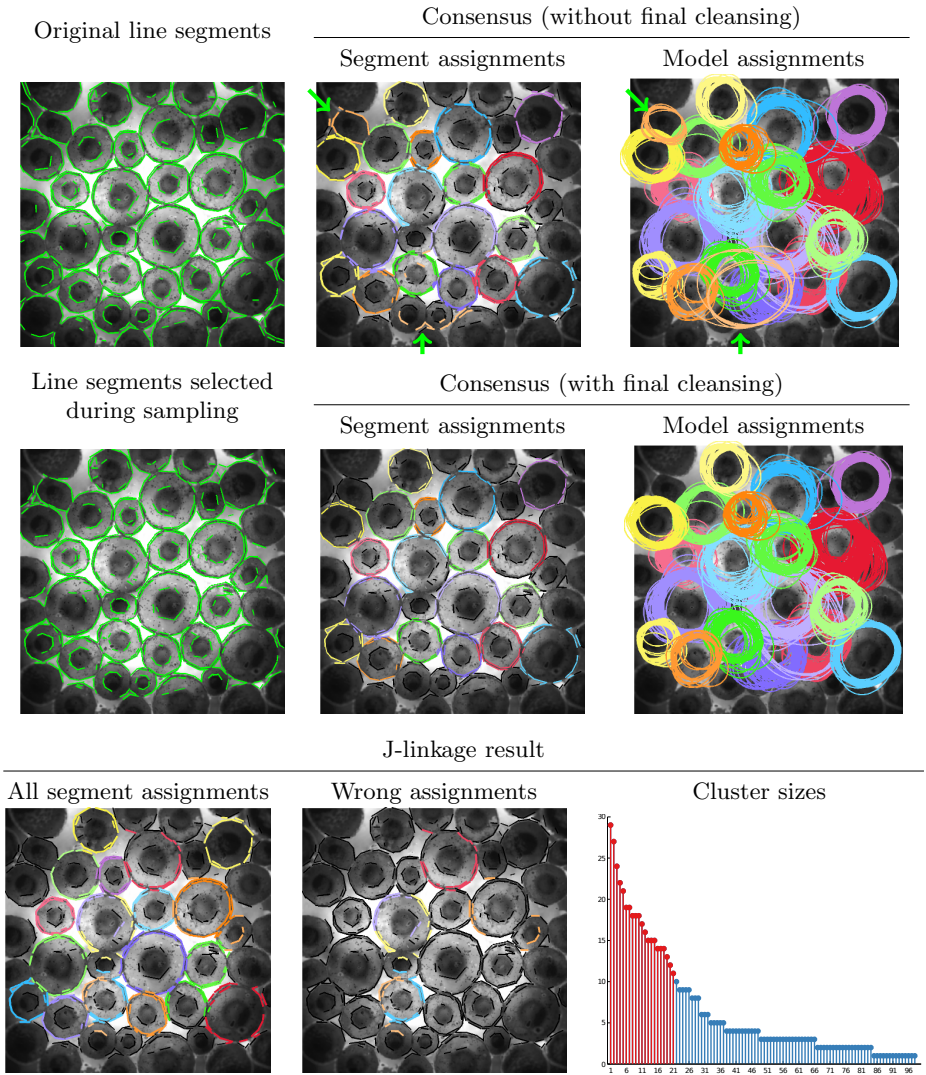

As an additional example, we developed a simple method for detecting cells in microscopy images, see Figure~\ref{fig:ellipses}. This is achieved by using ellipses as models (during sampling, we discard ellipses that are too elongated). Instead of using edge points as the base elements, we use line segments, detected with LSD~\cite{grompone10}, providing more robust detections and making the process of estimating an ellipse from an MSS more stable. For this example, we set $n=20000$, $\sigma=30$, $\delta=10$ and instead of using a zero-centered distance distribution, see Equation~(\ref{eq:distanceDistribution}), we set the mean distance to 80 pixels. Our method retrieves most of the ellipses (i.e., cell membranes) in the image without further tuning our generic algorithm. Notice that we employ a generic a contrario test, that is not completely adapted to this scenario: should the lengths of the segments be taken into account in the NFA, see definition~(\ref{eq:nfa}), the results would automatically be further refined and improved. This is noticeable for small ellipses (i.e., cell nuclei) that contain very few segments. Our goal in this work is to present the general detection framework, we leave this specific refinement for future work.

\begin{figure}[t]

	\centering
	\includegraphics[width=.35\textwidth]{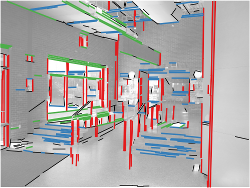}
	\includegraphics[width=.35\textwidth]{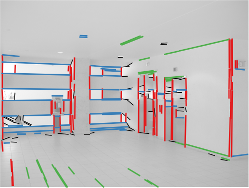} \\[2pt]
	\includegraphics[width=.35\textwidth]{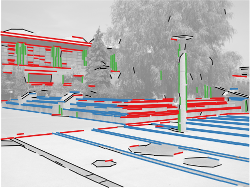}
	\includegraphics[width=.35\textwidth]{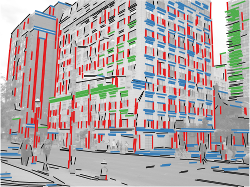}
	
	\caption{Vanishing point detection example. Our simple approach correctly recovers the vanishing points in the images. We only show the line segments in each bi-cluster, because the vanishing points might lie far away from the image in the image plane. The detection of 3D parallelism from 2D images is an ill-posed problem, as some information is inherently lost during the projection. For a handful of segments (e.g., the one marked with a green arrow in the second image), there is no way of determining from the observed projections their true 3D orientation. Despite this indeterminacy, the overall problem is still solvable.}
	\label{fig:vanishing_points}
\end{figure}

Finally, we present another application for the proposed framework: vanishing point detection in uncalibrated images. As with cells, in this application we use line segments as the base objects/elements of our method. The groups are formed by detecting the subset of lines that intersect at a given point in the image plane~\cite{tepper14vp}. In this case, since we only need 2 segments to compute a MSS (i.e., an intersection point)~\cite{tepper14vp}, and the number $m$ of segments in not too large, we compute all $m(m-1)/2$ MSSs, instead of sampling at random ($\sigma=1$, $\delta=10$ pixels). This helps to show that in the other cases the sampling procedure does not artificially boosts the performance, and only helps to speed up the algorithm. In Figure~\ref{fig:vanishing_points} we show some results from the York Urban database~\cite{denis08}, where we can robustly and reliably detect vanishing points.
As with ellipses, the results could be improved by considering the segment length in the a contrario test. Using more sophisticated and specific schemes to obtain the vanishing point candidates might also lead to further improvements~\cite{lezama14}. 

In these experiments we can observe that the bi-clustering approach provides:
(1) a good description for each detected model instance in terms of its objects;
(2) a compact overall description, by considering the reduced number of detected bi-clusters; and
(3) support for overlapping groups of objects.
These features of the proposed general approach lead to a rich characterization of the data in terms of parametric models.

\subsection{The boundaries of multiple parametric model estimation}
\begin{figure}[h]

	\centering
	\begin{tabular}{ccc}
		\multirow{2}{*}{Point configuration} &
		\multicolumn{2}{c}{Bi-clustering result} \\
		\cmidrule(lr){2-3}
		& \multicolumn{1}{c}{Point assignments} & \multicolumn{1}{c}{Model assignments} \\
	
		\includegraphics[width=.25\textwidth]{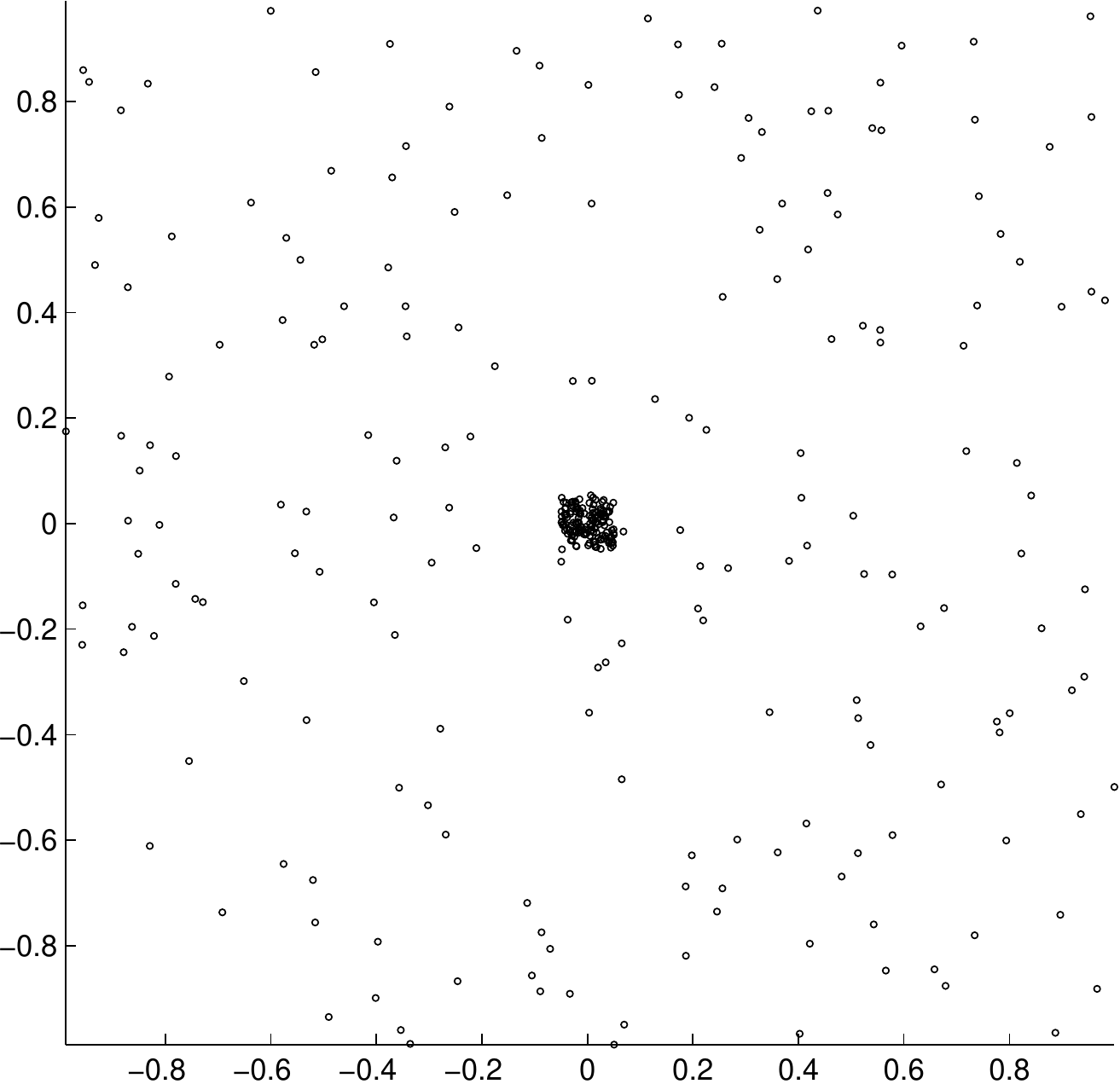} &
		\includegraphics[width=.25\textwidth]{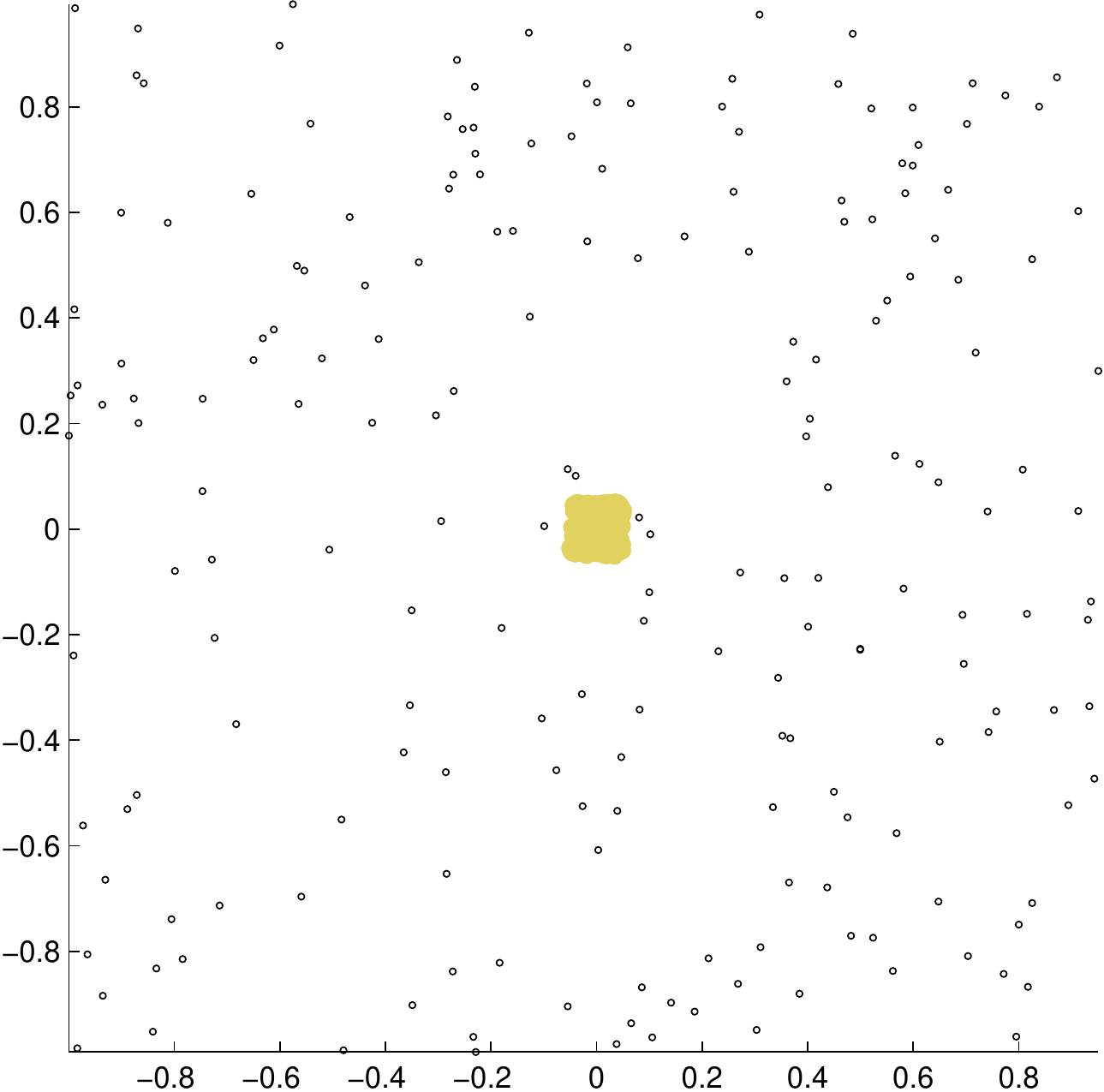} &
		\includegraphics[width=.25\textwidth]{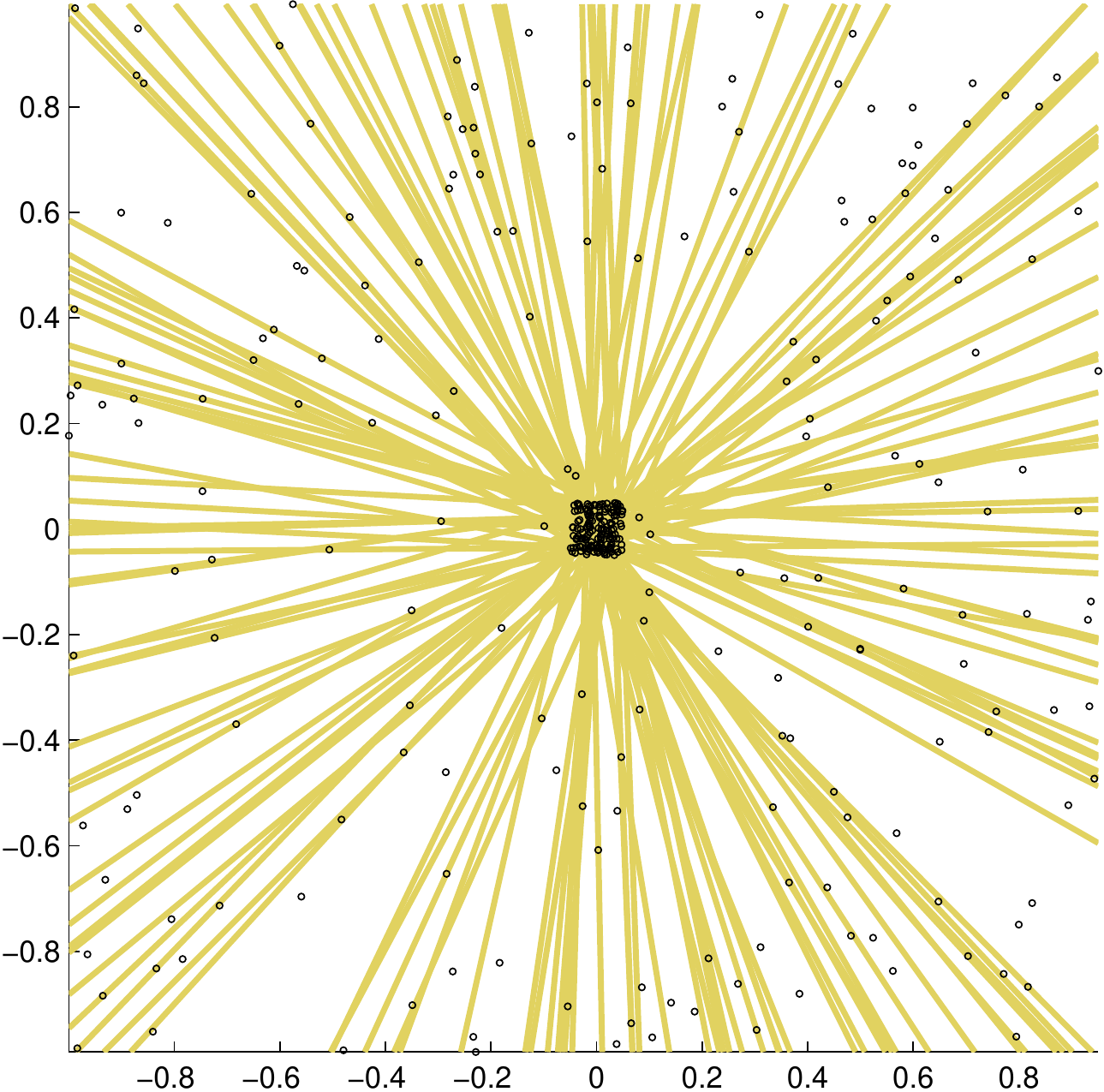} \\
	\end{tabular}
	
	\caption{An aberrant example, containing uniformly distributed points and a small cluster in the center. We run our line detection algorithm and recover the central cluster. The result is consistent with the proposed protocol, although one might claim that a line detection scheme should yield no detections in this case. A stricter definition of the a contrario pruning rule is needed to correct this effect: as before, we need a concentration of points in the orientation orthogonal to the line and \textit{additionally} a dispersion of points along the line.}
	\label{fig:dot}
\end{figure}

The proposed approach for detecting parametric models assumes that observing an unusual concentration of elements around a parametric model instance is enough to produce a robust candidate. In most situations this assumption is valid and works well in practice, as seen in the numerous previous examples. However, there are situations where these element-model distances are not enough to fully characterize a given configuration of elements. Let us illustrate this with a simple example, see Figure~\ref{fig:dot}. Using the classical definition of a consensus set, see~(\ref{eq:consensusSet}), every line that passes through the central cluster will have a large consensus set. This artificially fires many candidate detections, that end up creating a bi-cluster. Notice that (1) this result is completely consistent with our theoretical formulation, and (2) this is not a bi-product of using lines since a similar effect will occur if we use circles, for example. Neither RANSAC nor the Hough transform can discern between points lying along a line and points concentrated in a small cluster. In fact, the setup is missing a dispersion constraint along the model. This has been addressed for a particular case in~\cite{lezama14}, but many formulations could be employed to achieve the desired effect, such as Ripley's K and L functions~\cite{ripley1976second}. Additionally, this improved validation would greatly simplify the validation step of the proposed method.

The values of $\delta$ (the distance threshold) and $n$ (number of considered MSSs) are in general very important to obtain good results. Notice that this dependency is not introduced by our bi-clustering formulation, but common to most parametric model estimation approaches. For example, methods based on RANSAC and the randomized Hough transform share this dependency. In the general case, there is no theoretical formulation nor clear practical guideline for properly setting their values. In practice, many cases present a reasonably intuitive range for $\delta$. In our experience, setting the value of $n$ is not critical when time is not a concern, as we can simply select a large value. In time-critical applications, we can learn $n$ from training examples.

Of course, the choices of $n$ and $\tau_C$ are related. For a single model, the probability of drawing at least $\tau_C$ outlier-free MSSs out of $n$ (and thus guaranteeing its detection) is given by
\begin{equation}
	1 - \sum^{\tau_C - 1}_{k=0} \binom{n}{\tau_C} (p_\text{MSS})^k ( 1 - p_\text{MSS} )^{n - k} ,
\end{equation}
where $p_\text{MSS}$ is the probability of drawing an MSS of cardinality $b$ composed only of inliers (recall that $b$ is the minimum number of elements necessary to uniquely characterize a given parametric model). See~\cite{toldo08} for further details. This formulation has two shortcomings. First, $p_\text{MSS}$ is not a trivial quantity to set and/or estimate without deep knowledge about the structure of the dataset (consider that we are actually trying to discover this structure). Second, the above equation relates $n$ and $\tau_C$ for a single model; establishing a sound relation for multiple models requires knowing in advance the number of models. More research is much needed in this theoretically and practically important aspect, as all state-of-the-art approaches would benefit from it. As a side note, this problem is similar in nature to the Chinese restaurant and Indian buffet processes, but with a fixed and unknown number of tables/dishes~\cite{aldous1985,griffiths2005infinite}.

% !TEX root = main.tex
\section{Connection with the computational Gestalt theory}
\label{sec:computational_gestalt}

We now provide a brief discussion on the connections between the presented framework and the computational Gestalt theory.

The driving principle in the computational Gestalt theory is the Helmholtz principle~\cite{desolneux08}. In its simplest form, it states that we do not perceive any structure in a uniform random image~\cite{attneave54}. The computational Gestalt theory makes extensive use of a stronger form, namely that whenever some large deviation from randomness occurs, a structure is perceived.

The Helmholtz principle is generically illustrated as follows.
Let $\set{X}$ be the set of atomic objects present in an image. Let us assume that there is a subset $\set{X}_{\text{CF}} \subset \set{X}$ whose objects share a common feature, say, color, orientation, position, etc. We then face a decision problem: is this common feature happening by chance or is it significant enough to make $\set{X}_{\text{CF}}$ stand out as a perceptual group? To answer this question, let us make the following mental experiment: assume a priori that the considered feature is randomly and uniformly distributed on all objects of $\set{X}$, i.e., the observed feature is a realization of this uniform process. We finally ask: is this realization probable or not? If not, this proves a contrario that a grouping process (a gestalt) is at play. The Helmholtz principle states roughly that in such mental experiments, the object features are assumed to be uniformly distributed and independent~\cite{desolneux08}.

The theory can be interpreted as a multiple hypothesis testing framework for visual events and works by controlling the Number of False Alarms (NFA), a proxy of the expectation of the number of occurrences of a visual event under the stated a contrario model.

In Section~\ref{sec:mpme} we exploited this detection theory twice: (1) to cleanse the preference matrix, and (2) to adjust $\tau_R$, the minimum number of elements (rows) necessary to accept a bi-cluster. These a contrario tests allow us to assess the improbability of a given event or configuration. However, our framework also computes information regarding the repeatability of this configuration when probing the data. In this sense, we can interpret our framework as repeatedly querying the data at random and looking for configurations that arise over and over. The a contrario tests do not exploit this additional dimension that nonetheless plays a central role when assessing randomness: if a given configuration arises over and over, what are the chances that it is a realization of a random process? This simple discussion brings forward the need to develop statistical tests that also consider the repeatability of a configuration under random sampling, via $\tau_c$ in our framework.

As a side note, we would also like to point out the fact that, when $(\forall \theta)\ p(\mu(\theta),\delta)= p_\delta$, as in the case of 2D lines (see Appendix~\ref{sec:probaComputation}), $p_\delta \approx \norm{\mat{A}}{0} / (mn)$, where $\norm{\bullet}{0}$ stands for the pseudo-norm counting the number of non-zeros. This simple observation might lead to develop simpler, more general, and more extensible geometric probes, by employing a reduced set of assumptions for computing $p(m(\theta),\delta)$ than the ones used in Appendix~\ref{sec:probaComputation}, common in all a contrario literature.

The presented framework also has connections with the so-called Pr\"agnanz principle in Gestalt psychology: ``...of several geometrically possible organizations that one will actually occur which possesses the best, simplest and most stable shape,'' quoted in~\cite{zhu1999} from Koffka's book~\cite{koffka1935principles}. Analyzing the set of possible configurations, the proposed framework assigns a formal and mathematical meaning to the terms \emph{best, simplest, and most stable}. Let us analyze these characteristics one by one:
\begin{description}
	\item[Best.] Our resulting configurations (bi-clusters) are obtained as the solution of a non-parametric minimization problem with an intuitive interpretation, see problem~\ref{eq:nmf}.
	\item[Simplest.] The rank-one nature of the extracted bi-clusters means that they provide a minimalistic explanation for the set of configurations. This set is interpreted as a summation of rank-one bi-clusters, plus some noise.
	\item[More stable.] Configurations that are hard to reproduce exhibit low stability. We already discussed how our framework extracts the configurations that show more repeatability when probing the data. 
\end{description}
From this point of view, our framework might provide a mathematically sound way to formulate and implement the Pr\"agnanz principle, while integrating it with the Helmholtz principle, seen as a stopping criterion for the bi-clustering process. Psychophysical experiments are of course needed to fully validate this intuitive conceptual vinculation.

\section{Conclusion}
\label{sec:conclusions}

In this paper we proposed a framework and a new perspective for reaching consensus in grouping problems.
Our general characterization of grouping problems subsumes many different areas, e.g., clustering, community detection in networks, and multiple parametric model estimation.
We pose consensus grouping as a bi-clustering problem, obtaining a conceptually simple and descriptively rich modeling. We presented a simple but powerful bi-clustering algorithm, specifically tuned to the nature of the problem we address, though general enough to handle many different instances inscribed within our framework.

In particular, this is the first time that the task of finding/fitting multiple parametric models to a dataset was formally posed as a consensus bi-clustering problem. The equivalence of these tasks is highlighted by the proposed framework, and we devoted special attention to explain the rationale behind this new characterization. As a future line of research, we are currently investigating whether using a hard thresholding scheme is actually necessary. Instead of working with binary data, we could work with a real-valued object-model distance matrix, eliminating a critical parameter that has been haunting the RANSAC framework for years.

We also discussed the connection with the computational Gestalt program~\cite{desolneux08}, that seeks to provide a quantitative psychologically-inspired detection theory for visual events. We provided some cues that show the suitability of our approach as a new research direction in this field. We are working on exploiting these new connections to fully develop a new perspective on this fundamental problem.

We are also exploring how to use a contrario statistical tests in non-parametric scenarios. From this perspective, we point out again that we do not need a very sharp testing mechanism but a coarse procedure to avoid filling the preference matrix with a huge number of poor groups that could clutter the bi-clustering algorithm. This would allow for more freedom in the selection of the pool of input algorithms.

As an alternative, the framework could be extended with individual weights $w_{ij}$ (instead of column-wide weights) in the preference matrix. These weights would thus model a confidence measure of the $i$-th element belonging to the $j$-th group. It would be interesting to explore this possibility in depth.

Finally, we would like to stress that the proposed framework is not limited to the presented applications. It is flexible enough to handle any type of grouping problem and we plan to address other applications that can be formulated in this way. Two clear examples of this are image segmentation (seen as an extension of clustering with spatial constraints) and supervised classification, where we aim at fusing the output of different classifiers to obtain robustified results.

\appendix

% !TEX root = main.tex
\section*{}
\label{sec:nmf_algorithm}
We now show how to solve problem~(\ref{eq:nmf}).
The problem can be equivalently re-formulated as
\begin{equation}
    \min_{\mat{U}, \mat{V}, \mat{X}, \mat{Y}, \mat{E}} \norm{\mat{E}}{1} \quad \text{s.t.} \quad
    \begin{gathered}
    \mat{E} = \mat{A} - \mat{X} \mat{Y} \\
    \mat{X} = \mat{U}, \mat{Y} = \mat{V} \\
    \quad \mat{U}, \mat{V} \geq 0 ,
    \end{gathered}
    \label{eq:nmfEquiv}
\end{equation}
where $\mat{E} \in \Real^{m \times n}$, $\mat{X}, \mat{U} \in \Real^{m \times q}$, and $\mat{Y}, \mat{V} \in \Real^{q \times n}$.
We consider the augmented Lagrangian of~(\ref{eq:nmfEquiv}),
\begin{align}
    \mathscr{L} (\mat{X}, \mat{Y}, \mat{U}, \mat{V}, \mat{E}, \Lambda, \Phi, \Psi)
    &= \norm{\mat{E}}{1} 
    + \Lambda \bullet (\mat{X}-\mat{U})
    + \Phi \bullet (\mat{Y}-\mat{V}) + \nonumber \\
    &+ \Psi \bullet (\mat{A}- \mat{X} \mat{Y} - \mat{E}) 
    + \tfrac{\alpha}{2} \norm{\mat{X} - \mat{U}}{F}^2 + \nonumber \\
    &+ \tfrac{\beta}{2} \norm{\mat{Y} - \mat{V}}{F}^2
    + \tfrac{\gamma}{2} \norm{\mat{A} - \mat{X} \mat{Y} - \mat{E}}{F}^2 ,
\end{align}
where $\Lambda \in \Real^{m \times q}, \Phi  \in \Real^{q \times n},  \Psi \in \Real^{m \times n}$ are Lagrange multipliers, $\alpha, \beta, \gamma$ are penalty parameters, and $\mat{B} \bullet \mat{C} = \sum_{i,j} (\mat{B})_{ij} (\mat{C})_{ij}$ for matrices $\mat{B}, \mat{C}$ of the same size.

We use the Alternating Direction Method of Multipliers (ADMM) for solving~(\ref{eq:nmfEquiv}). The algorithm works in a coordinate descent fashion, successively minimizing $\mathscr{L}$ with respect to $\mat{X}, \mat{Y}, \mat{U}, \mat{V}, \mat{E}$, one at a time while fixing the others at their most recent values, i.e.,
\begin{subequations}
\begin{align}
    \mat{X}_{k+1} &= \argmin_{\mat{X}} \mathscr{L} (\mat{X}, \mat{Y}_{k}, \mat{U}_{k}, \mat{V}_{k}, \mat{E}_{k}, \Lambda_{k}, \Phi_{k}, \Psi_{k}) , \\
    \mat{Y}_{k+1} &= \argmin_{\mat{Y}} \mathscr{L} (\mat{X}_{k+1}, \mat{Y}, \mat{U}_{k}, \mat{V}_{k}, \mat{E}_{k}, \Lambda_{k}, \Phi_{k}, \Psi_{k}) , \\
    \mat{U}_{k+1} &= \argmin_{\mat{U} \geq 0} \mathscr{L} (\mat{X}_{k+1}, \mat{Y}_{k+1}, \mat{U}, \mat{V}_{k}, \mat{E}_{k}, \Lambda_{k}, \Phi_{k}, \Psi_{k}) , \\
    \mat{V}_{k+1} &= \argmin_{\mat{V} \geq 0} \mathscr{L} (\mat{X}_{k+1}, \mat{Y}_{k+1}, \mat{U}_{k+1}, \mat{V}, \mat{E}_{k}, \Lambda_{k}, \Phi_{k}, \Psi_{k}) , \\
    \mat{E}_{k+1} &= \argmin_{\mat{E}} \mathscr{L} (\mat{X}_{k+1}, \mat{Y}_{k+1}, \mat{U}_{k+1}, \mat{V}_{k+1}, \mat{E}, \Lambda_{k}, \Phi_{k}, \Psi_{k}) ,
\end{align}
and then updating the multipliers $\Lambda, \Phi, \Psi$, i.e.,
\begin{align}
    \Lambda_{k+1} &= \argmin_{\Lambda} \mathscr{L} (\mat{X}_{k+1}, \mat{Y}_{k+1}, \mat{U}_{k+1}, \mat{V}_{k+1}, \mat{E}_{k+1}, \Lambda, \Phi_{k+1}, \Psi_{k+1}) , \\
    \Phi_{k+1} &= \argmin_{\Phi} \mathscr{L} (\mat{X}_{k+1}, \mat{Y}_{k+1}, \mat{U}_{k+1}, \mat{V}_{k+1}, \mat{E}_{k+1}, \Lambda_{k+1}, \Phi, \Psi_{k+1}) , \\
    \Psi_{k+1} &= \argmin_{\Psi} \mathscr{L} (\mat{X}_{k+1}, \mat{Y}_{k+1}, \mat{U}_{k+1}, \mat{V}_{k+1}, \mat{E}_{k+1}, \Lambda_{k+1}, \Phi_{k+1}, \Psi) .
\end{align}
\end{subequations}
Each of these steps can be written in closed form as
\begin{subequations}
\begin{align}
    \mat{X}_{k+1} &= \left( \gamma \left(\mat{A} - \mat{E}_k \right) \transpose{\mat{Y}_k} + \alpha \mat{U}_k - \Lambda_k + \Psi_k \transpose{\mat{Y}_k} \right) \left( \mat{Y}_k \transpose{\mat{Y}_k} + \alpha \mat{I} \right)^{-1} , \\
    \mat{Y}_{k+1} &= \left( \transpose{\mat{X}_{k+1}} \mat{X}_{k+1} + \beta \mat{I} \right)^{-1} \left( \gamma \transpose{\mat{X}_{k+1}} \left(\mat{A} - \mat{E}_k \right) + \beta \mat{V}_k - \Phi_k +\transpose{\mat{X}_{k+1}} \Psi_k  \right) , \\
    \mat{U}_{k+1} &= \mathscr{P}_+ \left( \mat{X}_{k+1} + \alpha^{-1} \Lambda_k \right) , \\
    \mat{V}_{k+1} &= \mathscr{P}_+ \left( \mat{Y}_{k+1} + \beta^{-1} \Phi_k \right) , \\
    \mat{E}_{k+1} &= \operatorname{shrink} \left( \mat{A} - \mat{X}_{k+1} \transpose{\mat{Y}_{k+1}} + \gamma^{-1} \Psi_k \ ,\ \gamma^{-1} \right) , \\
    \Lambda_{k+1} &= \Lambda_{k} + \xi \alpha \left( \mat{X}_{k+1} - \mat{U}_{k+1} \right) , \\
    \Phi_{k+1} &= \Phi_{k} + \xi \beta \left( \mat{Y}_{k+1} - \mat{V}_{k+1} \right) , \\
    \Psi_{k+1} &= \Psi_{k} + \xi \gamma \left( \mat{A} - \mat{X}_{k+1} \transpose{\mat{Y}_{k+1}} - \mat{E}_{k+1} \right) ,
\end{align}
\label{eq:admm_nmf}
\end{subequations}
where $\mat{I}$ is the $q \times q$ identity matrix, and
\begin{subequations}
\begin{align}
(\mathscr{P}_+ (\mat{B}))_{ij} &= \max\left\{ (\mat{B})_{ij}, 0 \right\} ,\\
(\operatorname{shrink} (\mat{B}, \lambda))_{ij} &= \operatorname{sign}((\mat{B})_{ij}) \, \max\left\{ |(\mat{B})_{ij}| - \lambda, 0 \right\} .
\end{align}
\end{subequations}
These iterations define our algorithm, the initialization being done with a rank-$q$ SVD. In practice, we set $\alpha, \beta, \gamma, \xi$ to 1.

\section*{}
\label{sec:probaComputation}
We now describe how to compute $p(\mu(\theta),\delta)$ for the examples presented in Section~\ref{sec:mpme}.
For simplicity, we assume that the objects under the background model are independent and identically distributed, following a uniform law.
The presented tests are, in some cases, rather crude. Much tighter bounds can be found by carefully tuning the probabilistic models for each specific application. However, in this simpler forms, they are already useful for demonstrating the capabilities of the proposed framework, and are sufficient to lead to state-of-the-art results.

\textbf{2D Lines.}
An object in $\set{X}$ is a 2D point $\vect{x} \in \Real^2$.
We need $b=2$ points to define a line.
The parameter vector $\theta \in \Real^3$ of a line passing through two points $\vect{x}$, $\vect{x}'$, is given by $\theta = \begin{bsmallmatrix}\vect{x}\\1\end{bsmallmatrix} \times \begin{bsmallmatrix}\vect{x}'\\1\end{bsmallmatrix}$. Then, a line is the set of points (see Equation~(\ref{eq:model})) such that
\begin{equation}
    \mu(\theta) = \{ \vect{x} \in \Real^2 ,\ \begin{bmatrix} \vect{x}^T \ 1 \end{bmatrix} \theta = 0 \} ,
\end{equation}
and the distance between a point $\vect{x}$ and a line (see Equation~(\ref{eq:pointModelDistance})) with parameter vector $\theta = \transpose{[ A, B, C ]}$ can be written as
\begin{equation}
    \operatorname{e}_\mu (\vect{x}, \theta) = \left( A^2 + B^2 \right)^{-1/2} \ \begin{bmatrix} \vect{x}^T \ 1 \end{bmatrix} \theta.
\end{equation}
Let $a$ be the area of the bounding box enclosing the points (data). The longest line segment in the bounding box has length $D$, where $D$ is the diagonal length of the bounding box. For a given line (model) $\ell$, we accept points $\vect{q}$ such that $\operatorname{e}_\mu (\vect{q}, \theta) < \delta$. We can then compute the probability of a random point lying on a band with length $D$ and width $2\delta$~\cite{desolneux08}. This is given by $2 \delta D / a$, and we set $p(\mu(\theta),\delta)$ to this value for all $\theta$.

\textbf{2D Circles.}
An object in $\set{X}$ is a 2D point $\vect{p} \in \Real^2$. We need $b=3$ points to define a circle.
The parameter vector $\theta = \begin{bsmallmatrix} \vect{c} \\ \rho\end{bsmallmatrix}$ of a circle, with center $\vect{c} \in\Real^2$ and radius $\rho \in \Real^+$, passing through three points $\vect{p}$, $\vect{p}'$, $\vect{p}''$ can be found by solving the system of equations
$\norm{\vect{p} - \vect{c}}{2}^2 = \norm{\vect{p}' - \vect{c}}{2}^2 = \norm{\vect{p}'' - \vect{c}}{2}^2 = \rho^2$.
A circle is the set of points (see Equation~(\ref{eq:model})) such that
\begin{equation}
    \mu(\begin{bsmallmatrix} \vect{c} \\ \rho\end{bsmallmatrix}) = \{ \vect{x} \in \Real^2 ,\ \norm{\vect{x} - \vect{c}}{2} = \rho \} ,
\end{equation}
and the distance between a point $\vect{x}$ and a circle (see Equation~(\ref{eq:pointModelDistance})) can be written as
\begin{equation}
    \operatorname{e}_\mu (\vect{x}, \begin{bsmallmatrix} \vect{c} \\ \rho\end{bsmallmatrix}) = | \norm{\vect{x} - \vect{c}}{2} - \rho\ |.
\end{equation}

The probability of a random point lying on a band of width $2\delta$ around a circle with radius $\rho$ is given by $\pi \left[ (\rho+\delta)^2 - (\rho-\delta)^2 \right] / a$, where $a$ is the area of the bounding box enclosing the data. We set $p(\mu(\theta),\delta)$ to this value.

\textbf{3D Planes.}
An object in $\set{X}$ is a 3D point $\vect{p} \in\Real^3$. We need $b=3$ points to define a plane.
The parameter vector $\theta \in \Real^4$ of a plane passing through three points $\vect{p}$, $\vect{p}'$, $\vect{p}''$, can be found by solving the system of equations
$
    \transpose{\begin{bsmallmatrix} \vect{p} & \vect{p}' & \vect{p}'' \\ 1 & 1 & 1
    \end{bsmallmatrix}}
    \theta = 0
$.
A plane is the set of points (see Equation~(\ref{eq:model})) such that
\begin{equation}
    \mu(\theta) = \{ \vect{x} \in \Real^3 ,\ \begin{bmatrix} \vect{x}^T \ 1 \end{bmatrix} \theta = 0 \} ,
\end{equation}
and the distance between a point $\vect{x}$ and a plane (see Equation~(\ref{eq:pointModelDistance}))  with parameter vector $\theta = \transpose{[ A, B, C, D ]}$ can be written as
\begin{equation}
    \operatorname{e}_\mu (\vect{x}, \theta) = \left( A^2 + B^2 + C^2 \right)^{-1/2} \ \begin{bmatrix} \vect{x}^T \ 1 \end{bmatrix} \theta.
\end{equation}

Let $r$ be half the diagonal length of the 3D bounding box enclosing the points (data).
We compute the probability of a random point lying on a band of width $2\delta$ around a plane. This can be approximated by $2 \pi r^2\delta / (\tfrac{4}{3} \pi r^3)$, and we set $p(\mu(\theta),\delta)$ to this value for all $\theta$.

\textbf{Ellipses in images.}
An object in the image is a line segment, detected using LSD~\cite{grompone10}. We use $b=3$ segments to define an ellipse.
We can define an ellipse, with parameter vector $\theta = \transpose{[ A, B, C, D, E, F ]}$, as the set of points (see Equation~(\ref{eq:model})) such that
\begin{equation}
    \mu \left( \transpose{[ A, B, C, D, E, F ]} \right) = \{ \transpose{[x, y]} \in \Real^2 ,\ A x^2 + B x y + C y^2 + Dx + E y + F = 0 \} ,
\end{equation}
where $B^2 -4AC < 0$.
An ellipse passing through three line segments can be found by solving the system of equations detailed in~\cite{halir1998}.
We can compute the distance between a segment, with endpoints $\vect{x},\vect{x'}$, and an ellipse (see Equation~(\ref{eq:pointModelDistance})) as
\begin{equation}
	\operatorname{e}_\mu \left( \{ \vect{x},\vect{x'} \}, \transpose{[ A, B, C, D, E, F ]} \right) = \max_{ \vect{y} \in \mu \left( \transpose{[ A, B, C, D, E, F ]} \right) }
	\left\{
	\begin{gathered}
	\norm{ \vect{x} - \vect{y} }{2} \\
	\norm{ \vect{x'} - \vect{y} }{2}
	\end{gathered}
	\right\} .
\end{equation}
Solving this for each endpoint involves finding the roots of a quartic polynomial.
The probability of a random point lying on a band of width $2\delta$ around an ellipse with radii $\rho, \rho'$, is given by $\pi \left[ (\rho+\delta)(\rho'+\delta) - (\rho-\delta)(\rho'-\delta) \right] / a$, where $a$ is the area of the bounding box enclosing the data. We set $p(\mu(\theta),\delta)$ to this value.

\textbf{Vanishing points.}
We refer the reader to~\cite{tepper14vp} for a description of the a contrario test used in this case.

\bibliographystyle{hsiam}
%\bibliography{mtepper-clustering,mtepper-nmf,mtepper-communityDetection,mtepper-biclustering,mtepper-networks,mtepper-graph,mtepper-mpme,mtepper-perception}

\end{document}